\documentclass[a4paper,fleqn]{cas-sc}

\usepackage[utf8]{inputenc}
\usepackage[polish]{}
\usepackage{color}
\usepackage{subcaption}
\usepackage{tikz,pgfplots}
\pgfplotsset{compat=1.15}
\usepackage[nolist]{acronym}%
\usepackage[numbers,square,sort&compress]{natbib}%
\usepackage{amsmath,amssymb,amsfonts,amsthm}
\usepackage{bm}%
\usepackage{ulem}
\theoremstyle{plain}
\usepackage[labelfont=bf, font={small,sf}, hypcap=false]{caption}
\usepackage{multirow}
\usepackage{float}

\def\tsc#1{\csdef{#1}{\textsc{\lowercase{#1}}\xspace}}
\tsc{WGM}
\tsc{QE}
\tsc{EP}
\tsc{PMS}
\tsc{BEC}
\tsc{DE}

\begin{document}
\let\WriteBookmarks\relax
\def\floatpagepagefraction{1}
\def\textpagefraction{.001}
\shorttitle{An automatic selection of optimal recurrent neural network architecture}
\shortauthors{Laddach, K. et~al.}
\title [mode = title]{An automatic selection of optimal recurrent neural network architecture for processes dynamics modelling purposes}

\author[1]{Krzysztof Laddach}[orcid=0000-0001-9122-2167]
\address[1]{Department of Intelligent Control and Decision Support Systems, Gda\'nsk University of Technology, G. Narutowicza 11/12, 80-233 Gda\'nsk, Poland}
\author[1]{Rafa\l{} \L{}angowski}[orcid=0000-0003-1150-9753]
\cormark[1]
\ead{rafal.langowski@pg.edu.pl}
\author[1]{Tomasz A. Rutkowski}[orcid=0000-0001-8818-6126]
\author[1]{Bartosz Puchalski}[orcid=0000-0001-9834-6250]
\cortext[cor1]{Corresponding author}%

\begin{abstract}
A problem related to the development of algorithms designed to find the structure of artificial neural network used for behavioural (black-box) modelling of selected dynamic processes has been addressed in this paper. The research has included four original proposals of algorithms dedicated to neural network architecture search. Algorithms have been based on well-known optimisation techniques such as evolutionary algorithms and gradient descent methods. In the presented research an artificial neural network of recurrent type has been used, whose architecture has been selected in an optimised way based on the above-mentioned algorithms. The optimality has been understood as achieving a trade-off between the size of the neural network and its accuracy in capturing the response of the mathematical model under which it has been learnt. During the optimisation, original specialised evolutionary operators have been proposed. The research involved an extended validation study based on data generated from a mathematical model of the fast processes occurring in a pressurised water nuclear reactor.
\end{abstract}


\maketitle

\begin{acronym}[TDMA]
\acro{ann}[ANN]{artificial neural network}%
\acro{rnn}[RNN]{recurrent artificial neural network}%
\acro{dnas1}[DNAS1]{first developed NAS algorithm}%
\acro{dnas2}[DNAS2]{second developed NAS algorithm}%
\acro{dnas3}[DNAS3]{third developed NAS algorithm}%
\acro{dnas4}[DNAS4]{fourth developed NAS algorithm}%
\acro{ea}[EA]{evolutionary algorithm}%
\acro{pwr}[PWR]{pressurized water reactor}%
\acro{nas}[NAS]{neural network architecture search}%
\acro{siso}[SISO]{single-input single-output model}%
\acro{neat}[NEAT]{NeuroEvolution of Augmenting Topologies}%
\end{acronym}

\section{Introduction}\label{sec:introduction}
Nowadays, the use of advanced algorithms involved in the operation of widely understood industrial plants is very strongly related to the availability of accurate mathematical models of processes that occur in these plants. It can distinguish algorithms that perform the tasks of monitoring, diagnostics, estimation, or advanced control, etc. Therefore, the quality of performance of the above-mentioned algorithms tasks is closely related to the quality of mathematical models, but also to their time availability or the possibility of performing many simulations of the process with the required time regimes, as it is the case, for example, in a widely applied model predictive algorithm \cite{Tatjewski2007}.

Typically, a mathematical model of a given process (system) can be devised in the phenomenological or experimental (behavioural) way \cite{Roffel2006}. In the first case resulting model is called a white-box model and is based on the conservation laws (physics, chemistry, biology, etc.), and in general, describes the phenomena occurring in a given system. Hence, the white-box model is derived analytically, and its abstract mathematical structure is fundamentally related to the physical structure of processes, and the model parameters have a physical meaning and interpretation. On the other hand, a mathematical model developed experimentally (behaviourally) is referred to as a black-box model and it is built based on observation of the behaviour of a given system. Thus, this model is derived experimentally, and its structure does not have to be essentially related to the structure of the process, and the model parameters do not have a physical interpretation. Naturally, each of these types of models has both advantages and disadvantages. However, because the complexity of some plants is significant as well as the phenomena occurring in them are sophisticated, the white-box modelling might become difficult, time-consuming, expensive, and in the worst-case impossible to perform. Moreover, in many cases, the values of white-box model parameters are not exactly known. Therefore, developing a black-box model or a certain kind of hybrid of white- and black-box models so-called grey-box model may be more justified and reasonable. In this paper, black-box modelling is considered.

The black-box model of a given system may be provided using various types of tools. They are commonly based on either a statistical analysis of time series or computational (artificial) intelligence. The first group includes the input-output models such as, e.g., linear autoregressive moving-average model with exogenous inputs (ARMAX) or non-linear autoregressive moving average model with exogenous inputs (NARMAX) model structure \cite{Billings2013,Candy2006}. The ARMAX and NARMAX models, approximate the input-output system behaviour by the linear or non-linear difference equations defined in the finite-dimensional linear or non-linear discrete-time domain, respectively. In general, the process of identifying a black-box model involves determining the structure of the unknown linear or non-linear difference equation, estimating its parameters, and finally checking or validating the resulting model to ensure that it describes the modelled system accurately. A wide class of dynamical systems can be approximated with sufficient accuracy only by the linear expressions involving the variables which characterise the system. However, there are many practical cases when a linear description of a process is not sufficient, and a global, more accurate non-linear model is required. Additionally, the determination of the structure of non-linear functional dependencies between inputs and outputs of the modelled system is not an obvious and trivial task. Typically, for the group of the linear black-box models, the polynomial model structures are used. For the non-linear models, to overcome mentioned limitations, the common practice approximates the unknown high-dimensional and non-linear functional dependencies by using well-known and well-suited for those purposes methodologies using polynomials, wavelets, neural networks, or hybrid neural-fuzzy estimators. With those technologies, the high-dimensional and non-linear function is approximated by a set of appropriately organised lower-dimensional functions. A large group of those technologies is classified as the black-box models inspired by artificial intelligence \cite{Chen1996}. In this paper, a methodology based on \acp{ann} and their recurrent implementation, which are well-known to be universal approximators of non-linear dynamic systems, will be further considered \cite{Rios:2020,Kurkova1992,Norgaard:2000, Cybenko:1989,Hornik:1989}. In recent years, many interesting applications of ANN have been found in the literature, include classification tasks \cite{Dreiseitl2002,Saxena2007,SaiSindhuTheja:2021}, image (pattern) recognition \cite{Agarwal2010,Bhattacharyya2011,Krizhevsky2017}, smell recognition \cite{Brudzewski2004, Llobet1999}, speech recognition \cite{Dede2010}, text generation \cite{Sutskever2011}, prediction purposes \cite{Tam1992,Ahn2000, Kashiwao:2017}, modelling and control of dynamic systems \cite{Rios:2020, Norgaard:2000, Perrusquia:2021, Yu:2021}, state estimation, generating control signals and operating as diagnostic systems \cite{Sun2013,Stubberud1995, Gadoue2009}, fractional order operators approximations \cite{Puchalski2020a}, and many others, e.g., \cite{Fukuda1992,Ponulak2011}.

A common feature of practically all \acp{ann} applications is the need to select their architecture (optimal structure/topology) so that their performance, in terms of accuracy and time complexity, will be as high as is possible and satisfactory for the user. In particular, attributes/components of \ac{ann} such as neuron model, the connections between neurons, the number of layers, the number of neutrons in each layer, level of delays, and net input and activation functions should be determined. In a classic approach, manual selection of the optimal structure of \ac{ann} manually is not a trivial task - the exponential dependence of the number of parameters that the user should optimally set to the complexity of the \ac{ann} topology \cite{Haykin2011, Kurkova1992}. At the next stage, it is still necessary to train the chosen ANN structure, with the following considerations: the problem of initialisation of the weights, the type of learning algorithm, the lack of definition of the optimal number of iterations for the learning algorithms, the size of the training set and the value of the learning rate coefficient. Overall, the impact of the aforementioned architecture-related ANN attributes and its learning process on the \acp{ann} performance is verified using a trial-and-error approach. It should also be noticed that, in the case of the over-sized and under-sized \acp{ann} structure, the over-fitting and under-fitting problems (poor generalization, trap in local solution) can take place. In this area, the approach based on dividing a data set into training, validation, and testing subsets and further evaluation of networks using all subsets to minimise over-fitting may be useful \cite{Tetko1995}.

In general, the problem of neural \ac{nas} is still an open issue. It is because there are no general and certain rules in this topic. Hence, experimental methods based on heuristics, experience and intuition are often in favour by various authors. For example, the \ac{nas} based on the trial-and-error method and in-depth analysis of the obtained results has been discussed in \cite{Kavzoglu1999}. However, all these methods require considerable user involvement, and they are time-consuming. Therefore, a fully or partially automatic selection of \acp{ann} architecture is an attractive alternative. One of the approaches enabling such operations is to automate the trial-and-error technique using a computing environment with dedicated optimisation tools (solvers) that meet especially a multi-objective optimisation requirement \cite{Ellefsen:2020} (e.g. minimal \ac{ann} topology, and maximal \ac{ann} accuracy). Examples of such solvers are nature-inspired optimisation algorithms, which are commonly used to solve optimisation tasks for a wide class of real-world complex problems. An interesting and comprehensive review of this type of algorithms from the groups of evolutionary algorithms (e.g. genetic algorithm, evolutionary strategy, differential evolution), swarm intelligence (e.g. ant colony and particle swarm optimisation) or those inspired by the laws of physics and chemistry (e.g. simulated annealing), which were used for the optimal, partial or full selection of the \ac{ann} structure, can be found in \cite{Gupta:2019}. In addition to references to individual publications, an overview of articles from over the last two decades, the study indicates \cite{Gupta:2019}: (i) the range of automatically selected, optimised \ac{ann} parameter/parameters divided into seven groups: architecture and weights, connection and weights, hidden neurons, hidden neurons and hidden layers, hidden layers, hidden neurons and connections weights and bias; (ii) used optimisation algorithm; (iii) and its observed strengths and weaknesses.  The other papers related to using artificial intelligence that uses evolutionary algorithms to generate \ac{ann} can be found in the literature under the neuroevolution topic \cite{Yaot1993, Floreano2008,Nolfi1997,Siebel2007, Stanley2019, Stanley2002, Turner2014}. A wide overview of different aspects of neuroevolution methods and discusses their potential for application in the field of deep learning may be found in \cite{Stanley2019}. One well-known algorithm from this group is NeuroEvolution of Augmenting Topologies (NEAT) presented almost two decades ago \cite{Stanley2002,Stanley2009}. The basic \ac{neat} is a Topological and Weight Evolving Artificial Neural Networks (TWEANN) method that enables the learning of the structure of \acp{ann} at the same time it optimises their connectivity weights. The \ac{neat} method is based on direct encoding to encode the phenotypes (\acp{ann} structure) in the genotype, specialised operators of crossover, mutation and speciation (the introduction of the concept of species -- sub-populations), and other crucial issue related to the identification of similar neural network structures evolving independently in population and assigning each structure its historical markings (innovation number). They act as chronological indicators that facilitate crossover by identifying homologous sections between different neural networks. The innovation number of each gene is inherited by the offspring, facilitating the retaining of its historical origin throughout evolution. Generally, neural networks are grouped in appropriate sub-population based on their topological similarities expressed as compatibility distance. The sub-populations protects topological innovations in the neural network structure, and such individuals compete within their own niche instead of the entire population. The \ac{neat} algorithm starts the evolution with minimal structure and, in further iterations, introduce new nodes and connections via mutations (evolve to more complex networks structures) as long as they find useful after fitness evaluation. In the paper, \cite{Papavasileiou:2021}, comprehensive categorisation of the NEAT algorithms successors found in the literature is described in detail. Historically, the introduction of species has been also introduced in earlier studies, e.g., \cite{Michalewicz1996}. Another approach is based on genetic algorithms, where \acp{ann} are used to observe the state of the nuclear reactor for diagnostic purposes \cite{Basu1992, Bartlett1991}. In this approach, the "importance" of particular neurons, and thus the probability of their survival, depends on the influence of the particular neuron on the output of the entire network - the close link between the \acp{ann} structures and the problems for which they are addressed is shown.

In this paper, the authors' neuroevolution methods inspired by various methods described in the literature \cite{Gupta:2019, Azzini:2011, Papavasileiou:2021}, especially \acp{ea} to build a black-box model of a non-linear dynamic process are presented. The authors developed and simulation-verified four \ac{nas} algorithms. In general, the proposed algorithms using the $\mu + \lambda$ evolutionary strategy. The first two using roulette-based selection and elitism mechanism. In comparison to the first, the second one includes a specialised neuron mutation-deleting operator. In turn, the third and fourth algorithms incorporate the species concept. In comparison to the third, the fourth algorithm is distinguished because the weights of connections between neurons are selected using a gradient descent learning method (the hybrid algorithm, where \ac{ea} still selects \ac{ann} structure) and not jointly by \ac{ea} (\ac{ann} structure and weights). For all the above-mentioned algorithms, specialised evolutionary operators have been developed, i.e. mutation and crossover for weights and neurons (for new neuron formation or neuron death), respectively.  All algorithms use the direct encoding type to store the network structure. Additionally, proposed algorithms satisfy the postulate of automatic selection of the neural network structure to various degrees, ranging from the algorithm that selects two parameters of \ac{ann} (number of neurons in the hidden layer and connection weights) to the algorithm that selects six parameters of \ac{ann} (number of hidden layers, number of neurons in the hidden layer, input delay level, output delay level (feedback), connection weights and learning method). As an application the dynamics of fast processes in a \ac{pwr} is taken into account. A nuclear reactor is a non-linear, spatial, and non-stationary plant that belongs to the elements of critical infrastructure. Its processes are characterised by multi-scale and complex dynamics. Because of these reasons, there are many different mathematical models of nuclear reactors, which are used depending on the purpose, e.g., synthesis of control algorithms, modelling of physical processes, diagnostics, on-line monitoring, power demand scheduling, or fuel campaign planning. In this study, a mathematical model of \ac{pwr} reactor that has been originally developed for diagnostics and control purposes has been used \cite{Naghedolfeizi1990}. The model consists of a sub-model responsible for the description of point neutron kinetics in the reactor core with six groups of delayed neutron precursors, a sub-model responsible for the description of thermohydraulic phenomena related to heat exchange between the core and the coolant, a module for calculating reactivity feedbacks from fuel and coolant temperatures, and a sub-model of the actuator, which mimics the operation of control rod drive mechanism. The model uses a thermal-hydraulic structure consisting of a single fuel node - F and two coolant nodes - C (1F/2C) which is described in \cite{Naghedolfeizi1990,Puchalski2017,Puchalski2020,Kerlin1978}. The model of the actuator used is given in \cite{Puchalski2020}, whereas the overall mathematical model equation structures together with the necessary parameters are given in \cite{Puchalski2017,Puchalski2020}. The mathematical model of the \ac{pwr} presented in the aforementioned works takes into account only fast processes taking place in the reactor, i.e. neutron kinetics and heat exchange between the fuel and coolant. From the research point of view addressed in the paper, the nuclear reactor model used is certainly a non-trivial and challenging case. It should be noted also that the model has been used only to generate learning, validation, and test data, and mainly for this reason a detailed description of it is not presented in the body of the paper as modelling of processes taking place in a nuclear reactor is not the subject of the presented research. 

To summarise the main aim of this work is to develop and verify the authors' algorithms of optimal artificial neural network architecture search for black-box (behavioural) modelling purposes. As a type of \ac{ann}, the recurrent network has been chosen - as a universal approximator of a non-linear dynamic system. Its architecture is automatically selected by solving an appropriately defined optimisation task. The obtained neural network operates as a black-box model of the fast processes in a \ac{pwr}. As an input to the black-box model the position of control rods is used whereas the scaled thermal power of a \ac{pwr} is an model output. The optimality has been understood as achieving the desired trade-off between the size of obtained \ac{ann} and the accuracy of black-box model responses. Hence, the main contributions of this paper are as follows:
\begin{itemize}
    \item four algorithms of \acp{nas} using either \ac{ea} based itself or \ac{ea} with gradient descent learning methods, which create the desired recurrent neural network models are devised and verified based on the selected case study (fast processes in a \ac{pwr}),
    \item the specialised evolutionary operators, i.e. mutation and crossover for weight, neuron (for new neuron formation or neuron death) and delays for proposed \ac{nas} algorithms are delivered and their performance is verified in comparison to the basic NEAT algorithm and exhaustive search algorithm,    
    \item the black-box model of the selected processes in a \ac{pwr} is obtained with delivered by authors \acp{nas} algorithms and verified based on the black-box models delivered by the basic \ac{neat} algorithm and exhaustive search algorithm.
\end{itemize}

The paper is organised as follows. The problem formulation is presented in section \ref{sec:problem_statement}. The four various authors' algorithms of artificial neural network architecture search for black-box modelling purposes are delivered in section \ref{sec:Designed_NAS_agorithms}. The obtained \ac{siso} of the fast processes in \ac{pwr} and the performance of its operation verified in a simulation way are described in section \ref{sec:case_study}. The paper is concluded in section \ref{sec:conclusions}.

\section{Problem statement}\label{sec:problem_statement}
In general, a concept of \acp{ann} is based on an investigation of processes taking place in biological neural networks \cite{Bhushan2009,Kandel2012,Vincent2006}. Typically, \acp{ann} can be classified in the following way \cite{Bishop2006,Alom2019,Haykin2011}:
\begin{itemize}
    \item feed--forward artificial neural networks (FNNs) – where signals are transmitted only in one direction, i.e. from the network input to its outputs through the network layers;
    \item \acp{rnn} – these networks are characterised by an internal state; in other words there are feedbacks in the \acp{rnn} from the outputs of individual neurons to their inputs, or from the network outputs to its inputs; this causes that a change in the state of individual neuron can be transferred through feedback to the other neurons, invoking transient states and generally leading to another state of the network; thus thanks to feedbacks \acp{rnn} have their internal state, which allows them to model dynamic plants;
    \item cell artificial neural networks (CeNNs) – where connections between particular neurons occur only in the closest neighbourhood; these connections are generally non-linear and described by a system of differential equations; this type of networks is mainly used for clustering of input data, and often the method of teaching them, in which the teacher does not exist (unsupervised learning), is based on the Hebb rule; an example of CeNNs is a Kohonen’s map of features \cite{Alom2019}.
\end{itemize}

As it has been mentioned in section \ref{sec:introduction} as an application the fast processes occurring in \ac{pwr} are considered. Similarly to the vast majority of real plants, also \ac{pwr} has internal feedbacks. Thus, the natural choice of a type of \ac{ann} has been \ac{rnn} but it has been decided to use only feedback from the network output to its input in this paper.Searching for a network with such architecture is equivalent to looking for a dynamic function meeting the discrete equation (\ref{eq:discrete equation}):
\begin{equation}\label{eq:discrete equation}
    y(k)=f\left(y(k-1),y(k-2),y(k-3),...,y(k-n),u(k),u(k-1),u(k-2),u(k-3),...,u(k-m) \right),
\end{equation}
where:
$y(\cdot)$ is the output at the discrete--time instant specified by $(\cdot)$;
$u(\cdot)$ denotes the input at the discrete--time instant specified by $(\cdot)$;
$f(\cdot)$ signifies the function specified by $(\cdot)$;
$k, m, n$ are the discrete--time instants.

\begin{figure}
\centering
    \includegraphics[width=0.8\textwidth]{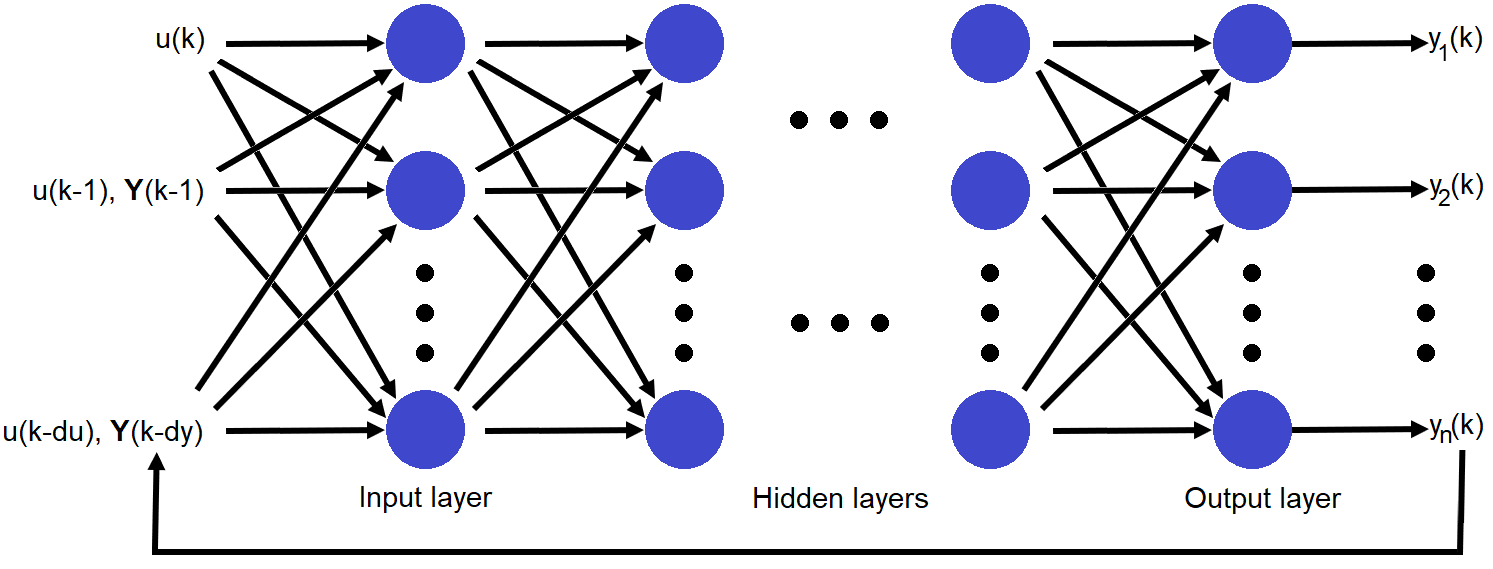} 
    \captionof{figure}{An architecture of exemplary \ac{rnn}.}
    \label{fig:RANN_structure}
\end{figure}

In turn, an architecture of exemplary \ac{rnn} is presented in Fig.~\ref{fig:RANN_structure}. 

As it can be noticed the following layers are distinguished:
\begin{itemize}
    \item an input layer, which is responsible for the normalisation of data (the most often it is re-scaling the input data to a given range),
    \item hidden layers, which are responsible for the signals processing, and,
    \item output layer, there are at least as many neurons in it as there are outputs from the network.
\end{itemize}

The particular symbols in Fig.~\ref{fig:RANN_structure} denote:\\
$\bm{Y}(\cdot)$ -- the vector of feedback outputs at the discrete--time instant specified by $(\cdot)$;\\
$y_{i}(\cdot)$ -- the outputs at the discrete--time instant specified by $(\cdot)$, $i=\overline{\mathrm{1},\mathrm{n}}$;\\
$du$ -- maximal level of input delay;\\
$dy$ -- maximal level of feedback outputs delay.

The main task of the first layer (input layer) is re-scaling inputs data. However, in many cases, also in this work, the inputs data are already normalised, so the input layer is skipped, and it can be understood as transmitting input signals directly to the next (hidden) layers. Therefore, the developed algorithms enable selecting of the number of hidden layers and the number of neurons in these layers in a given \ac{rnn}. Moreover, they also enable selecting of $du$ and $dy$ delays. In order to perform this task, it is necessary to define the neuron model. The simple perceptrons with the weighted sum, and with bipolar sigmoid in the hidden layers and linear activation function in the output layer are used. These activation functions are chosen primarily because of:
\begin{itemize}
    \item the monotony of the derivative ensures the correct operation of gradient methods of ANN learning, which are used in one of the developed algorithms,
    \item the bipolarity of functions increases the acceptable field of searching for the optimal solution,
    \item since the task of neurons in the output layer is the linear transformation of the sum of the previous layer outputs the linear activation functions in this layer are selected. 
\end{itemize}

\begin{figure}
\centering
    \includegraphics[width=0.65\textwidth]{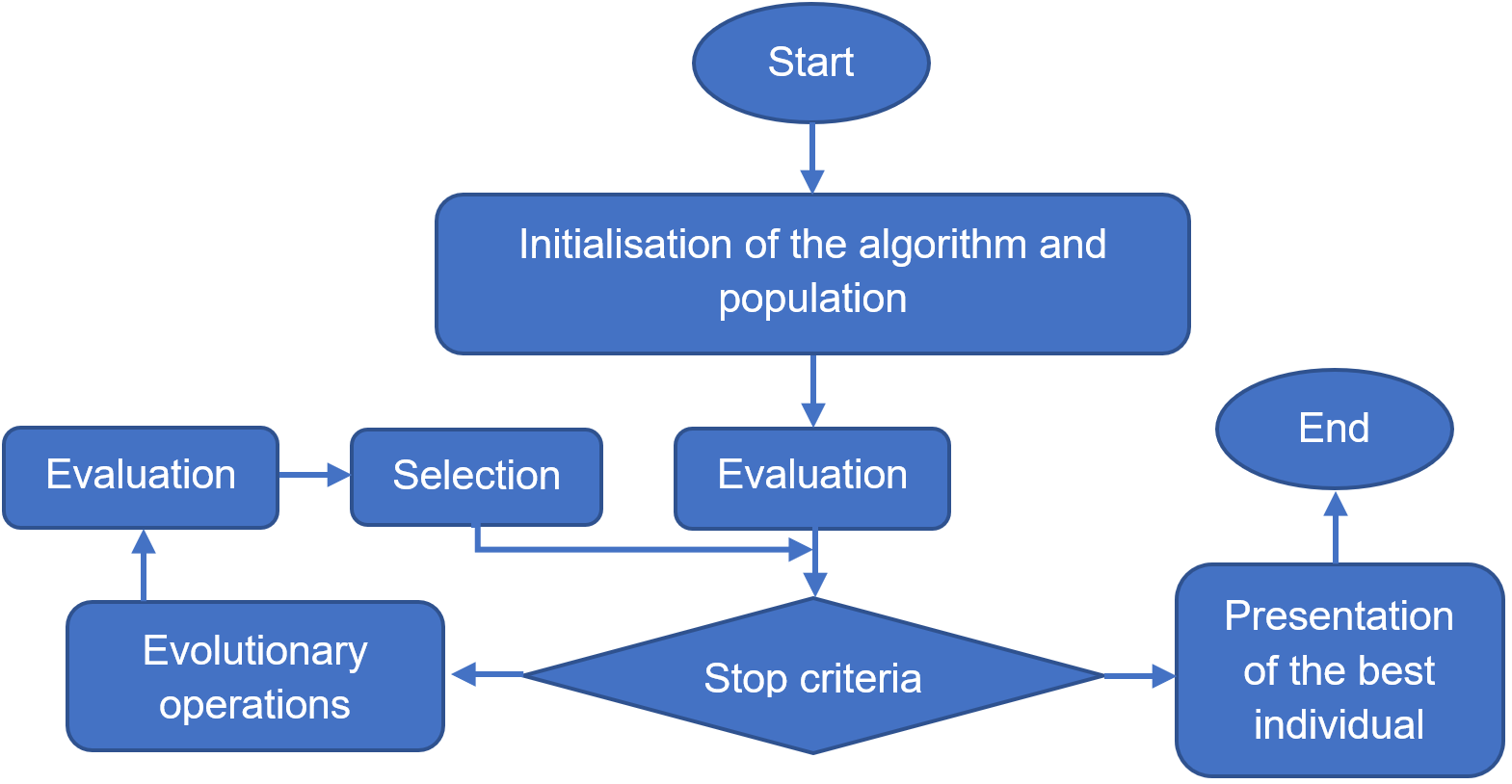} 
    \captionof{figure}{A general scheme of an evolutionary algorithm.}
    \label{fig:EA_scheme}
\end{figure}

In order to perform an automatic selection of the number of hidden layers, the number of neurons in these, the weights of neurons, and delays in a given \ac{rnn} the \acp{ea} are used either independently or in combination with gradient-based methods for the network learning process. It is since an \ac{ea}, i.a., enables solving non-trivial optimisation tasks \cite{Michalewicz1996, Stanley2019}. Undoubtedly searching for the optimal architecture of \ac{ann} belongs to a category of this type. It is because there are different types of variables involved in the task, i.e. real numbers, e.g., the values of \ac{ann} weights as well as integer and binary variables, e.g., the values of maximal levels of used delays and the number of layers and the number of neurons in each layer. A general scheme of an \ac{ea} is presented in Fig.~\ref{fig:EA_scheme}.

The development of evolutionary operators has a crucial meaning for black-box modelling when \acp{rnn} are involved. It is well-known that, in general, two types of evolutionary operators can be distinguished, i.e. mutation and crossover \cite{Michalewicz1996,Stanley2002,Koza1998,Haykin2011}. The mutation operator is primarily responsible for exploring the field of solutions whereas the crossover operator is responsible for exploiting this field. In the further part of this paper, the developed (dedicated) mutation and crossover operators are described in detail.

As it has been mentioned above, four \ac{nas} algorithms have been proposed using \acp{ea}. The \ac{dnas1} enables only the selection of the number of neurons in the hidden layer and the values of their weights. In contrast, the \ac{dnas2} and \ac{dnas3} allow the selection of the maximal levels of delays $du$ and $dy$ also, using an \ac{ea} with dedicated evolution operators. In turn, in the \ac{dnas4} an \ac{ea} is used for searching for the number of hidden layers, the number of neurons in these layers, $du$ and $dy$ and one of the three gradient-based \acp{rnn} learning methods. To summarise, the classification of the \ac{nas} algorithms proposed in the paper, from their main features point of view, is presented in table~\ref{tab:summary}. Thus, each individual in the population represents a \ac{rnn} architecture that is related to searched black-box model.

\begin{table}
   \caption{The summary of main features of proposed \ac{nas} algorithms.}
   \label{tab:summary}
   \begin{tabular}{|c|c|c|c|c|}
   \hline
   \multicolumn{5}{|c|}{\textbf{Type of \ac{ann} architecture - \ac{rnn}}}           \\ \hline \hline
{\textbf{Component name}} & {\textbf{\ac{dnas1}}} & {\textbf{\ac{dnas2}}} & {\textbf{\ac{dnas3}}} & {\textbf{\ac{dnas4}}} \\ \hline \hline  
 number of hidden layers & hyper-parameter & hyper-parameter & hyper-parameter & parameter \\ \hline
 number of neurons in hidden layers & parameter & parameter & parameter & parameter \\ \hline
 $du$ & hyper-parameter & parameter & parameter & parameter \\ \hline
 $dy$ & hyper-parameter & parameter & parameter & parameter \\ \hline
 connection weights & parameter & parameter & parameter & parameter \\ \hline
 learning methods & hyper-parameter & hyper-parameter & hyper-parameter & parameter \\ \hline
 number of neurons in output layer & hyper-parameter & hyper-parameter & hyper-parameter & hyper-parameter \\ \hline
 model of neuron & hyper-parameter & hyper-parameter & hyper-parameter & hyper-parameter \\ \hline
 net input function & hyper-parameter & hyper-parameter & hyper-parameter & hyper-parameter \\ \hline
 activation function in hidden layers & hyper-parameter & hyper-parameter & hyper-parameter & hyper-parameter \\ \hline
 activation function in output layer & hyper-parameter & hyper-parameter & hyper-parameter & hyper-parameter \\ \hline
\end{tabular}
\end{table}

\section{Designed \ac{nas} algorithms}\label{sec:Designed_NAS_agorithms}
During the developing of each \ac{nas} algorithm a certain set of input parameters has been determined. These parameters are listed in table~\ref{tab:input_parameters}.
\begin{table}
   \caption{A set of input parameters for proposed \ac{nas} algorithms.}
   \label{tab:input_parameters}
   \begin{tabular}{|c|c|}
   \hline
   \multicolumn{2}{|c|}{\textbf{Input parameters} } \\ \hline \hline 
   \textbf{Symbol} & \textbf{Description} \\ \hline
    $maxLay$ & the maximal number of hidden layers \\ \hline
    $maxNinLay$ & the maximal initial number of neurons in each hidden layer \\ \hline
    $du$ & the maximal initial level of input delay \\ \hline
    $dy$ & the maximal initial level of feedback outputs delay \\ \hline
    $popSize$ & the initial number of the population individuals\\ \hline
    $pCross$ & the probability of crossover \\ \hline
    $p_i$ & the values of weights in a fitness function \\ \hline
    $minDelta$ & the minimal module of changing the weights in mutation operator \\ \hline
    $maxDelta$ & the maximal module of changing the weights in mutation operator \\ \hline
    $pMutW$ & the probability of weights mutation \\ \hline
    $pMut$ & the probability of mutation in \ac{dnas4} \\ \hline
    $pMutNewN$ & the probability of generation of a new neuron \\ \hline
    $pMutD$ & the probability of mutation of delays \\ \hline
    $pMutDelN$ & the probability of delete of neuron \\ \hline
    $minW$ & the minimal initial weights values \\ \hline
    $maxW$ & the maximal initial weights values \\ \hline
    $hmBest$ & the number of the best individuals subject to elitism \\ \hline
    $pRetrain$ & the probability of selecting an individual for retraining \\ 
    \hline
\end{tabular}
\end{table}
\subsection{\ac{dnas1} -- \ac{dnas3} algorithms} \label{subsec:dnas1dnas3}

In this section the \ac{dnas1} -- \ac{dnas3} algorithms, which operate according to general scheme presented in Fig.~\ref{fig:EA_scheme} are described. Firstly the algorithms initialisation, fitness function evaluation, selection and dedicated evolutionary operators, i.e.  mutation and crossover are presented. Next, the unique features of a given algorithm are discussed.

\subsubsection*{Initialisation}\label{subsec:initialisation}

At the beginning of operation of the \ac{dnas1} -- \ac{dnas3} algorithms each individual in an initial population is created in accordance with the following procedure. Firstly, the values of $du$ and $dy$ are randomly taken from the ranges (0, $duMax$) and (0, $dyMax$), respectively or they are assumed to be fixed (see tables \ref{tab:input_parameters} and \ref{tab:input_parameters_values}). It is worth adding that, the assumed large initial values of $du$ and $dy$, on the one hand, can improve the mapping of target responses, but on the other hand, it directly degrades the individual fitness evaluation. Next, the number of neurons for the hidden layer is randomly selected from the range of (1, $maxNinLay$). In turn, in the output layer, only one neuron occurs because the network has only one output -- \ac{siso} model of \ac{pwr}. Then, for each neuron, a random value for appropriate number of ($N_\mathrm{rw}$) is drawn within the range ($minW$, $maxW$). The appropriate number of random weights is understood here as the number of inputs to each neuron that results from the network structure. For the first layer, it is:
\begin{equation}\label{eq:init}
    N_\mathrm{rw} = u_k + du + dy + b,
\end{equation}
where: $u_k$ = 1 denotes current input sample; $b$ = 1 stands for the bias.

Whereas for the next layers, the number of needed weights equals the number of neurons in the previous layer increased by one, i.e. the weight corresponding to the bias. The value of the bias weight can be understood as a direct value of the bias.

\subsubsection*{Mutation}\label{subsec:mutation}

As it has been aforementioned, during operation of \acp{ea} mutation is mainly responsible for exploring the field of decision space. In the \ac{dnas1} -- \ac{dnas3} algorithms, the mutation of individual weights are designed and programmed in the following way. The probability of mutation for each weight, as the value of one of the algorithm parameters, has been set to $pMutW$. Then a random decision is taken as to whether or not there will be a mutation of a given weight. The weight is changed by adding to it the corresponding value from the $\Delta w$ vector. The values of $\Delta w$ vector depends on the value of the fitness evaluation of the individual for whom the mutation occurred. The formula linking the value of the fitness evaluation with the $\Delta w$ has been developed so that the weights of better-adapted individuals change by smaller values. Assuming that $fit$ is a vector of the values of the individual fitness evaluation, the procedure for calculating the $\Delta w$ vector is as follows:
\begin{align}\label{eq:fitness_evaluation}
    fit &= -fit,\\
    fit &= fit + |min(fit)|,\\
    minFit &= min(fit),\\
    maxFit &= max(fit),\\
    \Delta w &= \frac{\left(maxDelta - minDelta\right) \left(fit - minFit\right)}{\left(maxFit - minFit\right)} + minDelta.
\end{align}

\begin{figure}
\centering
    \includegraphics[width=0.5\textwidth]{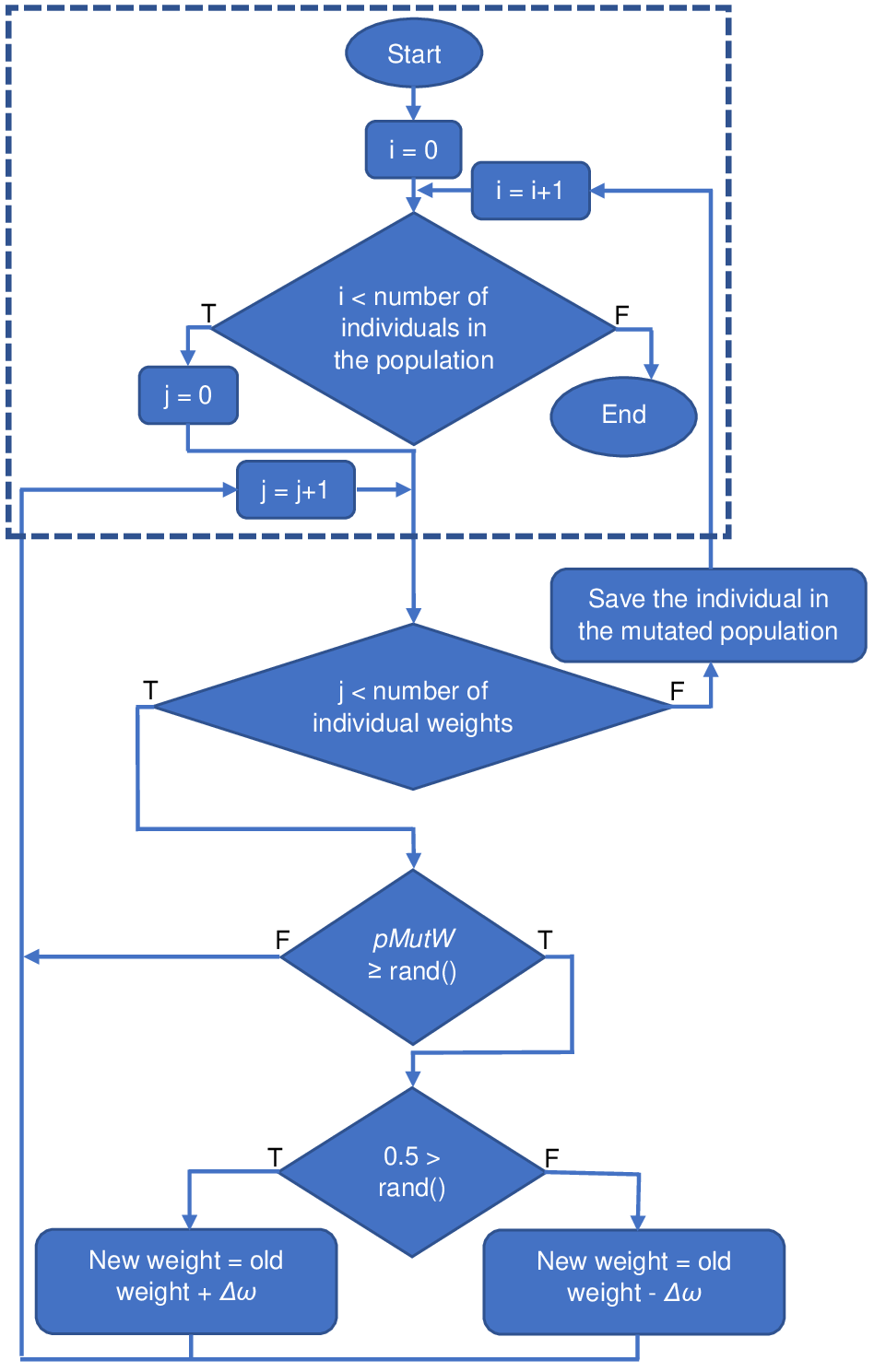} 
    \captionof{figure}{A functional diagram of the weight mutation operator.}
    \label{fig:m_weight_scheme}
\end{figure}

The vector $fit$ is first inverted, i.e. multiplied by -1, in order to assign the highest value resulting from the fitness evaluation to the individual who is the least fit. Then it is ensured that all its elements are positive, and next, the maximum and minimum values from the $fit$ vector are found. In the last line of the procedure, the $\Delta w$ vector is calculated. Hence, it is a linear mapping of the value of the $fit$ vector into a range of ($minDelta$, $maxDelta$). The results of this procedure are that for the best-fit individual the module of change is $minDelta$, and for the least fit $maxDelta$. These values are the parameters of a given algorithm, just like the probability of weights mutation $pMutW$ (see table \ref{tab:input_parameters}). If a mutation occurred in one or more weights in a given individual, a new individual containing this new weight or new weights are added to the population of mutated individuals. A functional diagram of the weight mutation operator is shown in Fig.~\ref{fig:m_weight_scheme}. The dashed blue box in Fig.~\ref{fig:m_weight_scheme} marks the part of the diagram that is shared with the following figures (Figs. \ref{fig:m_neuron_scheme} and \ref{fig:m_delays_scheme}). Thus, this part is not repeated on them.

It should be noticed that a feature of such evaluation of the weights adjustment is the fact that the adjustment for the most poorly fitted individuals is bigger. Moreover, this mechanism reduces the risk that an individual who is close to the optimum will jump over it because the change in weights will be smaller for him. Compared to conventional solutions, in which the weights changes values depending on the generation number, i.e. the longer the algorithm operates, the weights change by smaller value, the proposed approach eliminates the problem of marginal weight changes in the final stage of the algorithm operation \cite{Michalewicz1996}. Moreover, it allows the small value to change the more accurate exploitation of the most promising areas - the weights of the most-fitted individuals. The proposed solution also has an advantage over approach, where the value of the changes depends on the average value of the fitness function of individuals of the entire population \cite{Michalewicz1996}. It is because, it is not necessary to determine a level of satisfaction, i.e. the value of the fitness function at which the changes would reach the minimum level.

\clearpage

It is well-known that every human being during its life is subject to a continuous learning process. It should be noticed that learning is not only improving what the brain already knows (improving network’s weights – weights mutation), and what once had to be learned, but also learning completely new things. This process occurs due to the building of new connections between existing neurons and by creating new neurons that build new paths for nerve signals, thus allowing the network to extend its capability. This fact in neuro-science is known under the broad concept of neuroplasticity \cite{Bishop2006}. A similar feature may be required in the studied \acp{ann}, which even after achieving optimal weights for a given structure, may not sufficiently map learning data. According to Kolmogorov's theorem \cite{Kurkova1992}, the solution to this problem is to increase the number of neurons in the network structure. The above ideas have been implemented in devised algorithms in the following way. For each individual in the population, the number of free places, where the neuron could occur, is calculated according to the formula:
\begin{equation}\label{eq:number_free_places}
    N_\mathrm{fp} = maxNinLay - N_\mathrm{an},
\end{equation}
where: $N_\mathrm{fp}$ denotes the number of free places; $N_\mathrm{an}$ is an actual number of neurons in a given layer.

Then, for each free place with a probability equal to $pMutNewN$, a draw takes place to determine whether a new neuron is to be created in it. If the draw is successful a new neuron with random weights and bias is generated in the same way as it has been described in the initialisation part of this section. The newly formed neuron is always located in the first free place in a given layer. Of course, the creation of a new neuron requires modification of neurons’ weights in the next layer. Hence, for each neuron in the next layer, a new random weight is added at the place corresponding to the connection with the newly formed neuron. Each addition of a neuron to an individual creates a new individual in the mutated population. The operation of the described procedure is presented in Fig.~\ref{fig:m_neuron_scheme}.

\begin{figure}
\centering
    \includegraphics[width=0.5\textwidth]{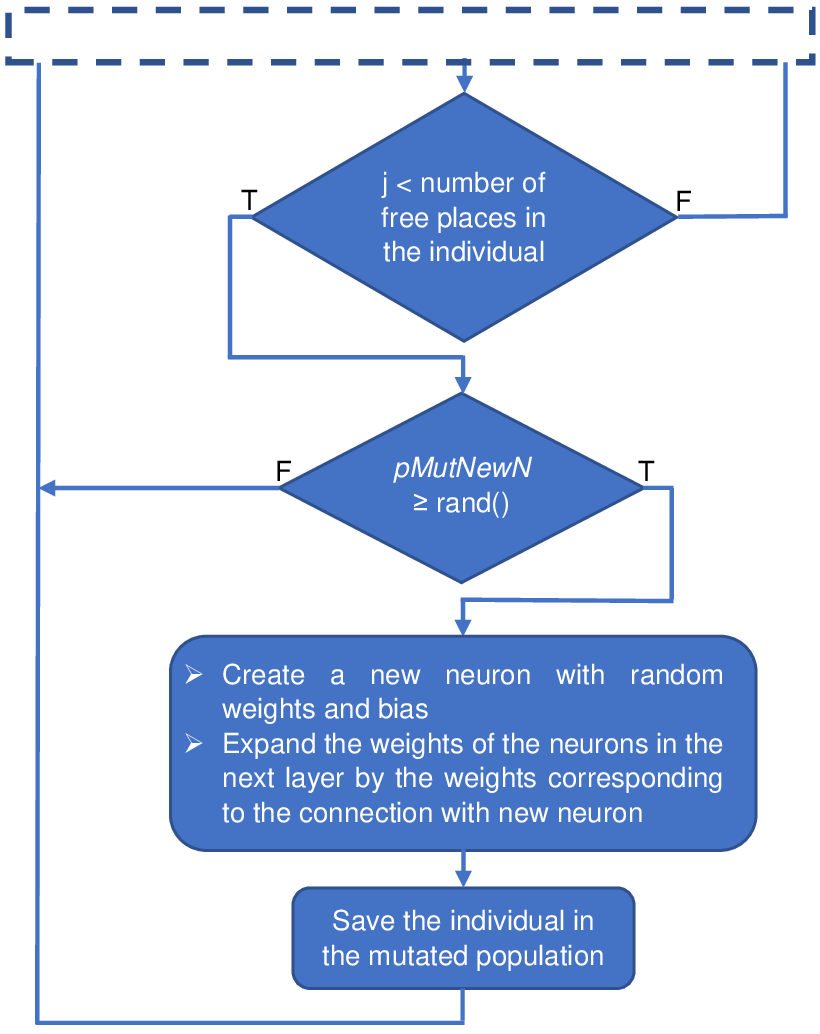} 
    \captionof{figure}{A functional diagram of the neuron mutation operator.}
    \label{fig:m_neuron_scheme}
\end{figure}

\clearpage

In opposition to the above, there is also the natural phenomenon of neuron death. This phenomenon may be compared to the creation of individuals with fewer neurons in a given layer of \acp{ann}. In designed \ac{nas} algorithms, the possibility of neuron death is given to every existing neuron in the hidden layer for each individual in the population with probability equal to $pMutDelN$. Of course in the next layer, the weights corresponding to the removed neuron are also removed. This evolutionary operator works very similar to the procedure illustrated in Fig.~\ref{fig:m_neuron_scheme}.

The last part of the mutation operator in the developed \ac{nas} algorithms is the delays mutation. The mutation of delays consists of randomly increasing or decreasing the delay by one time instance. The fact of change of delay is drawn with a probability equal to $pMutD$ for each type of delays  - $du$ and $dy$ separately. In the case of decreasing the level of delay, the corresponding input with its weights is removed from the network. In turn, in the case of increasing the level of delay, the set of saved inputs is enlarged and the weights corresponding to this input are drawn from the range [$minW$, $maxW$]. The probability of increase and decrease is equal to 50$\%$ for each. Of course, each delay mutation generates a new individual, i.e. a new individual has at most one delay difference than the source individual. A functional diagram of the delays mutation operator is shown in Fig.~\ref{fig:m_delays_scheme}.

\begin{figure}
\centering
    \includegraphics[width=0.8\textwidth]{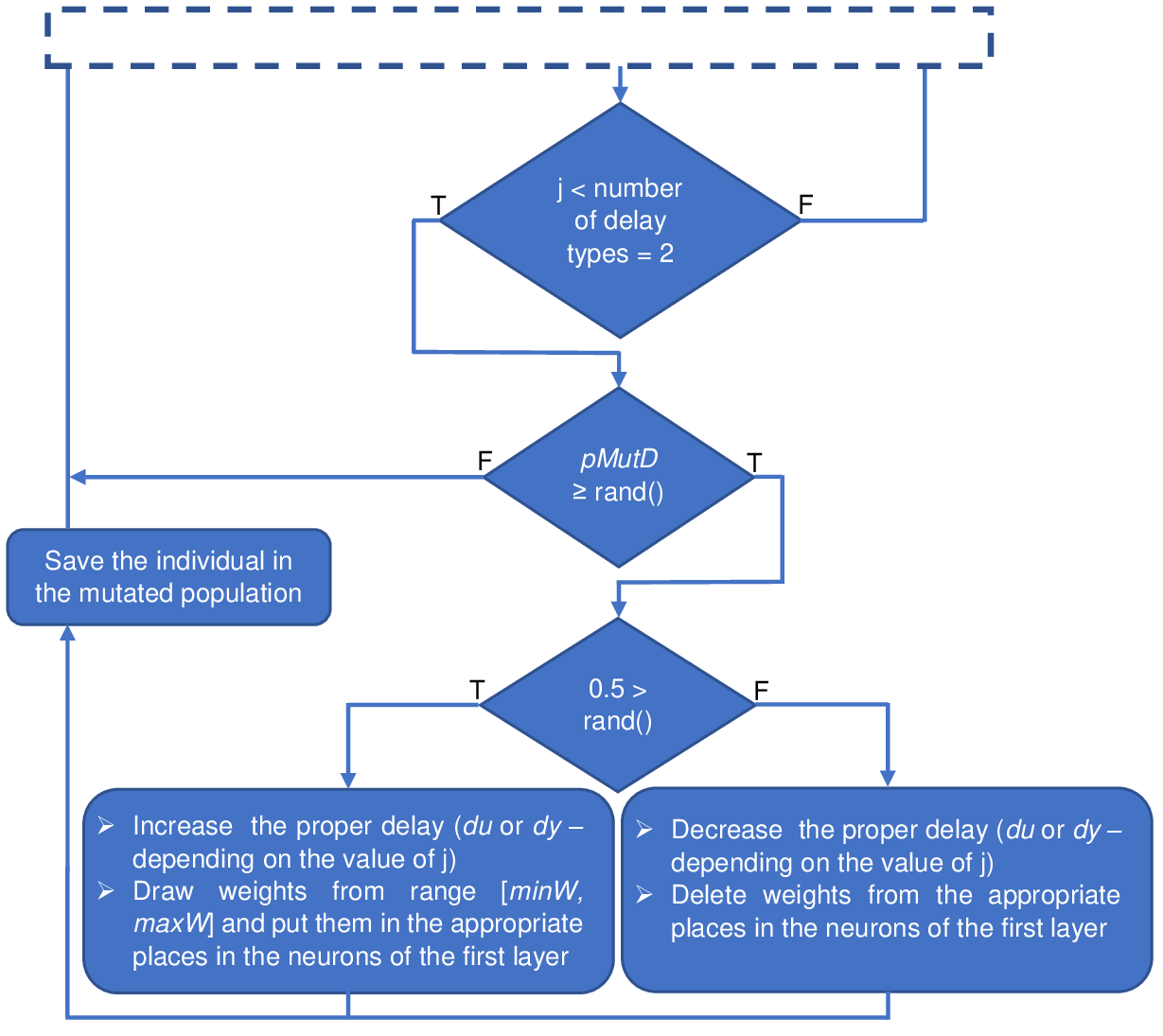} 
    \captionof{figure}{A functional diagram of the delays mutation operator.}
    \label{fig:m_delays_scheme}
\end{figure}

\subsubsection*{Crossover}\label{subsec:crossover}
\begin{figure}
\centering
    \includegraphics[width=0.6\textwidth]{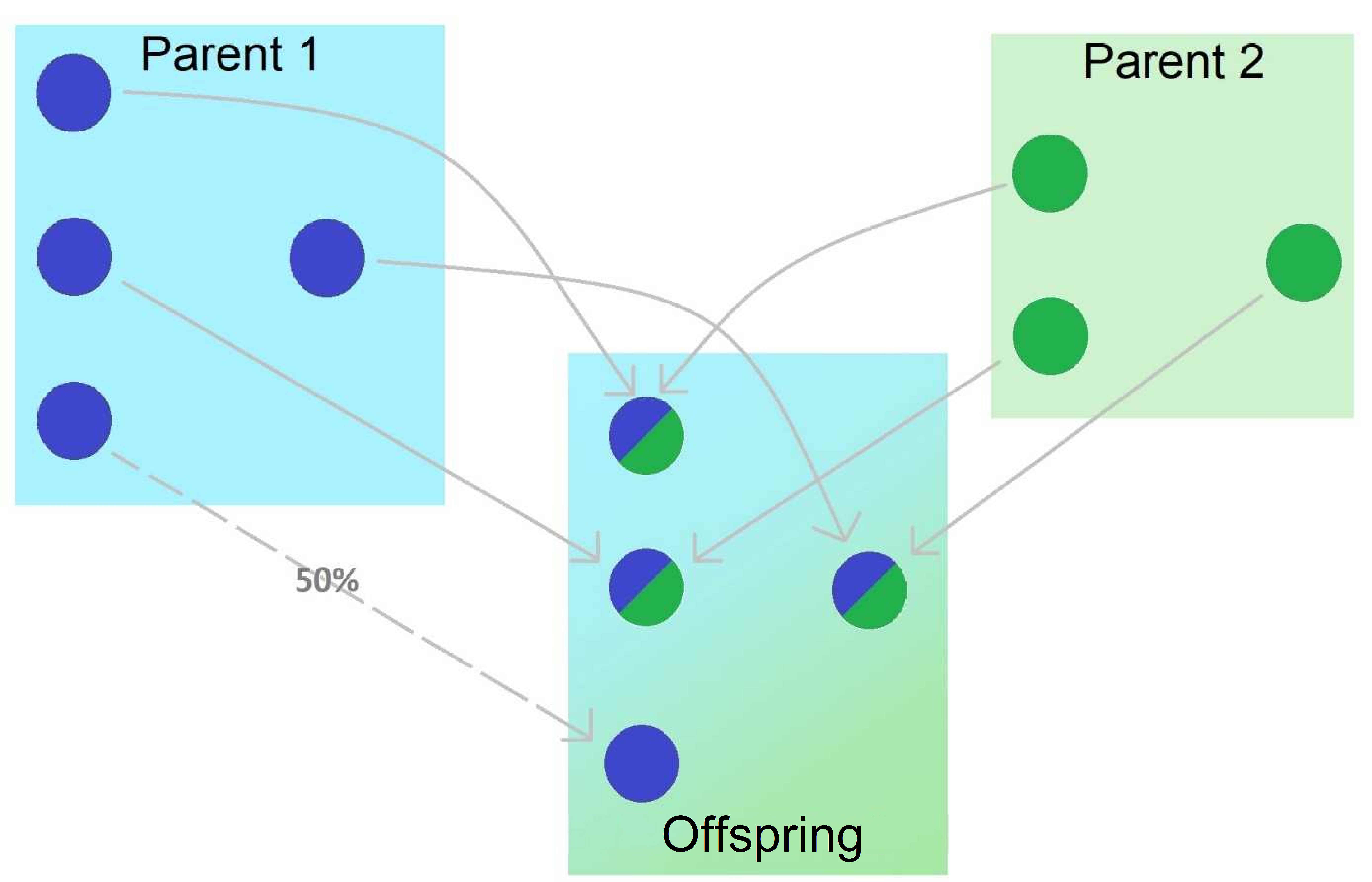} 
    \captionof{figure}{A general diagram of the crossover.}
    \label{fig:crossing_scheme}
\end{figure}

In general, in a short scale of time, an essential part of evolution is a result of crossover (recombination) primarily \cite{Michalewicz1996}. The main advantage of that method of reproduction is the diversity of offspring (children). Among others for this reason, as it has been aforementioned, during operation of \acp{ea} crossover operator is mainly responsible for exploiting the field of solutions. In the developed \ac{nas} algorithms, the crossover operator ensures the exchange of all network's architecture features. A general diagram of the crossover is shown in Fig.~\ref{fig:crossing_scheme}.

In the first step, parents with $pCross$ probability are drawn from a given population. It is assumed that each individual of a given population can be selected on a parent only once in a given generation of the algorithm, and participate in creating only one offspring. Next, the pairs are drawn from the parents' group. Each neuron of \ac{rnn}, which represents parent 1 is crossed with the corresponding neuron of \ac{rnn}, which represents parent 2. However, parents in a pair may differ not only by weight values but also by other \ac{rnn} structure components. This leads to a situation where the corresponding neuron from parent 1 neuron may not exist in the parent 2. This problem is resolved by introducing a 50$\%$ probability of copying to a child a neuron found only in one parent. The copied neuron always occupies the first free place after the last neuron in a given \ac{rnn} layer (see Fig.~\ref{fig:crossing_scheme}). Hence, in the situation when crossover occurs involving parents with different architectures, the described procedure produces a child with a number of neurons that is not greater than the maximum number of neurons in one of the parents. It is also possible to generate offspring with fewer number of neurons that parents have. On the one hand, this is an advantage of the developed operator, because it allows to find the \ac{ann} with the least number of neurons. On the other hand, this action in the initial phase of the algorithm's operation can lead to the elimination of all individuals with larger structures. As a consequence, in the further phase of the algorithm's operation, it may not be possible to create individuals who will adapt to the learning data with satisfactory accuracy. The predominance of individuals with a smaller number of neurons in the initial phase of the algorithm's operation, where due to low weight matching, network’s responses are far from perfect. This results from the smaller penalty part responsible for the size of the network (see \textit{fitness function} in this section). In order to prevent the phenomenon of bad quality of mapping caused by an insufficient number of neurons, an additional evolutionary operator is introduced responsible for adding new neurons to the network -- the neuron mutation operator (see Fig.~\ref{fig:m_neuron_scheme}).

Besides the different numbers of neurons in the layers, individuals differ in the levels of delays, i.e. the number of weights of neurons in the first hidden layer. The devised \ac{nas} algorithms enable that only the weights corresponding to the same inputs are crossed. For example, the weight from parent 1 responsible for the input signal delayed in the second level $u(k-2)$ can be crossed only with the corresponding weight from parent 2, i.e. the weight responsible for input signal delayed in the second level. Hence, in the first step of the crossover operator delays of parents are crossed as follows:
\begin{align}\label{eq:cross_delays}
    du_\mathrm{new} &= round \left(r du_\mathrm{p1} + (1-r) du_\mathrm{p2}\right),\\
    dy_\mathrm{new} &= round \left(r dy_\mathrm{p1} + (1-r) dy_\mathrm{p2}\right),
\end{align}
where:
$du_\mathrm{new}, dy_\mathrm{new}$ denote input delay level and recursive delay level of offspring, respectively;
$du_\mathrm{p1}, du_\mathrm{p2}$ are levels of input delays of parents 1 and 2;
$dy_\mathrm{p1}, dy_\mathrm{p2}$ signifies levels of recursive delays of parents 1 and 2;
$r$ stands for the random number in the range [0,1], drawn separately for each equation. 

Differences in the structures and delays levels of parents' also lead to a different number of weights in the corresponding neurons. If a given weight exists in both parents, the weight in the child is calculated from the following linear combination of parents’ weights \cite{Michalewicz1996}:
\begin{equation}\label{eq:cross_weight}
    w_\mathrm{new} = r w_\mathrm{p1} + (1-r) w_\mathrm{p2},
\end{equation}
where:
$w_\mathrm{new}$ denotes weight of offspring; 
$w_\mathrm{p1}, w_\mathrm{p2}$ are weights of parents 1 and 2.

If the weight corresponding to an input to a neuron exists only in one of the parents, it is rewritten. Because each neuron has a bias, the value of the offspring bias is always calculated according to (\ref{eq:cross_weight}). In order to make the above description more transparent, a scheme of the crossover operator is shown in Fig.~\ref{fig:crossover_operator}.

\clearpage

\begin{figure}
    \begin{subfigure}{0.99\textwidth}
    \centering
    \includegraphics[width=0.95\textwidth, height=0.8\paperheight]{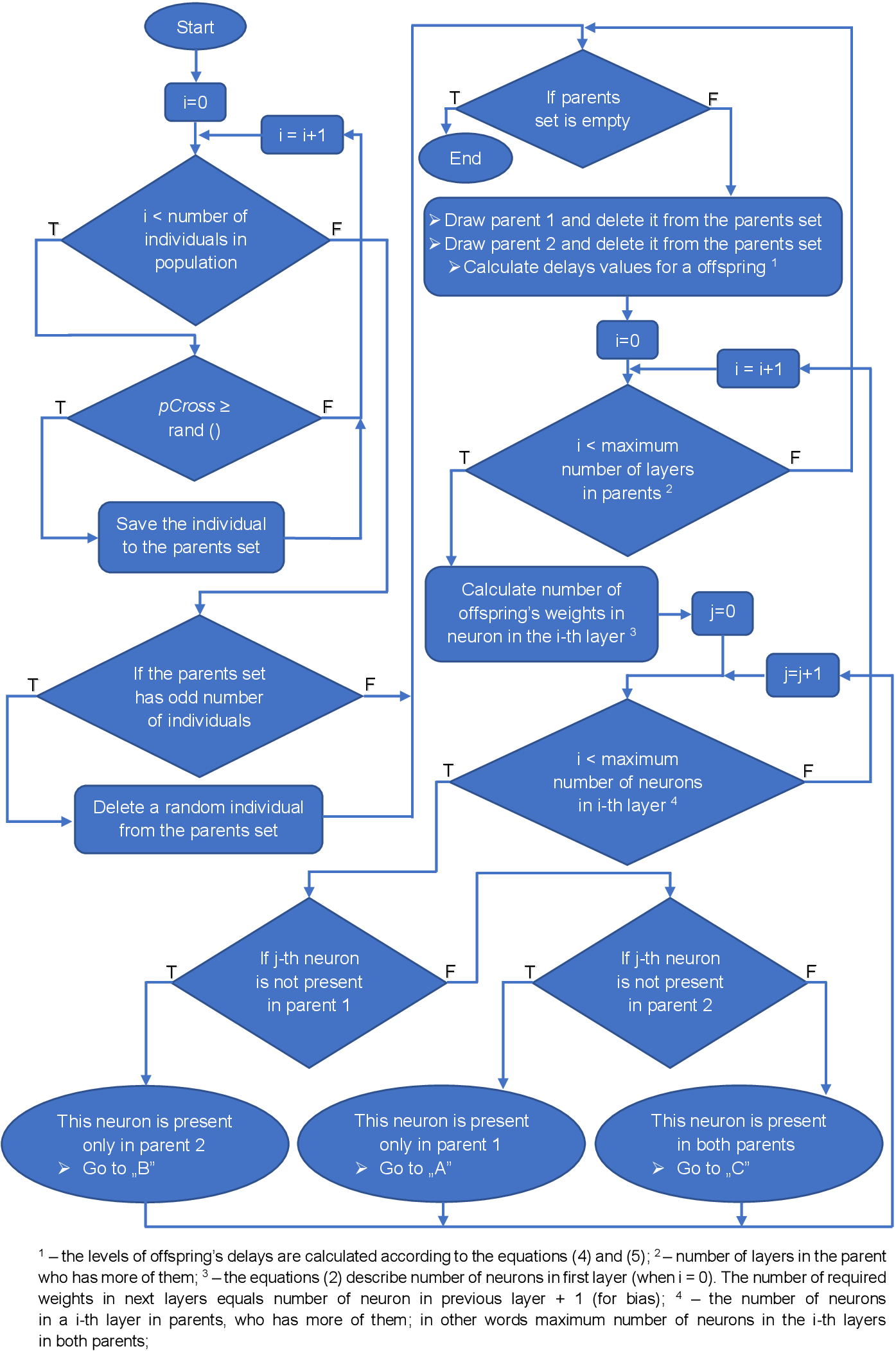}
    \caption{A scheme of crossover operator (continuation, i.e. "go to sections": A, B or C is presented in Fig. 7b)}
    \end{subfigure}
\end{figure}

\begin{figure}
\addtocounter{figure}{-1}
\centering
\renewcommand{\thesubfigure}{b}
    \begin{subfigure}{0.99\textwidth}
    \centering
    \includegraphics[width=0.9\textwidth]{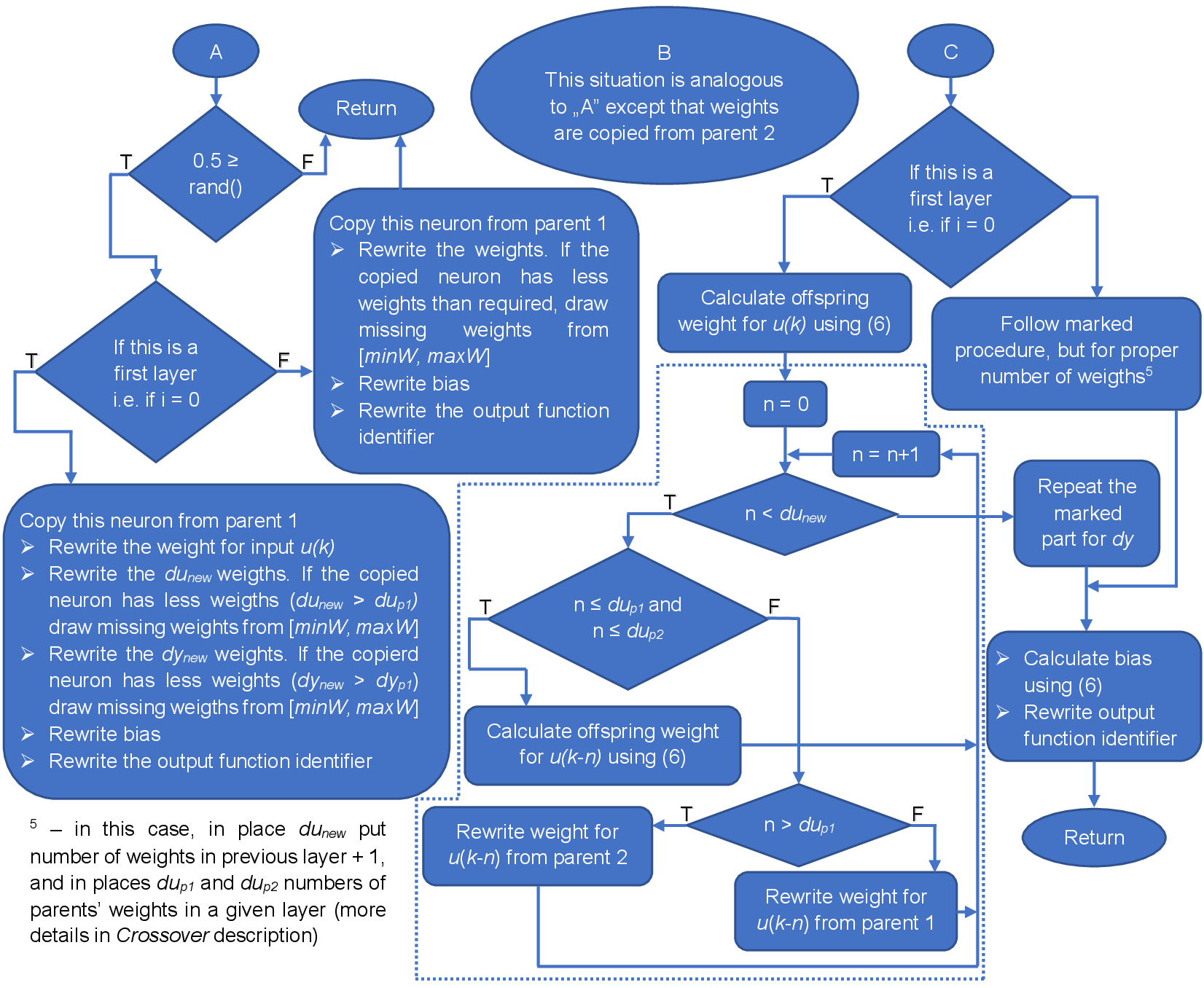}
    \label{fig:crossover_operator_b}
    \caption{A scheme of crossover operator (continuation of the scheme from Fig. 7a)}
    \end{subfigure}
\caption{A scheme of crossover operator.}    
\label{fig:crossover_operator}
\end{figure}
\clearpage
It should be noticed that this way of exchanging information between parents (recombination) is justified. The weight of a given input in parent 1 is crossed only with the weight corresponding to a given input in parent 2 or is completely rewritten. The same operation happens with entire neurons. If both parents have a neuron in a given place of the network, they are crossed with each other. However, if only one parent has a neuron in a given place, then the child inherits it with a probability of 50$\%$, and this neuron is located in the first free place in the given layer. Moreover, thanks to the neuron placement in the first free place in a given layer the transfer of significant neurons, i.e. those with the main positive effect on the network response, to the upper positions of the layers is ensured. This makes it possible to transfer their characteristic features (set of weights) to the neurons of other individuals. This action also saves the memory space used for the algorithm, because the gaps created after neurons are not saved. It is worth adding that copying of whole groups of neurons is not considered. In the \acp{ann} of relatively small size, this is not a disadvantage. However, in the case of creating networks of large sizes, an algorithm should be developed that allows crossing networks based on entire groups of neurons.

\subsubsection*{Fitness function}\label{subsec:fitness_function}

In order to evaluate the fitness of an \ac{ann}, it is necessary to determine its response to the learning data, and then with selected measure check their fit to target response. In the \ac{rnn} the calculation of the first samples of the network’s responses requires the completion of the delayed input and output samples. It depends on the considered problem, and in this work under these values the nominal position of the control rods and the nominal thermal power of the reactor (see section~\ref{sec:introduction}) scaled to the value 1, have been written. The value of the fitness function is calculated as follows:
\begin{equation}\label{eq:fittness_function}
    f_\mathrm{adapt,i} = 10 - p_\mathrm{1} \overline{e}_i - p_\mathrm{2} N_i - p_\mathrm{3} D_i,
\end{equation}
where:
$f_\mathrm{adapt,i}$ is the value of fitness function for $i$th individual;
$p_\mathrm{1}, p_\mathrm{2}, p_\mathrm{3}$ denote the values of weights in the fitness function;
$D_i = du + dy$ stands for the sum of levels of delays occurring in the $i$th individual;
$N_i$ signifies the total number of neurons in the $i$th individual;
$\overline{e}_i$ is the mean error (represents the fit to the target response) for the $i$th individual calculated as:

\begin{equation}\label{eq:error}
    \overline{e}_i = \frac{1}{n} \sum_{k=1}^{n} |y_{i,k} - y_{ref,k} |,
\end{equation}
where:
$n$ is the number of samples in target response;
$y_{i,k}$ denotes subsequent samples of the $i$th individual response;
$y_{ref,k}$ stands for subsequent target response samples.

The value of the fitness function depends not only on the quality of the \ac{rnn} fit but also on its architecture. This fact is included in the penalty part of (\ref{eq:fittness_function}), i.e. fitness function is decreased depending on the number of neurons present in the network and the number of delays in the input and output signals. The values of weights in this part, i.e. $p_\mathrm{1}$, $p_\mathrm{2}$ and $p_\mathrm{3}$ are 1, 0.01 and 0.0001, respectively. They have been chosen so that a network containing one more neuron and the same degree of delays have to generate a response whose average error would be at least 0.01 lower. The value 0.01 results from the specifics of the problem under consideration. Clearly, by dividing the original \ac{pwr} reactor heat output (target response) by its nominal power the response can be understood as the percentage power output. The value of 1 corresponds to the nominal power of the reactor, i.e. 100$\%$ of its load. It has been subjectively assumed that a network with fewer neurons can be wrong by 1$\%$ more of power on each signal sample. Similar reasoning is used to determine the value of $p_\mathrm{3}$, except that a network having one degree of delay less can have 0.01$\%$ greater error on each sample. This value is smaller than for $p_\mathrm{2}$, because the computational cost introduced due to a longer delay is less than the computational cost associated with introduction of additional neuron.

It is worth mentioning that the values of fitness function are evaluated in every developed \ac{nas} algorithm in the same way according to (\ref{eq:fittness_function}). This allows for a direct comparison of the results obtained by the investigated algorithms.

\subsubsection*{Selection}\label{subsec:selection}

It is well-known that in nature the well-fitted individuals, and thus those with a high-value fitness function, live longer. Moreover, they have a greater chance of multiplication, generating at the same time a greater number of offspring who inherit their features. This operator, such as in nature provides a greater likelihood of reproduction for more fitted individuals. It is because these individuals are more likely to survive to the next population, where they will have the chance to be crossed again. The selection operator $\mu + \lambda$ type is used in the devised \ac{nas} algorithms. Clearly, the chance of transition to a new population is given to both individuals formed in a given population and in the previous one, and in detail the roulette method with elitism is used \cite{Michalewicz1996}.

\subsubsection{\ac{dnas1} algorithm} \label{subsec:dnas1}
The domain of searched \ac{rnn} architecture is limited by setting the $du$ and $dy$ delays of all individuals in a given population to a constant value (see table \ref{tab:summary}) of $5$ in the \ac{dnas1} algorithm. Moreover, in order to enable the building of \acp{rnn} of larger sizes and ensuring better fit to learning data the neuron mutation-deleting operator is not used. The selection operator consists of the roulette method as well as elitism. The $hmBest$ number of the best individuals is directly transferred to the next population. In turn, the remaining individuals, i.e. $popSize-hmBest$ are drawn using the roulette method.
\subsubsection{\ac{dnas2} algorithm} \label{subsec:dnas2}

In the \ac{dnas2} algorithm the domain of searched \ac{rnn} architecture also includes adjustment of the $du$ and $dy$ values (see table \ref{tab:summary}). These values are changed in the mutation and crossover operators. Other operating conditions of \ac{dnas2} algorithm are identical to \ac{dnas1} algorithm.
\subsubsection{\ac{dnas3} algorithm} \label{subsec:dnas3}

In order to limit random events (e.g. drawing of missing weights during the crossover) and reasonable evaluation time of the \ac{dnas2} algorithm, subsequent mechanisms inspired by nature have been introduced. These include biogeographical zones, ecological divergence and competitive exclusion \cite{Futuyma2018}. 
Algorithm \ac{dnas2} developed in this way is the third \ac{nas} algorithm, so-called \ac{dnas3} algorithm. An exemplary effect of using the above mechanisms is the crossover of individuals with similar features (species). Thus, features that improve the fitness of an individual are transferred more often. The consequence of this is that individual features in a population are isolated and strengthened, and dominate over others. Hence, considering the problem of missing weights, the \acp{rnn} with a different number of neurons in the hidden layer are treated as separate species. This approach described in \cite{Bishop2006,Kirkpatrick1983}, utilise the segregation of individuals of the population concerning the similarity of their structure. Therefore, a crossover operator can operate only on individuals (networks) belonging to one species and only within it (intra-species crossover). Such operation is aimed at faster convergence to the optimal \ac{rnn} weights for a given structure. However, it cannot be guaranteed that in a given population, e.g., the initial one, there will be at least one individual of the optimal species, i.e. a network with the appropriate number of neurons in the hidden layer. Also, in order to avoid the evolution operations for species that differ significantly from the currently optimal species, the evolution operators have been provided, that is responsible for creating species slightly different from the original. It enables reaching from one species (e.g. not optimal) to another even significantly different (e.g. optimal) through numerous small changes, which is consistent with the theory of gradualism \cite{Futuyma2018}.

\begin{figure}
    \centering
    \includegraphics[width=0.75\textwidth]{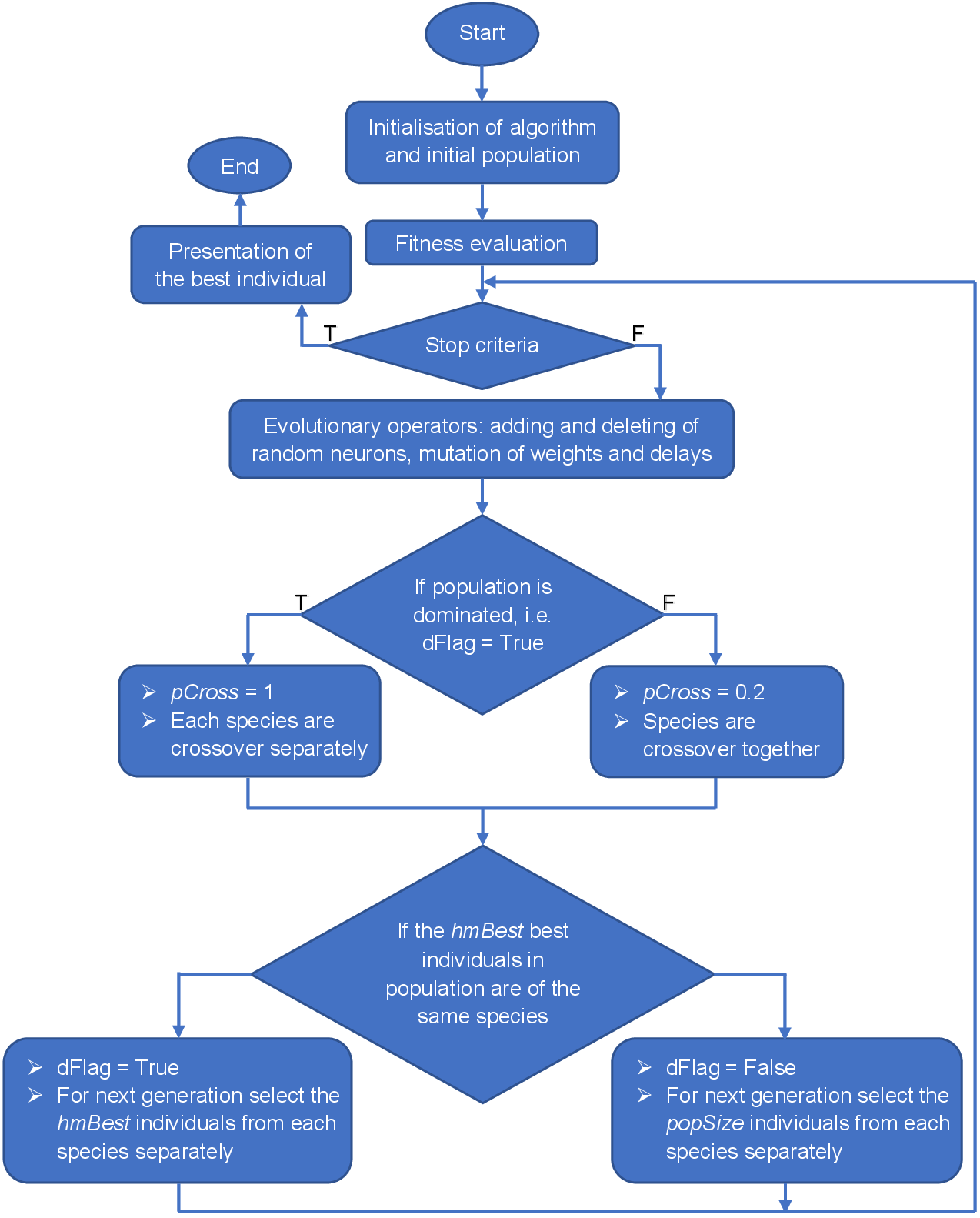} 
    \caption{A general scheme of the designed \ac{dnas3} algorithm.}
    \label{fig:dnas3_scheme}
\end{figure}

The mechanisms presented above have been implemented in \ac{dnas3} algorithm in the following way. The population is initialised with $popSize$ number of individuals, allowing all of them to crossover with all of the rest. When a given population is dominated by one species, only the $hmBest$ number of the best individuals from the dominant species, and the same number of individuals from two secondary species are be allowed to develop further through the intra-species crossover. The secondary species have one neuron more or less than the dominant species in the hidden layer. The intra-species crossover means that the offspring can be formed only from parents belonging to one species. At the same time, due to the small number of individuals of the dominant species and of the secondary species, the probability of crossover $pCross$ rate increases to 1. The used crossover operator does not differ from the one used in \ac{dnas2} algorithm (see section \ref{subsec:dnas1dnas3}). A given species becomes dominant if the $hmBest$ number of the most fit individuals in the population belongs to it. If a species loses domination in the population, the inter-species crossover is restored, $pCross$ is then reduced to 0.2, and the population's size is increased to a maximum number expressed by $popSize$, giving the chance to dominate the population by the new species. The creation of new species has been implemented using the described function of creating new neurons and deleting already existing neurons (see section \ref{subsec:dnas1dnas3} - the neuron mutation operator). Keeping in mind that, new species can occur as a result of the crossover. In comparison to \ac{dnas1} and \ac{dnas2} algorithms, the roulette method is not used to speed up the convergence of the algorithm. The next population receives a certain number of best-fitness individuals ("full elitism") depending on whether the population is dominated or not. In the case of a non-dominated population, this is the maximum number ($popSize$) of best-fitness individuals. Because the number of all existing individuals in a given population may be smaller, then it is possible that all individuals will pass to the next population. In the situation of dominance to the next population, as it has been mentioned above, passes $hmBest$ number of individuals of each species separately, i.e. $hmBest$ number of individuals of the dominant species and $hmBest$ number of each of the two secondary ones, in total $3hmBest$ of individuals. A general scheme of the designed \ac{dnas3} algorithm is presented in Fig.~\ref{fig:dnas3_scheme}.
\subsection{\ac{dnas4} algorithm} \label{subsec:dnas4}

The idea of the \ac{dnas4} algorithm is different from the algorithms presented previously. It is because in \ac{dnas4} algorithm the classic gradient descent methods are used to change the weights, whereas the \ac{ea} is used to select the number of hidden layers, the number of neurons in every layer, levels of delays $du$ and $dy$, and a type of gradient descent algorithms. The direct motivation to develop this approach is related to the concluded tests of \ac{dnas1} -- \ac{dnas3} algorithms. In details, it has been observed during the tests that the optimal or suboptimal \ac{rnn} structure is mainly determined in the initial phases of activity of the \ac{nas} algorithm. Whereas the subsequent \ac{rnn} improvement through the mutation of weights takes a large number of generations. Taking this into account, the \ac{dnas4} algorithm using the common gradient learning methods has been developed.

As with \ac{dnas1} -- \ac{dnas3} algorithms, the general scheme of \ac{ea} in \ac{dnas4} algorithm is in line with Fig.~\ref{fig:EA_scheme}. However, the selection, mutation and crossover operators have been changed compared to \ac{dnas1} -- \ac{dnas3} algorithms. Moreover, retraining is used in this algorithm. The retraining is repeated learning, i.e. reselection of \ac{rnn} weights of a given individual with other initial values of weights. Nevertheless, this operation does not always ensure the improvement of fitness of a given individual. Therefore, in every generation, only selected individuals are given to retraining. These individuals are randomly selected with the $pRetrain$ probability. The well-known gradient methods have been chosen for learning purposes. These are Levenberg-Marquardt algorithm \cite{Fukuda1992,Ponulak2011}, backward propagation of Bayesian regularisation \cite{Chua1988} and backward propagation of scaled gradient \cite{Stanley2002}. The $\mu + \lambda$ selection is applied again, and the maximum $popSize$ of the best-fitness individuals (elitism) is selected. Hence, the selection operator has been modified in such a way that only the best individual with a given architecture, i.e. only the best individual of a given species is transferred to the next population. The individuals belong to the same species when they have the same number of layers, neurons in each layer, $du$, $dy$ and method of training. Thanks to this, the variable population size has been obtained, which reduces the number of calculations and accelerates the result's achievement.

As it has been aforementioned, the mutation in \ac{dnas4} algorithm differs from this operator in the algorithms presented previously. It is because the \ac{rnn} weights are not mutated. In turn, the structure and the rest of the \ac{rnn} parameters represented by each of the individuals from the population has a chance to mutate with a probability of $pMut$. If an individual is mutated, it is equally likely that the number of hidden layers, the number of neurons in each layer, both levels of delays or the learning method will be mutated. Nevertheless, only one feature for a given individual is changed during one generation of the algorithm. The mutation of the number of hidden layers involves a random removal of a layer or inserting a new layer with a random number of neurons in a random place of the network. In turn, the neurons' mutation involves the deletion or the addition of one neuron to each layer. A random draw is done separately for each individual’s layer, and neuron's addition or removal, and is equally likely to 50$\%$ chance of addition or deletion. In the situation, when the last neuron is removed from a layer, the entire layer is deleted. The mutation of levels of delays $du$ and $dy$ is the equally likely to increase or decrease for each level by one. Whereas the mutation of the learning method is based on the equally probable selection from among all available methods, except that which has been already assigned to the individual at the time of the mutation.

The crossover operator consists of the exchange of individuals’ features concerning the \ac{rnn} structure, and learning method. In contrast to \ac{dnas1} -- \ac{dnas3} algorithms, weights are not crossed, which results in the required training of new individuals. In the first step, similarly to \ac{dnas1} -- \ac{dnas3} algorithms, the pairs of parents are randomly selected. Next, the following features of offspring are calculated:
\begin{itemize}
    \item the number of hidden layers as:
    \begin{equation}\label{eg:number_hidden_layers_dnas4}
        L_\mathrm{new} = round\left(rL_\mathrm{p1}+(1-r)L_\mathrm{p2}\right),
    \end{equation}
    where:
    $L_\mathrm{new}$ is the number of offspring hidden layers;
    $L_\mathrm{p1}, L_\mathrm{p2}$ denote the number of hidden layers in parent 1 and 2, respectively;
    \item the number of neurons in the $i$th hidden layer of offspring as:
    \begin{equation}\label{eg:number_neurons_hidden_layers_dnas4}
        N_{i,\mathrm{new}} = round\left(rN_{i,\mathrm{p1}}+(1-r)N_{i,\mathrm{p2}}\right),
    \end{equation}
    where:
    $N_{i,\mathrm{new}}$ is the number of neurons in $i$th offspring hidden layer;
    $N_{i,\mathrm{p1}}, N_{i,\mathrm{p2}}$ denote the number of neurons in $i$th hidden layers in parent 1 and 2, respectively;\\
    It should be noticed that if a given layer occurs only in one parent, then the number of neurons is calculated on the assumption that the number of neurons corresponding to the second parent equals zero.
    \item the level of the $i$th delay as:
    \begin{equation}\label{eg:level_delay_dnas4}
        d_{i,\mathrm{new}} = round\left(rd_{i,\mathrm{p1}}+(1-r)d_{i,\mathrm{p2}}\right),
    \end{equation}
    where:
    $d_{i,\mathrm{new}}$ is the level of the $i$th delay for offspring;
    $d_{i,\mathrm{p1}}, d_{i,\mathrm{p2}}$ denote the levels of $i$th delays in parent 1 and 2, respectively;
    \item the offspring inherits learning method from one of the parents, where each parent has a 50$\%$ chance to transfer his method.
\end{itemize}

\section{Case study}\label{sec:case_study}

In general, the \ac{dnas1} -- \ac{dnas4} algorithms have been simulation-verified in an analogous way. However, for the \ac{dnas1} -- \ac{dnas3} algorithms, $100$ calls of a particular algorithm have been made and each of them counted $100$ generations. In turn, the \ac{dnas4} algorithm has been called $46$ times with $25$ generations. This was due to the long computation time of this algorithm. As it has been mentioned above, the aim has been to find an \ac{rnn} with an optimal (minimum) architecture while at the same time obtaining satisfactory accuracy of its response. Such a network is a desirable \ac{siso} black-box model of the fast processes in \ac{pwr}. The values of the particular input parameters of devised \ac{nas} algorithms are shown in table~\ref{tab:input_parameters_values}.
\begin{table}
   \caption{The values of the input parameters for \ac{dnas1} -- \ac{dnas4} algorithms.}
   \label{tab:input_parameters_values}
   \begin{tabular}{|c|c|c|c|c|c|}
  \hline
  \multicolumn{6}{|c|}{\textbf{Values of input parameters}} \\ \hline \hline
\multicolumn{1}{|c|}{\multirow{2}{*}{\textbf{Symbol}}} & \multicolumn{4}{c|}{\textbf{Value}} & \multirow{2}{*}{\textbf{Comment}} \\ \cline{2-5}
\multicolumn{1}{|c|}{} & \textbf{\ac{dnas1}} & \textbf{\ac{dnas2}} & \textbf{\ac{dnas3}} & \textbf{\ac{dnas4}} & \\ \hline
$maxLay$ & 1 & 1 & 1 & 1 & - \\ \hline
$maxNinLay$ & 20 & 20 & 20 & 20 & \begin{tabular}[c]{@{}c@{}}this value has been chosen experimentally \\ and it depends on the correlation between \\ the input and output (target) data; during \\ numerical experiments, it has been observed \\ that if the smaller correlation appears \\ then the bigger number of neurons \\ is required (see also \cite{Kavzoglu1999})\end{tabular} \\ \hline
$du$ & 5 & ($1,duMax$) & ($1,duMax$) & ($1,duMax$) & \begin{tabular}[c]{@{}l@{}}the value $duMax$ has been chosen \\ experimentally and it is equal to 50\end{tabular} \\ \hline
$dy$ & 5 & ($1,dyMax$) & ($1,dyMax$) & ($1,dyMax$) & \begin{tabular}[c]{@{}l@{}}the value $dyMax$ has been chosen \\ experimentally and it is equal to 50\end{tabular} \\ \hline
$popSize$ & 50 & 50 & 50 & 50 & \begin{tabular}[c]{@{}c@{}}in \ac{dnas3} and \ac{dnas4} algorithms the size \\ of the population are changing during their \\ work\end{tabular} \\ \hline
$pCross$ & 0.8 & 0.8 & 0.8 & 0.8 & - \\ \hline
$p_i$ & \begin{tabular}[c]{@{}c@{}}1, \\ 0.01, \\ 0.0001\end{tabular} & \begin{tabular}[c]{@{}c@{}}1, \\ 0.01, \\ 0.0001\end{tabular} & \begin{tabular}[c]{@{}c@{}}1, \\ 0.01, \\ 0.0001\end{tabular} & \begin{tabular}[c]{@{}c@{}}1, \\ 0.01, \\ 0.0001\end{tabular} & see section \ref{subsec:dnas1dnas3} \\ \hline
$minDelta$ & 0.0001 & 0.0001 & 0.0001 & 0.0001 & - \\ \hline
$maxDelta$ & 0.1 & 0.1 & 0.1 & 0.1 & - \\ \hline
$pMutW$ & 0.2 & 0.2 & 0.2 & 0.2 & - \\ \hline
$pMut$ & 0.2 & 0.2 & 0.2 & 0.2 & - \\ \hline
$pMutNewN$ & 0.2 & 0.2 & 0.2 & 0.2 & - \\ \hline
$pMutD$ & - & 0.2 & 0.2 & - & - \\ \hline
$pMutDelN$ & - & - & 0.2 & - & - \\ \hline
$minW$ & -1 & -1 & -1 & - & (*) \\ \hline
$minW$ & 1 & 1 & 1 & - & (*) \\ \hline
$hmBest$ & 5 & 5 & 5 & 5 & - \\ \hline
$pRetrain$ & - & - & - & 0.2 & - \\ \hline
\multicolumn{6}{|c|}{\begin{tabular}[c]{@{}c@{}}(*) the values [-1,1] of the initial weights [$minW$, $maxW$] are because: \\ a) this range is bipolar and symmetrical with respect to zero, \\ therefore ensures that negative values for the activation function are accounted for; \\ b) the range of the input signals to the designed \ac{ann} is [-2.196, 0] \\ whereas the range of the target output signal is [0, 1.184] (see section \ref{subsec:pwr}),\\ thus the weights in considered interval should ensure calculating (re-scaling) input signal to the desirable levels \\ of the output signal; \end{tabular}} \\ \hline
\end{tabular}
\end{table}
\subsection{Framework} \label{subsec:framework}

In order to solve a given task using the \acp{ea}, the task can be either transferred to a proper form for the algorithm or the algorithm can be adapted to suit the task \cite{Michalewicz1996, Ahmadizar2015}. The classic genetic algorithms are associated with the first approach. In turn, the use of evolutionary algorithms allows for omitting \ac{ann}'s architecture encoding and decoding for operators, which can speed up \ac{ea} operation \cite{Michalewicz1996, Ahmadizar2015}. However, the adaptation of evolutionary operators to the \ac{nas} problem requires a proper way of saving data. In other words, a proper manner of coding solutions. It is because the efficiency of a given algorithm depends largely on this operation \cite{Michalewicz1996}.

\begin{figure}
\centering
    \includegraphics[width=0.8\textwidth]{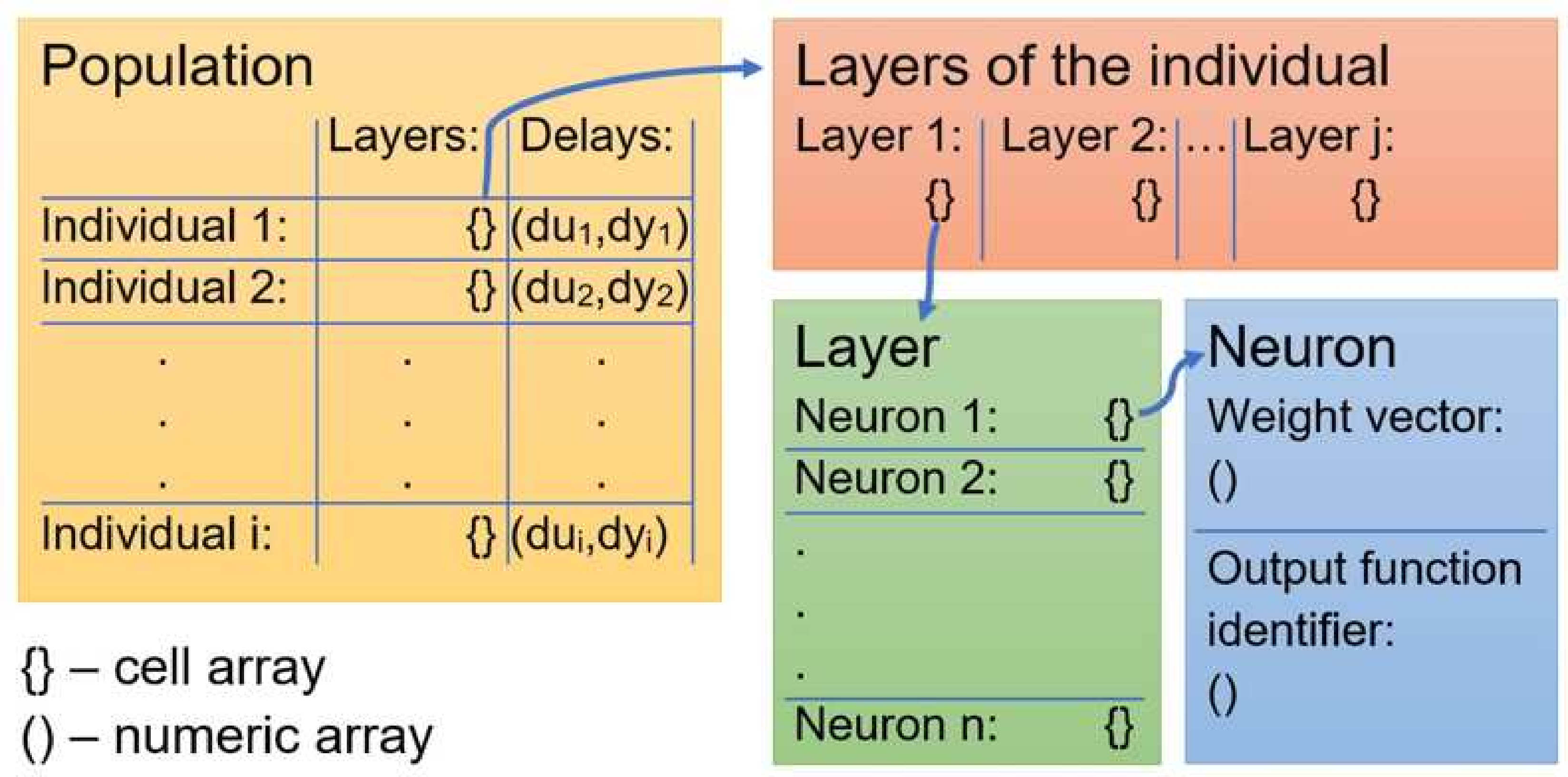} 
    \captionof{figure}{The way of saving of a given \ac{rnn} architecture in the devised algorithms.}
    \label{fig:RNN_saving}
\end{figure}

The devised algorithms have been implemented in the Matlab environment. All variables and constants in these algorithms have been saved in the decimal floating-point format. Whereas the structure of \acp{rnn} is saved in the cell arrays. These are a data type with indexed data containers called "cells". Each cell can contain every data type, including other cell or cell array. This feature is used to repeatedly nest subsequent cell arrays. The authors are aware that this way of saving the structure of \acp{rnn} does not ensure the fastest possible operation of algorithms. However, this has been done because the additional aim of the implementation has been to make the code easy to interpret by other users. Moreover, this approach allows relatively easy transfer of prepared software to open source environments, e.g., Python. As it has been aforementioned, in developed algorithms populations consist of individuals representing the architectures of \acp{rnn}. In turn, the \acp{rnn} are composed of layers, and the layers consist of neurons. Each neuron consists of a set of weights (with bias) and an identifier of the activation function. Therefore, a table of cells is available in which under the next indexes individuals of populations are placed. Each individual consists of the cell array, wherein each next array corresponds to the next level of penetration of the network, and a numeric array, in which two non-negative integers are placed determining the level of inputs and recursive outputs delays. The first array corresponds to the whole network and contains as many elements as the given network has layers. Each layer is built from neurons, so each array, which corresponds to the given layer, is built from next cell arrays, which correspond to neurons. Each neuron array consists of two numerical arrays. In the first numerical array are determined the neuron weights and bias whereas in the second a numeric value takes place, which identifies the activation function of a given neuron. The above description is illustrated in Fig. \ref{fig:RNN_saving} to ensure its transparency.

\subsection{PWR} \label{subsec:pwr}

\begin{figure}
\centering
\begin{minipage}{.45\textwidth}
    \includegraphics[width=0.99\textwidth]{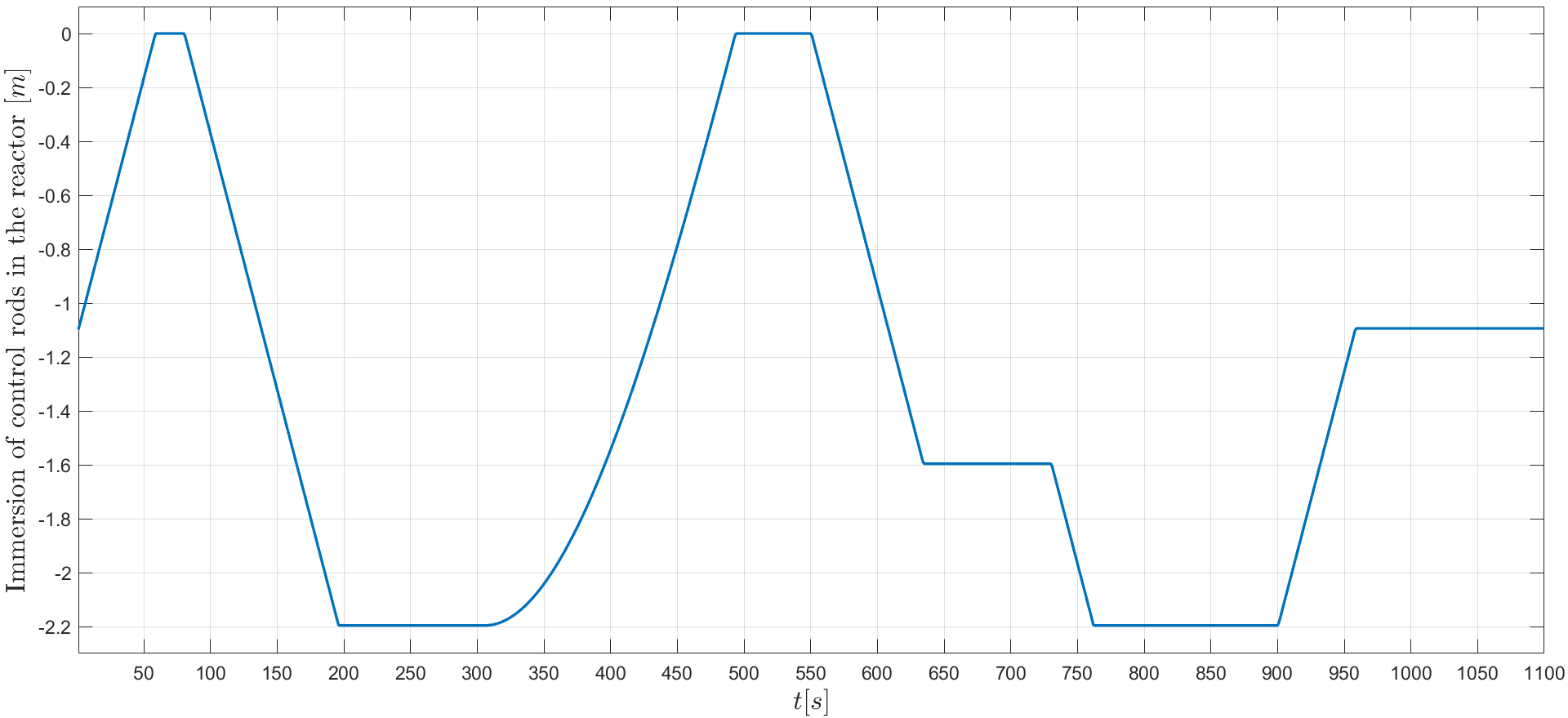} 
    \captionof{figure}{The target position \\\hspace{\textwidth} of the control rods in a \ac{pwr}.}
    \label{fig:ref_control_rods}
\end{minipage}
\begin{minipage}{.45\textwidth}
    \includegraphics[width=0.99\textwidth]{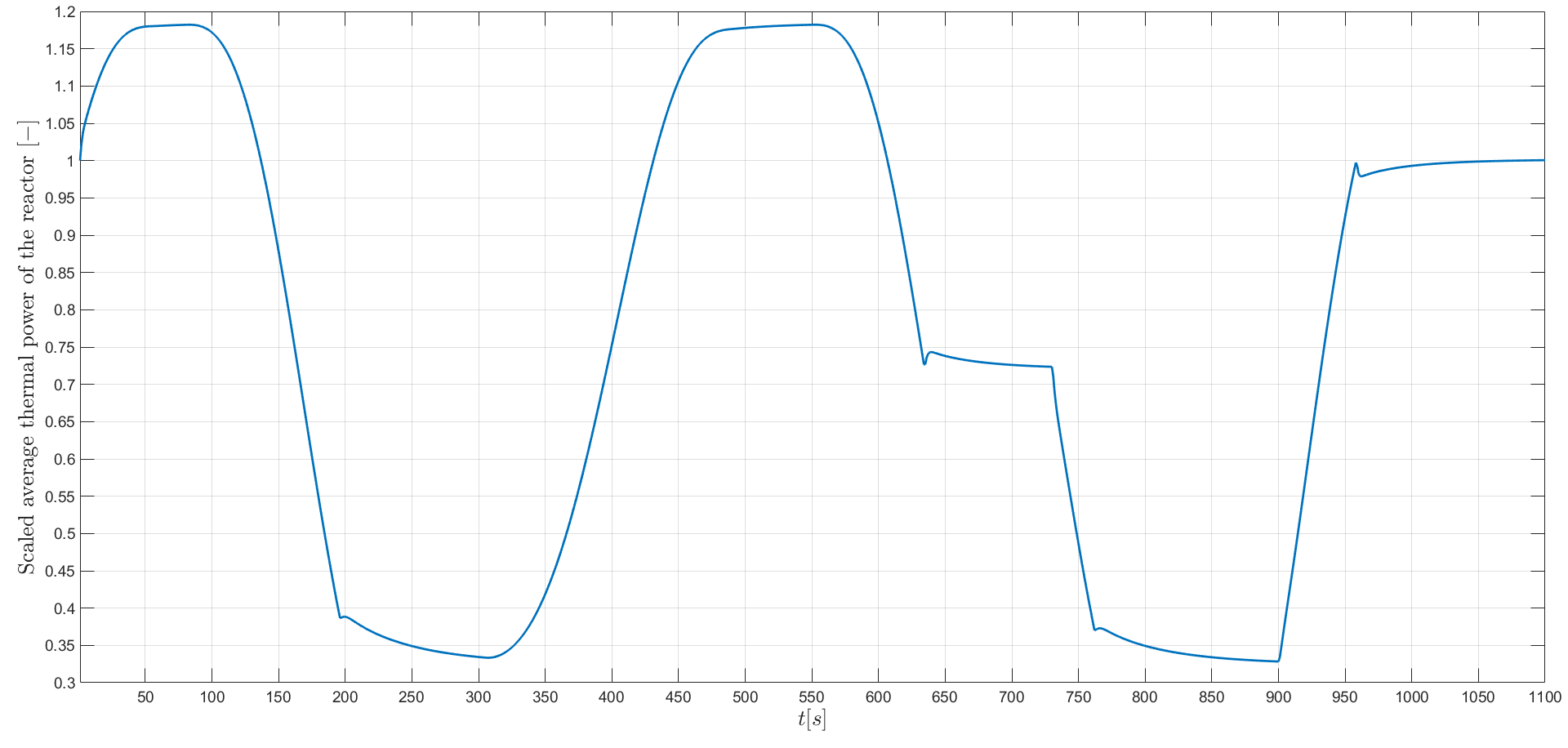} 
    \captionof{figure}{The target trajectory\\\hspace{\textwidth} of thermal power in a \ac{pwr}.}
    \label{fig:ref_thermal_power}
\end{minipage}
\end{figure}

In accordance with the considerations presented before, the \ac{dnas1} -- \ac{dnas4} algorithms have been developed for creating the \ac{siso} black-box model of the fast processes in \ac{pwr}. Therefore, the position of the control rods and the thermal power of the \ac{pwr} are the selected input and output signals, respectively. They are presented in Figs.~\ref{fig:ref_control_rods} and \ref{fig:ref_thermal_power} \cite{Puchalski2018,Puchalski2020}. It should be added that the trajectory in Fig.~\ref{fig:ref_thermal_power} is scaled. Clearly, the physical sense of this reference response is the average thermal power of the \ac{pwr} scaled by dividing it by the nominal average thermal power of the reactor. This means that the value 1 corresponds to the nominal thermal power (see table~\ref{tab:pwr_parameters_values}).

\begin{table}
   \caption{The values of \ac{pwr} parameters.}
   \label{tab:pwr_parameters_values}
   \begin{tabular}{|c|c|c|}
   \hline
    \multicolumn{3}{|c|}{\textbf{Values of \ac{pwr} parameters}} \\ \hline \hline
    \textbf{Name} & \textbf{Value} & \textbf{Unit} \\ \hline
The minimal location of the control rods & -2.196 & $\mathrm{m}$ \\ \hline
The nominal location of the control rods & -1.098 & $\mathrm{m}$ \\ \hline
The maximal location of the control rods & 0 & $\mathrm{m}$ \\ \hline
The nominal thermal power of the reactor & 3 436 & $\mathrm{MW}$ \\ \hline
\end{tabular}
\end{table}
\subsection{Results} \label{subsec:results}
In this section, the simulation results illustrating the performance of the proposed algorithms are presented. First, the learning (training) phase using trajectories from Figs.~\ref{fig:ref_control_rods} and \ref{fig:ref_thermal_power} is shown. Next, the verification phase is discussed. Moreover, the results obtained by the proposed algorithms are presented against the results generated by the NARX-based exhaustive search algorithm from Matlab \cite{Narx:2021} and the basic \ac{neat} algorithm implemented in Python \cite{Stanley2002,Neat:2021}. During the performed experiment with NARX-based exhaustive search algorithm the following parameters were set: number of hidden layers $= 1$, number of neurons in hidden layers $= max 25$, $du = max 50$, $dy = max 50$, learning method - Levenberg-Marquardt algorithm. The experiment was repeated 10 times. Whereas, during the performed experiment with the basic \ac{neat} algorithm, its task  was simplified because the penalty part dependent on the delay levels ($du$ and $dy$) was not considered in the objective function. This was since the \ac{neat} algorithm can create feedbacks of different types \cite{Stanley2002,Neat:2021}. The stop condition was taken as 100 generations and algorithm was called 100 times. 

\subsubsection{Learning phase} \label{subsubsec:learning_phase}
\begin{figure}
\centering
   \begin{subfigure}[t!]{0.45\textwidth}
    \includegraphics[width=\textwidth]{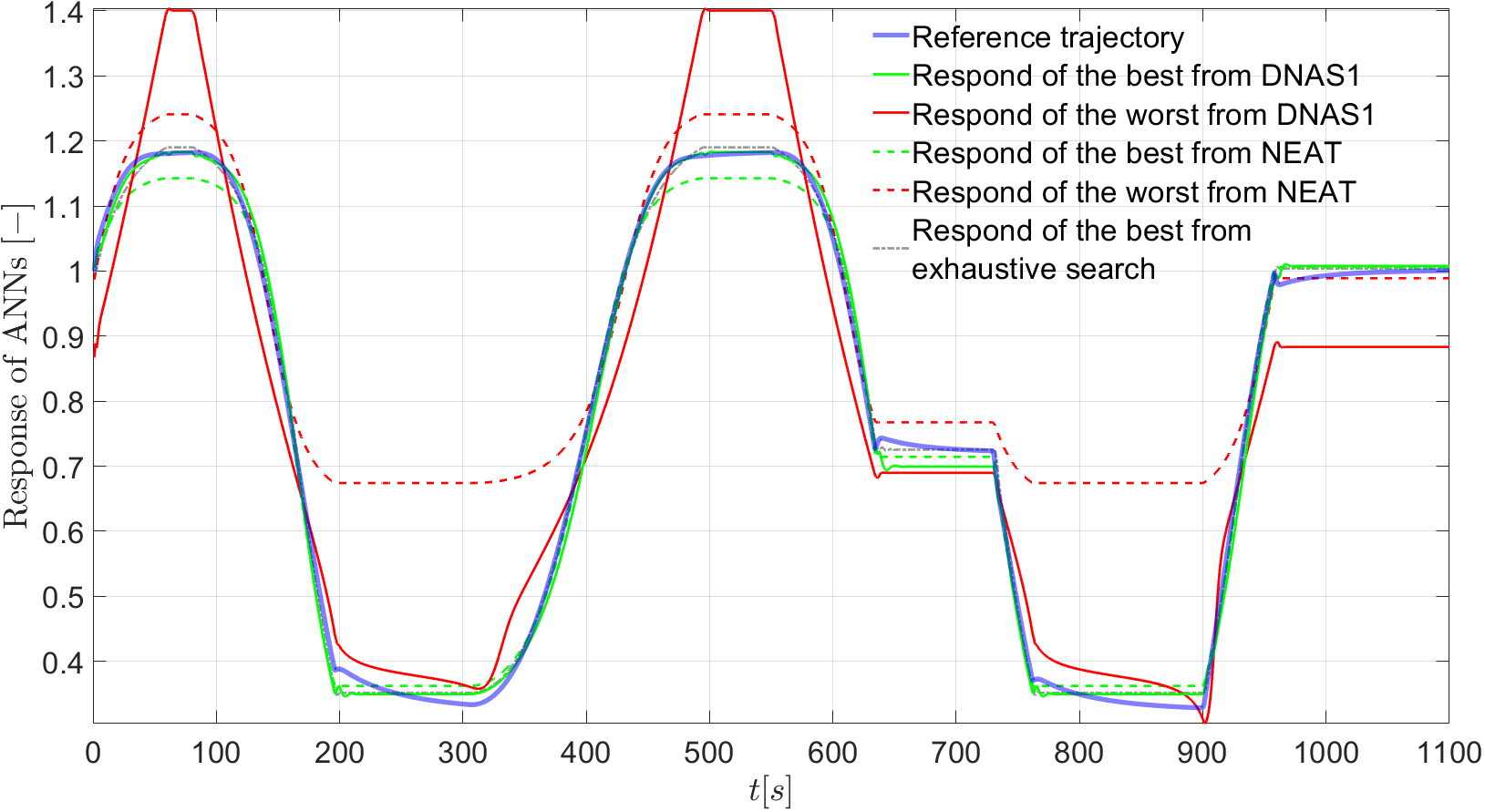} 
    \caption{The average thermal power from \ac{dnas1} algorithm} \label{fig:tp_dnas1}
  \end{subfigure}
  \hskip 2em
  \begin{subfigure}[t!]{0.45\textwidth}
    \includegraphics[width=0.99\textwidth]{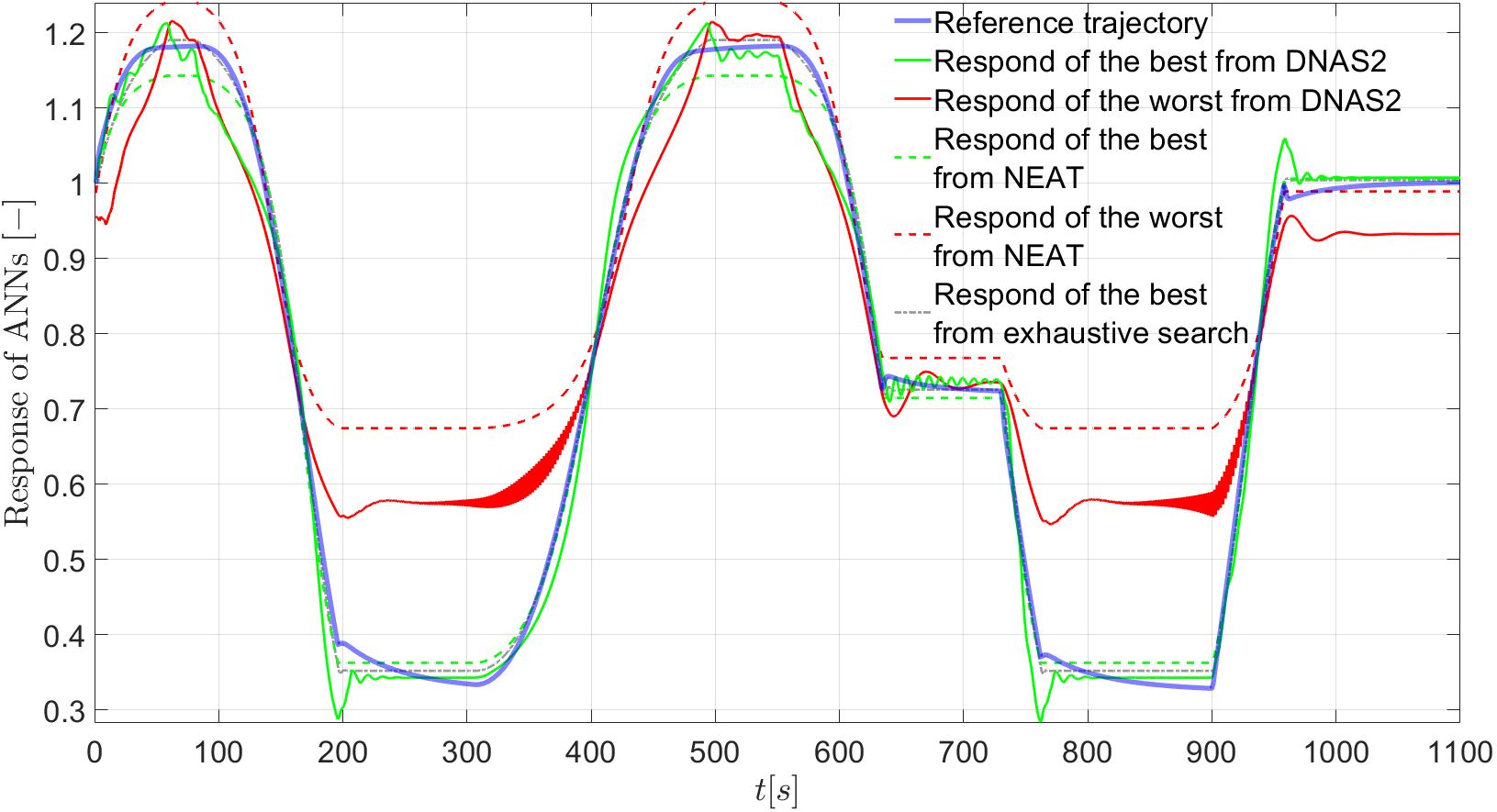} 
    \caption{The average thermal power from \ac{dnas2} algorithm} \label{fig:tp_dnas2}
  \end{subfigure}
  \vskip 2em

  \begin{subfigure}[t!]{0.45\textwidth}
    \includegraphics[width=0.99\textwidth]{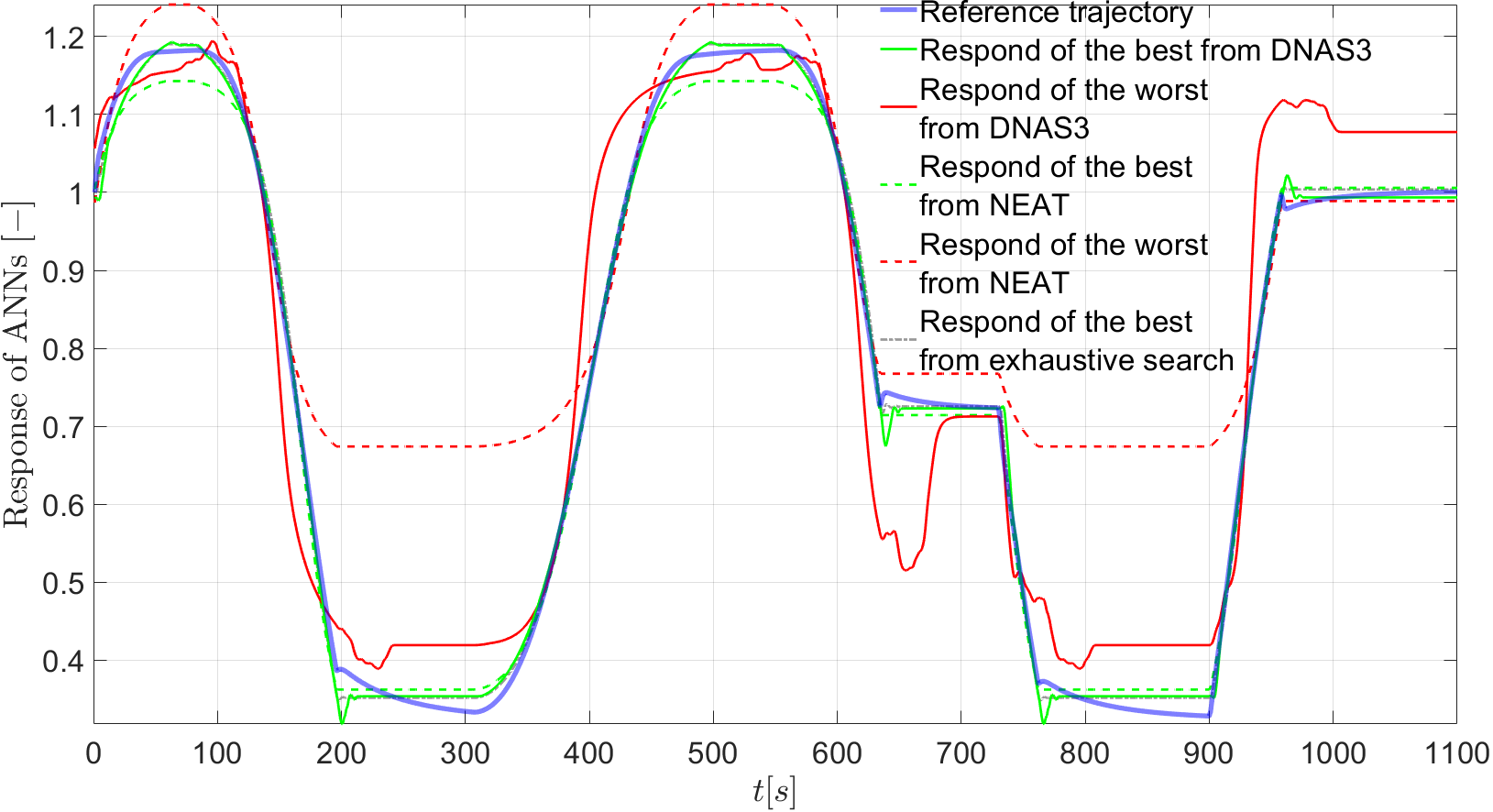} 
     \caption{The average thermal power from \ac{dnas3} algorithm} \label{fig:tp_dnas3}
  \end{subfigure}
  \hskip 2em
  \begin{subfigure}[t!]{0.45\textwidth}
    \includegraphics[width=0.99\textwidth]{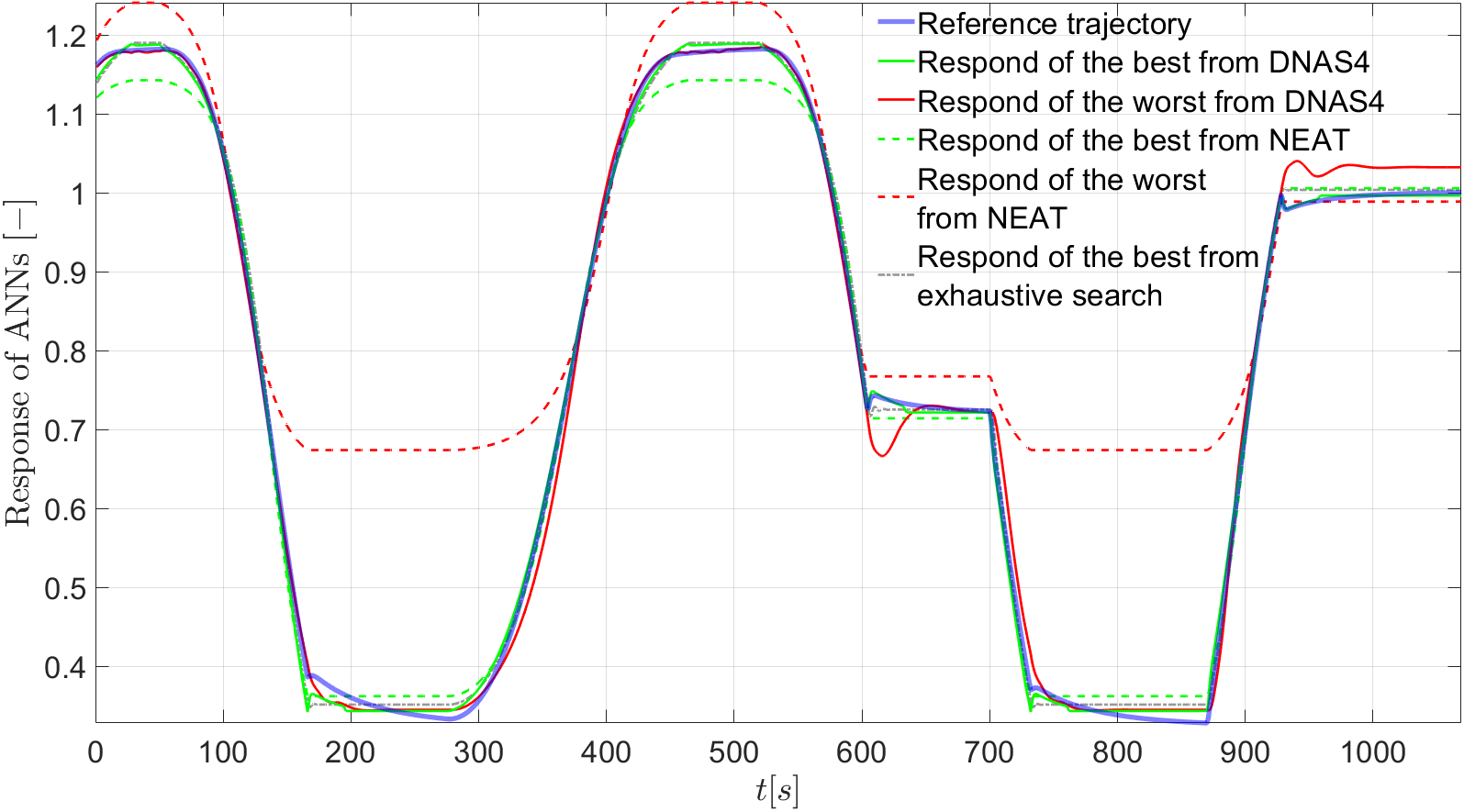} 
     \caption{The average thermal power from \ac{dnas4} algorithm} \label{fig:tp_dnas4}
  \end{subfigure}
  \vskip 1em
\caption{The generated trajectories of the average thermal power of the reactor in the learning phase.}\label{fig:thermal_power_learning_phase}
\end{figure}

\begin{figure}
\centering
    \includegraphics[width=0.9\textwidth]{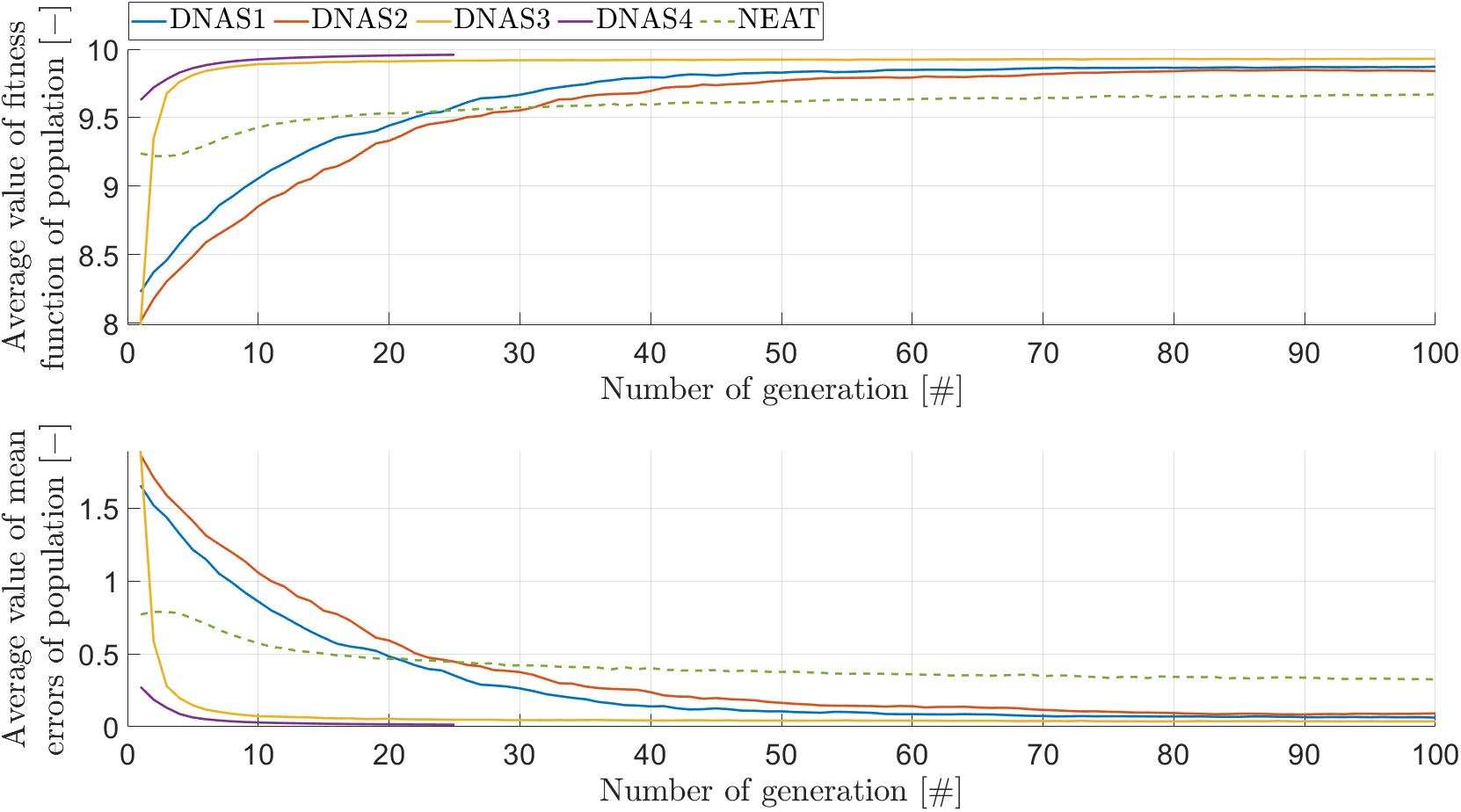} 
    \caption{The trajectories of the average value of the fitness functions and the mean errors.}
    \label{fig:ff_err_learning_phase}
\end{figure}

\begin{figure}
    \begin{subfigure}[t!]{0.45\textwidth}
    \includegraphics[width=0.99\textwidth]{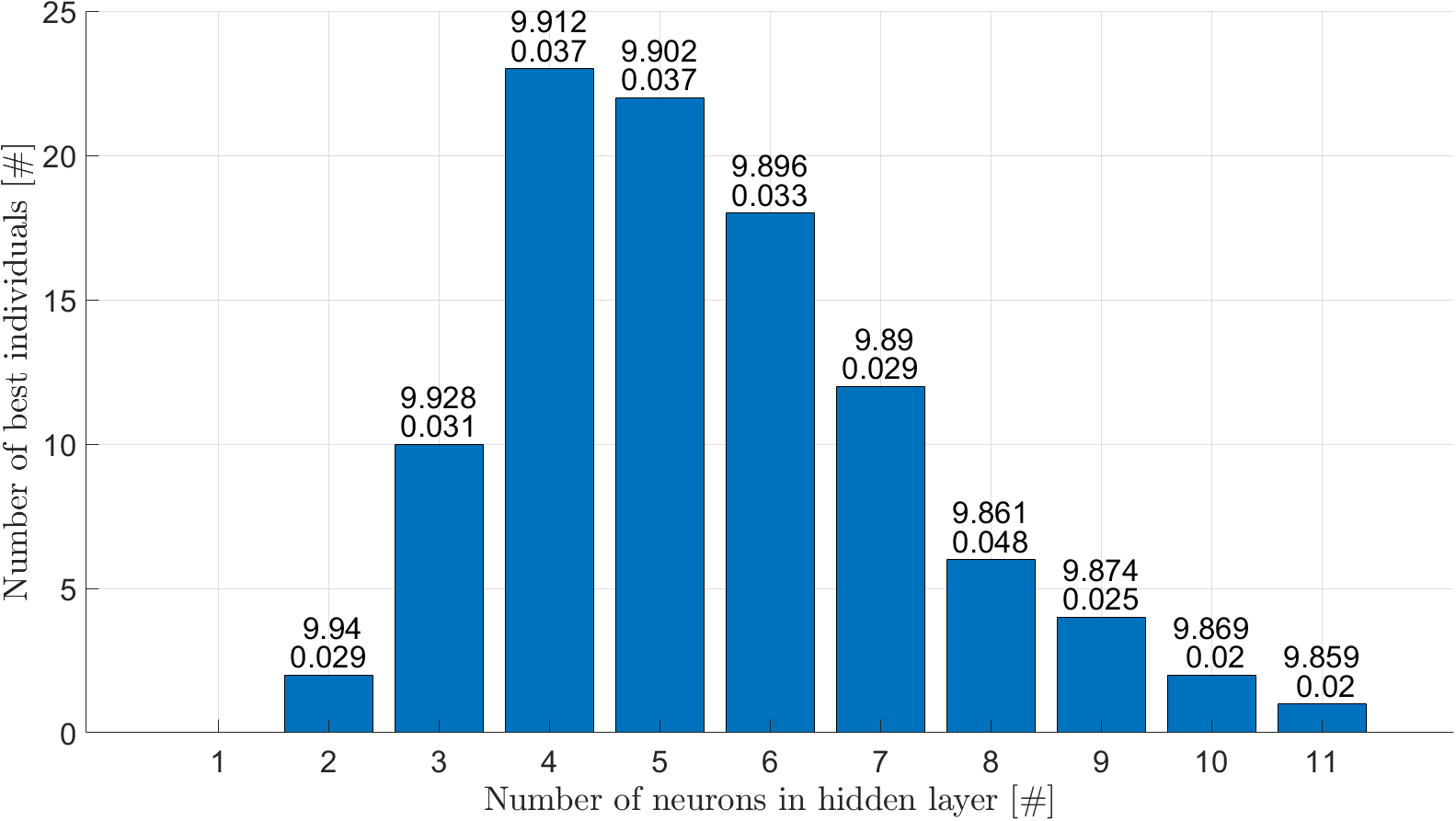} 
    \caption{The distributions of the number of best individuals in \ac{dnas1} algorithm} \label{fig:nn_dnas1}
  \end{subfigure}
  \hskip 2em
  \begin{subfigure}[t!]{0.45\textwidth}
   \includegraphics[width=0.99\textwidth]{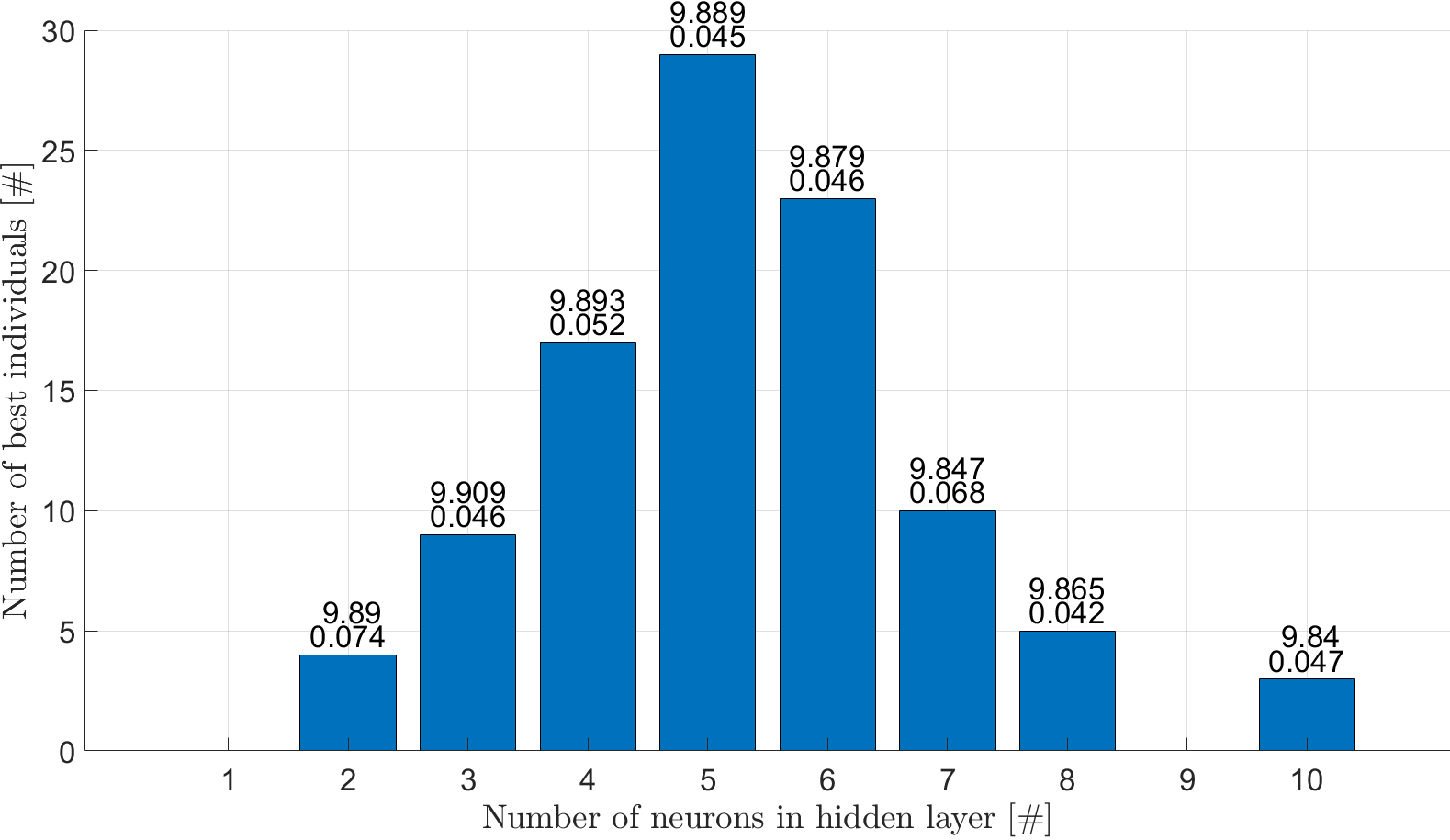} 
    \caption{The distributions of the number of best individuals in \ac{dnas2} algorithm} \label{fig:nn_dnas2}
  \end{subfigure}
  \vskip 2em

  \begin{subfigure}[t!]{0.45\textwidth}
    \includegraphics[width=0.99\textwidth]{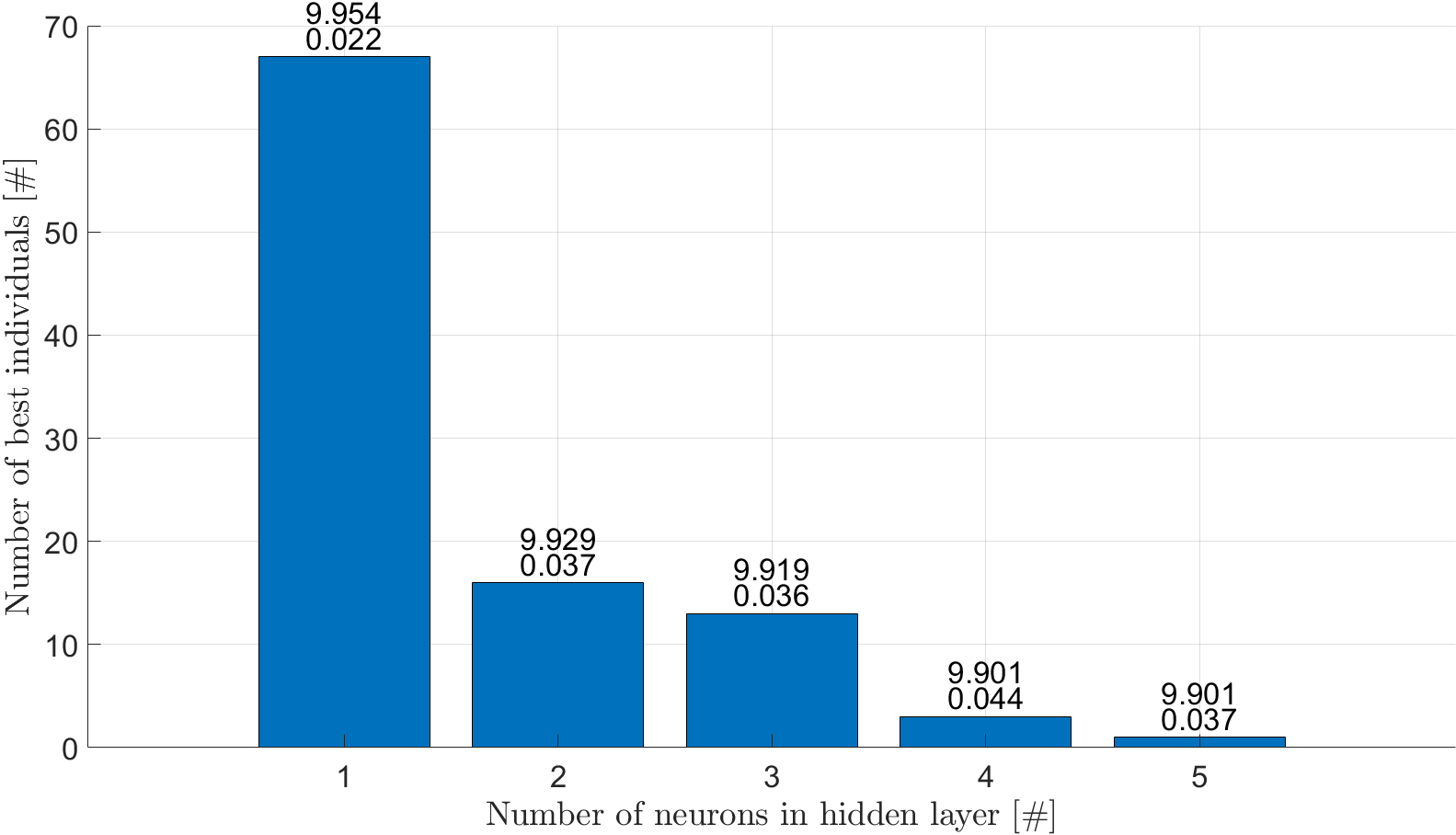} 
    \caption{The distributions of the number of best individuals in \ac{dnas3} algorithm} \label{fig:nn_dnas3}
  \end{subfigure}
  \hskip 2em
  \begin{subfigure}[t!]{0.45\textwidth}
    \includegraphics[width=0.99\textwidth]{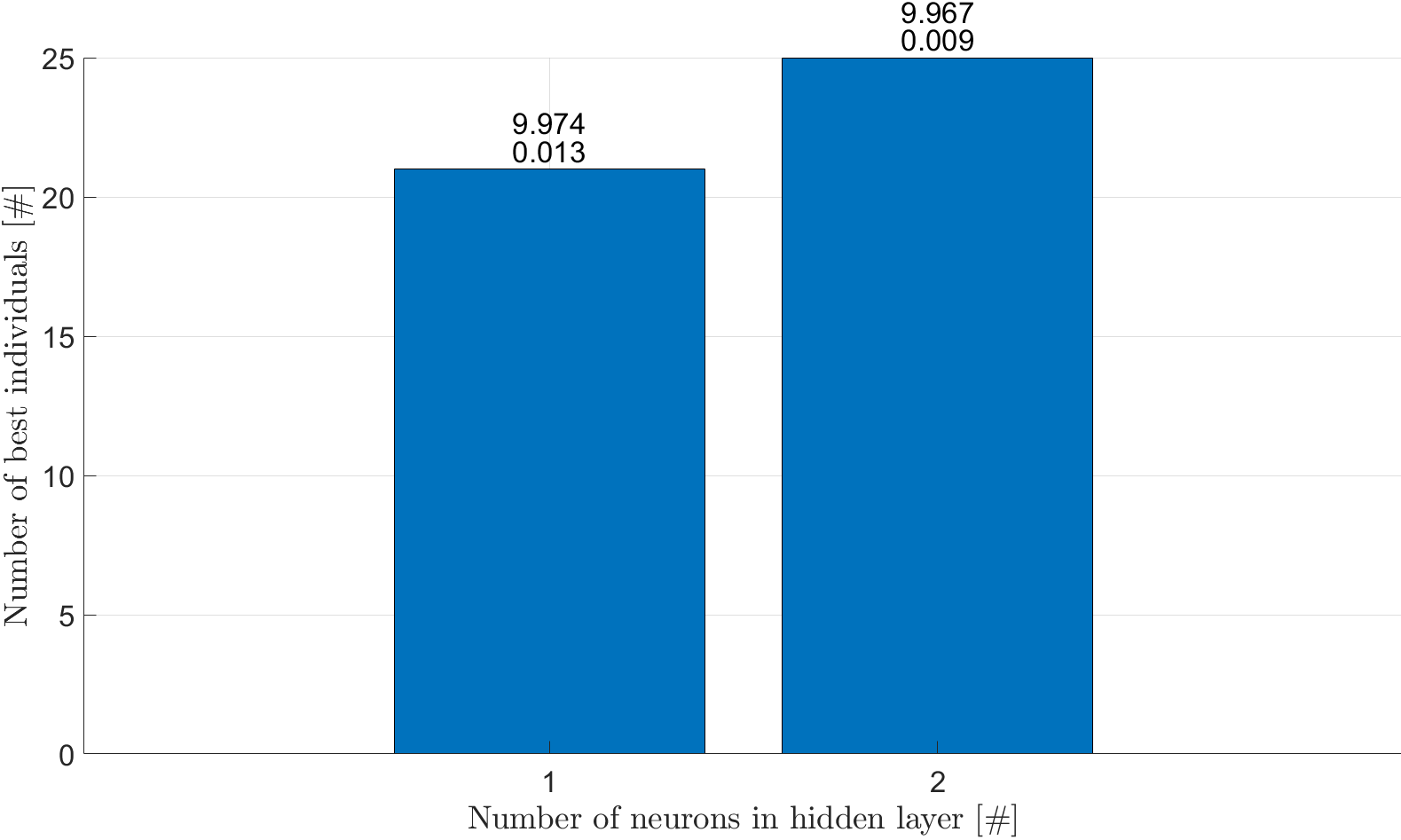} 
    \caption{The distributions of the number of best individuals in \ac{dnas4} algorithm} \label{fig:nn_dnas4}
  \end{subfigure}

  \hskip 1em
  \caption{The distributions of the number of best individuals in the learning phase.}
\label{fig:neuron_number_learning_phase}
\end{figure}

The simulation results showing the generated trajectories of the average thermal power of the reactor by the best and worst individual from the set of best individuals (relative to the value of the fitness function) for the \ac{dnas1} -- \ac{dnas4} algorithms are shown in Fig.~\ref{fig:thermal_power_learning_phase}. Whereas the trajectories representing the average value of the fitness functions of the whole population and the average value of the mean errors are illustrated in Fig.~\ref{fig:ff_err_learning_phase}. In turn, the distributions of the number of best individuals with a given number of neurons in the hidden layer (with a given \ac{rnn} structure), obtained during calls of the particular algorithms are presented in Fig.~\ref{fig:neuron_number_learning_phase}. It should be noticed that over each bar, there are two numbers. The first of them specify the average value of the fitness function of individuals whereas the second denotes the mean error calculated according to \eqref{eq:error}. Thus, for example for \ac{dnas1} algorithm, the highest number of best individuals ($23$) possess $4$ neurons in the hidden layer.

\clearpage

Analysing Figs.~\ref{fig:thermal_power_learning_phase} --  \ref{fig:neuron_number_learning_phase} it can be noticed that the best performance of the generated responses by the devised \acp{rnn} while having the smallest network architecture is provided by the \ac{dnas3} and \ac{dnas4} algorithms. Clearly, the obtained \acp{rnn} from \ac{dnas3} and \ac{dnas4} algorithms seem to be promising candidates for the black-box model of the fast processes in a \ac{pwr}. Hence, the optimality which is understood as achieving the desired trade-off between the size of obtained \ac{rnn} and the accuracy of black-box model responses may be ensured. In order to further illustrate this, several indicators are summarised in table~\ref{tab:indicators_values}. Taking into account the values of indicators from table~\ref{tab:indicators_values} it can be observed that the performance of the generated response by the \ac{dnas2} algorithm is worse than the results obtained with the \ac{dnas1} algorithm while significantly increasing the computation time. This is due to the incomparably larger space of searched solutions in the \ac{dnas2} algorithm. This problem is eliminated in \ac{dnas3} algorithm, where the search space is the same as in \ac{dnas2} algorithm. However, the efficiency of this algorithm as measured by the computation time as well as the accuracy of the response it generates is significantly higher. Moreover, the size of the provided network is much smaller than in \ac{dnas1} and \ac{dnas2} algorithms. As it can be noticed above completely different is \ac{dnas4} algorithm, which gets its advantage in the performance of \ac{rnn} response through the use of gradient selection of weights, but it pays off with the longest calculation time. Of course, the final test for the developed algorithms is the verification phase, where the input data have not been used during the learning phase. The results obtained are presented in the next section.

\begin{table}
\caption{The values of indicators for \ac{dnas1} -- \ac{dnas4} algorithms.}
   \label{tab:indicators_values}
      \begin{tabular}{|c|c|c|c|c|}
   \hline
    \multicolumn{5}{|c|}{\textbf{Values of indicators}} \\ \hline \hline
    \textbf{Indicator} & \textbf{\ac{dnas1}} & \textbf{\ac{dnas2}} & \textbf{\ac{dnas3}} & \textbf{\ac{dnas4}} \\ \hline
The average value of mean errors & 0.0343 & 0.0500 & 0.0270 & 0.0108 \\ \hline
The average value of the number of neurons in the hidden layer & 5.43 & 5.26 & 1.55 & 1.54 \\ \hline
The average value of the $du$ & $const$ = 5 & 25.21 & 24.10 & 25.35 \\ \hline
The average value of the $dy$ & $const$ = 5 & 23.43 & 20.41 & 11.00 \\ \hline
The average value of the computational time for a hundred generations & 37 $\mathrm{min}$ & 68 $\mathrm{min}$ & 18 $\mathrm{min}$ & 912 $\mathrm{min}$* \\ \hline
\multicolumn{5}{|c|}{\begin{tabular}[c]{@{}c@{}} * the duration of calculations for $100$ generations have been calculated from \\ the proportion using the registered calculation times for $25$ generations \end{tabular}} \\ \hline
\end{tabular}
\end{table}

\subsubsection{Verification phase} \label{subsubsec:verification_phase}
\begin{figure}
    \begin{subfigure}[t!]{0.45\textwidth}
    \includegraphics[width=0.99\textwidth]{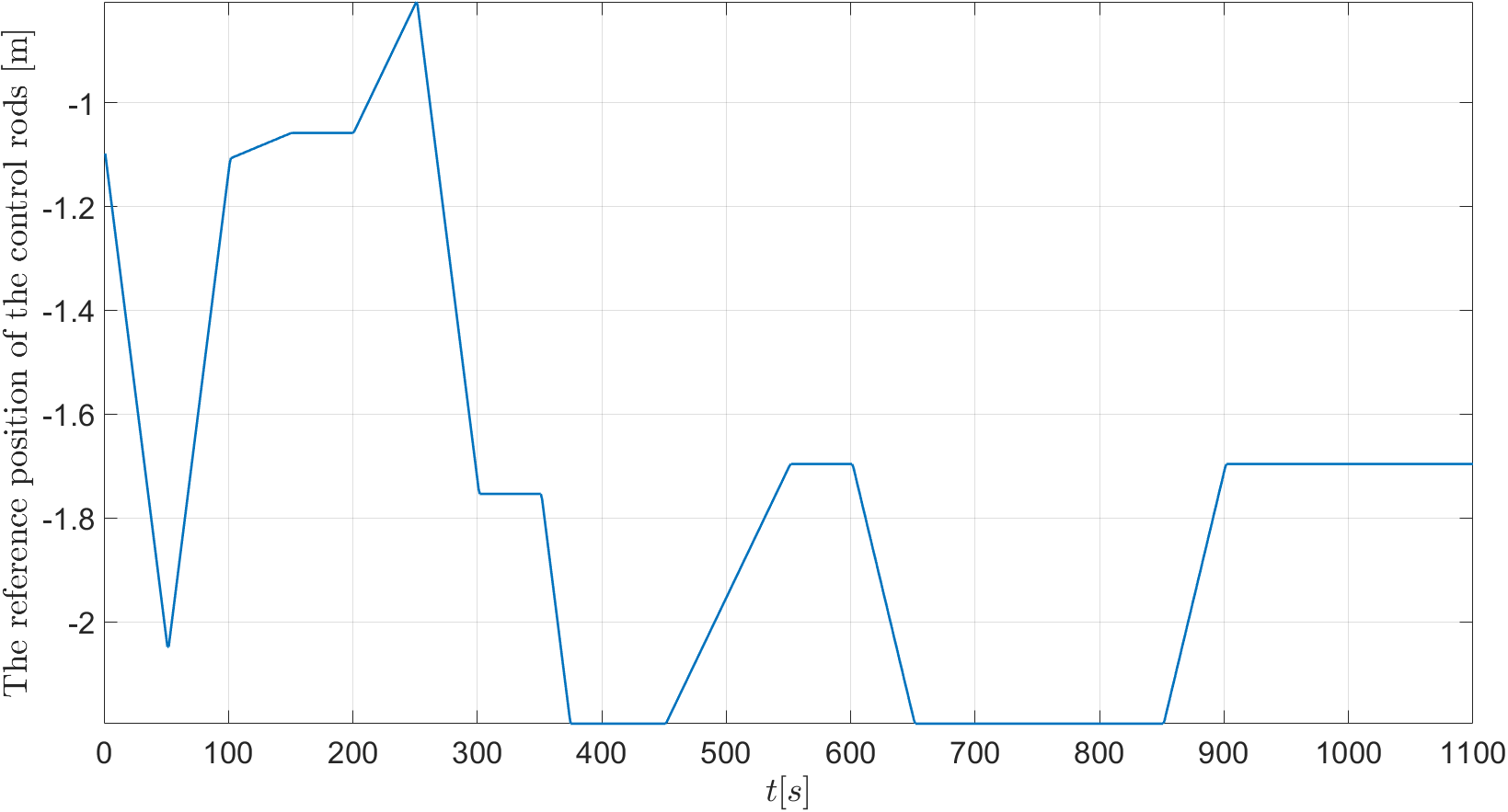} 
    \caption{The reference position of the control rods in a \ac{pwr} -- data set 1.} \label{fig:rods_set_1}
  \end{subfigure}
  \hskip 2em
  \begin{subfigure}[t!]{0.45\textwidth}
    \includegraphics[width=0.99\textwidth]{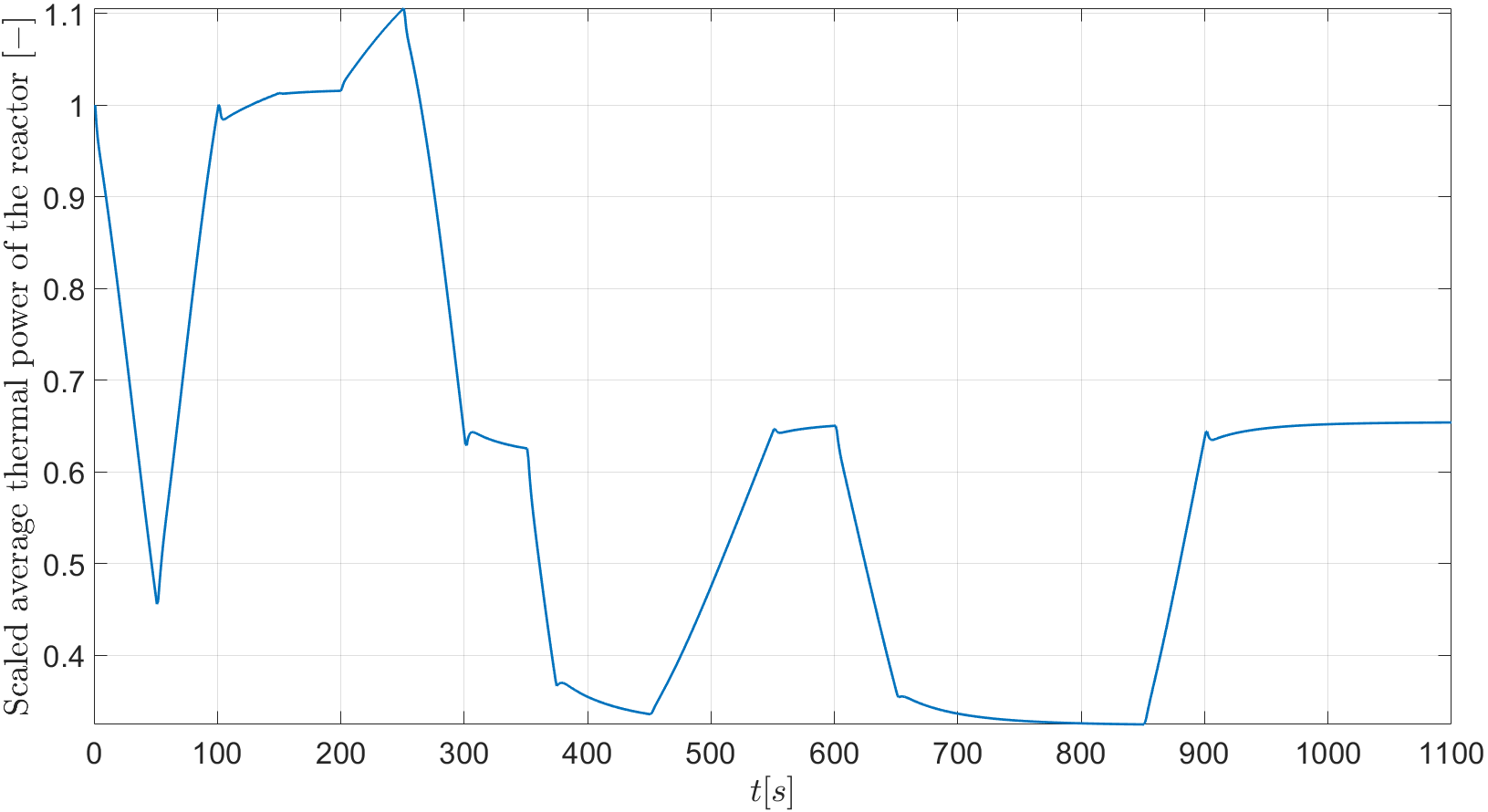} 
    \caption{The reference trajectory of thermal power in a \ac{pwr} -- data set 1.} \label{fig:power_set_1}
  \end{subfigure}
  \vskip 2em

  \begin{subfigure}[t!]{0.45\textwidth}
    \includegraphics[width=0.99\textwidth]{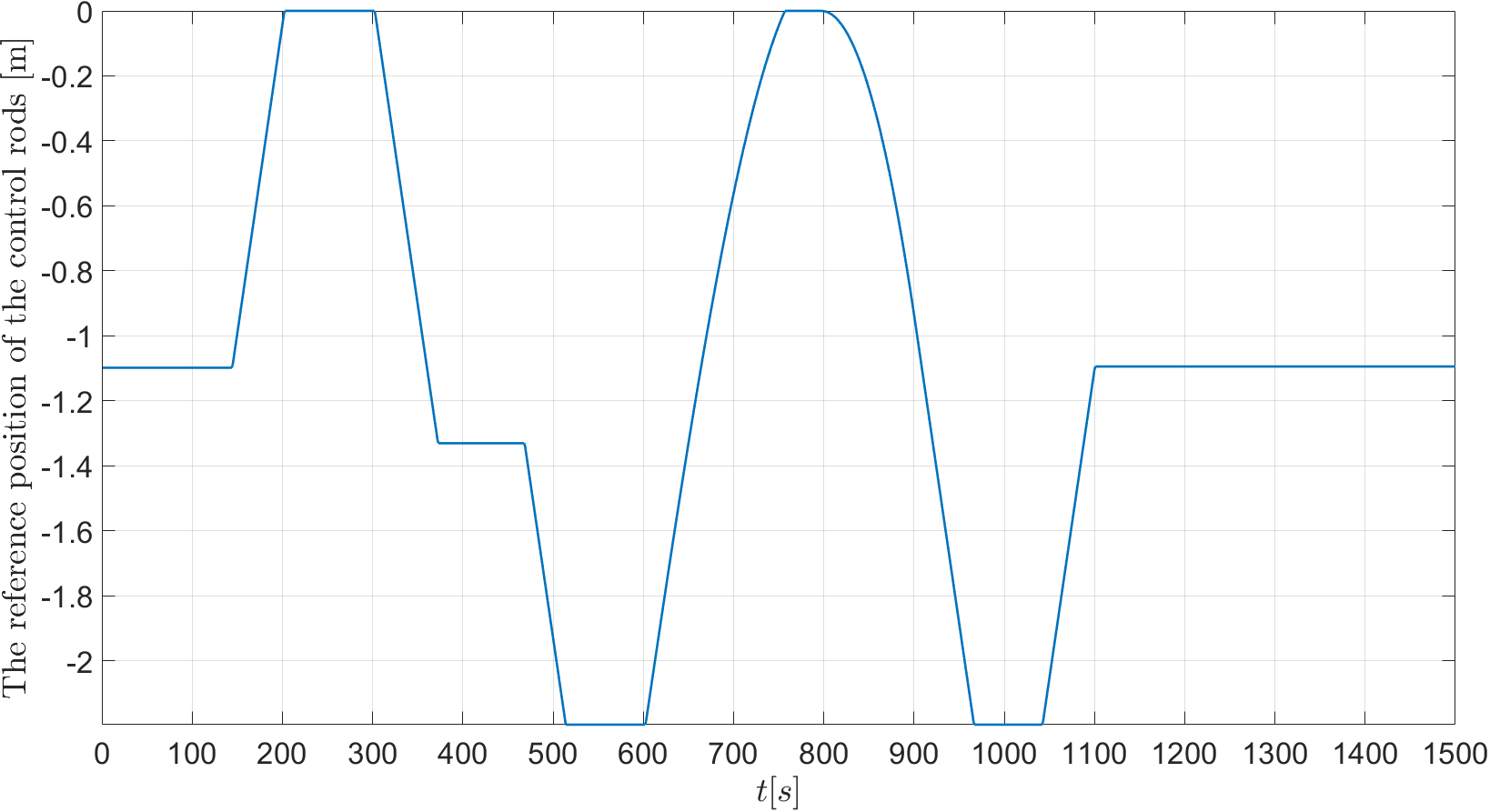} 
    \caption{The reference position of the control rods in a \ac{pwr} -- data set 2.} \label{fig:rods_set_2}
  \end{subfigure}
  \hskip 2em
  \begin{subfigure}[t!]{0.45\textwidth}
    \includegraphics[width=0.99\textwidth]{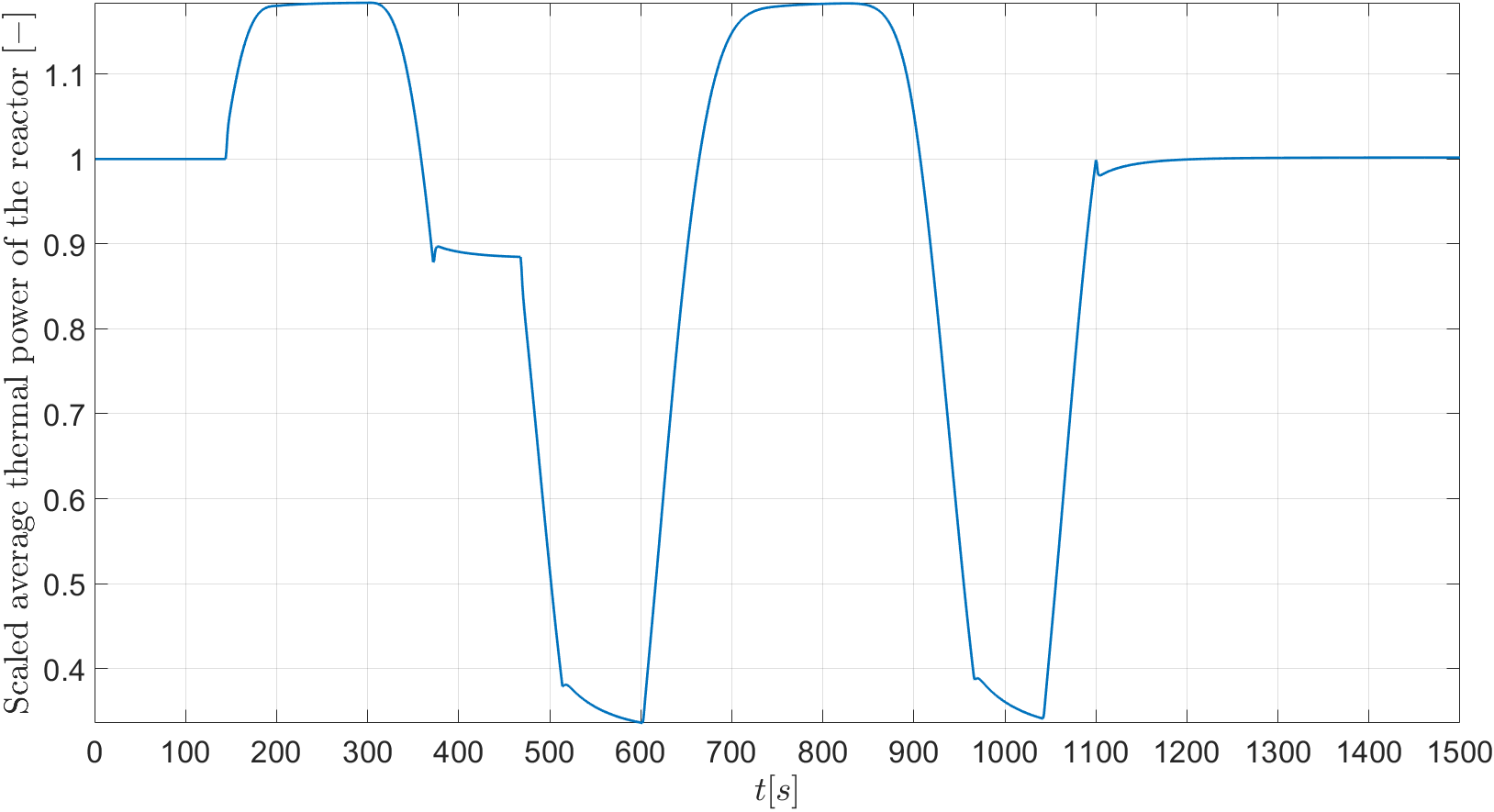} 
    \caption{The reference trajectory of thermal power in a \ac{pwr} -- data set 2.} \label{fig:power_set_2}
  \end{subfigure}
  \vskip 1em
\caption{The reference trajectories of position of the control rods and thermal power in a \ac{pwr} during the verification phase.}\label{fig:ref_verification_phase}
\end{figure}

\begin{figure}
    \begin{subfigure}[t!]{0.45\textwidth}
    \includegraphics[width=0.99\textwidth]{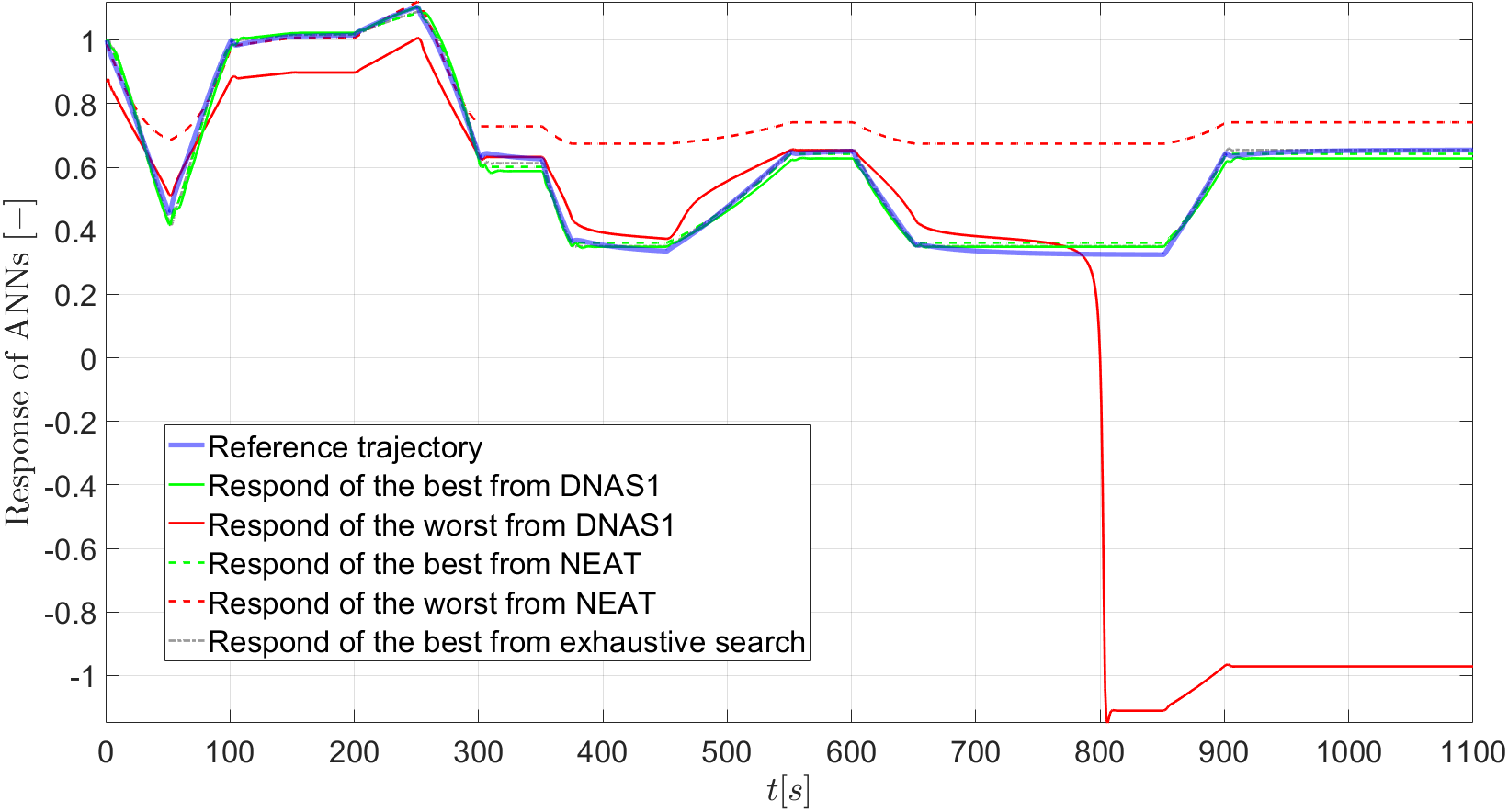} 
    \caption{The average thermal power from \ac{dnas1} algorithm -- data set 1} \label{fig:tp_dnas1_set_1}
  \end{subfigure}
  \hskip 2em
  \begin{subfigure}[t!]{0.45\textwidth}
    \includegraphics[width=0.99\textwidth]{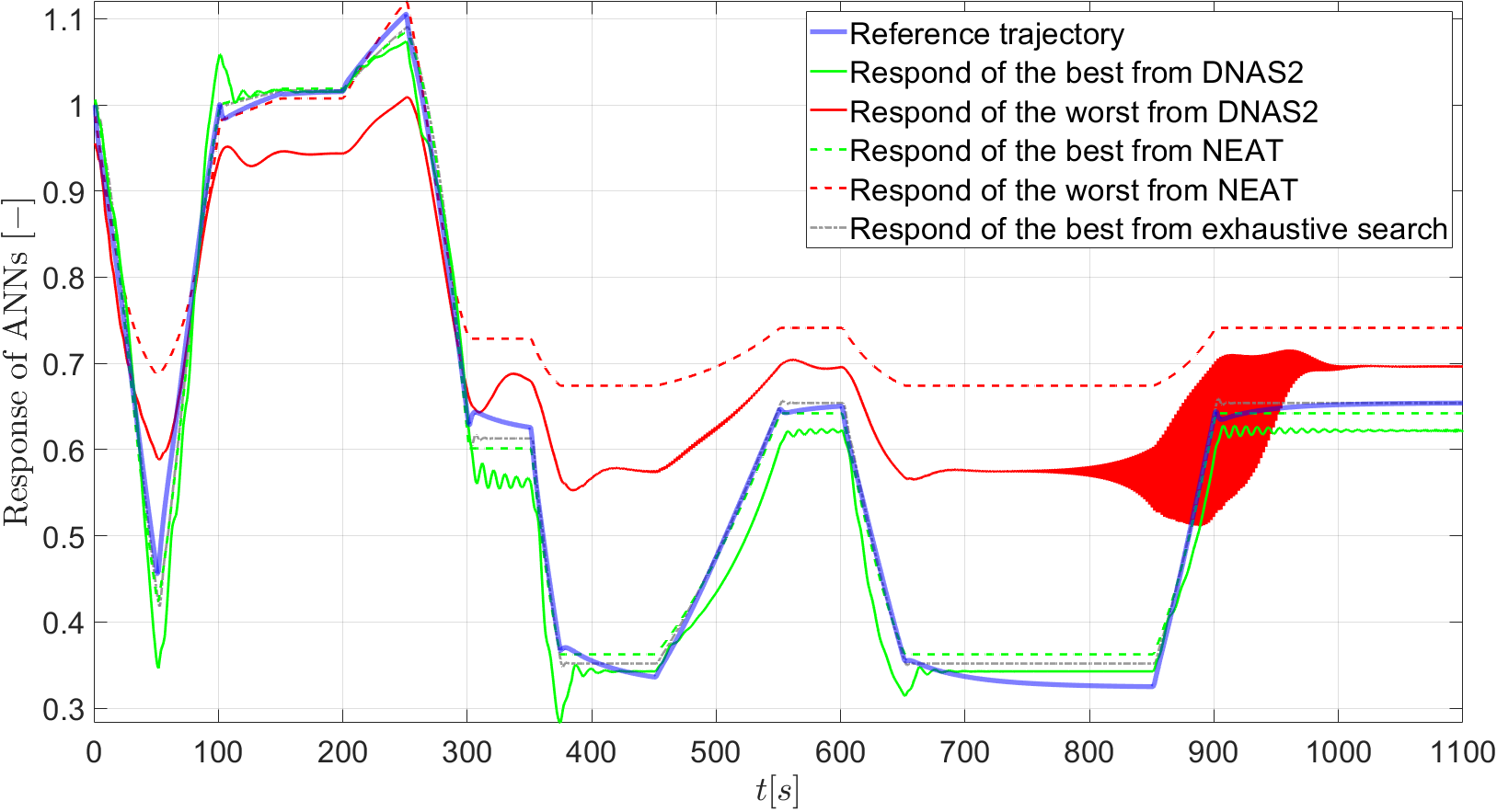} 
    \caption{The average thermal power from \ac{dnas2} algorithm -- data set 1} \label{fig:tp_dnas2_set_1}
  \end{subfigure}
  \vskip 2em
  \begin{subfigure}[t!]{0.45\textwidth}
    \includegraphics[width=0.99\textwidth]{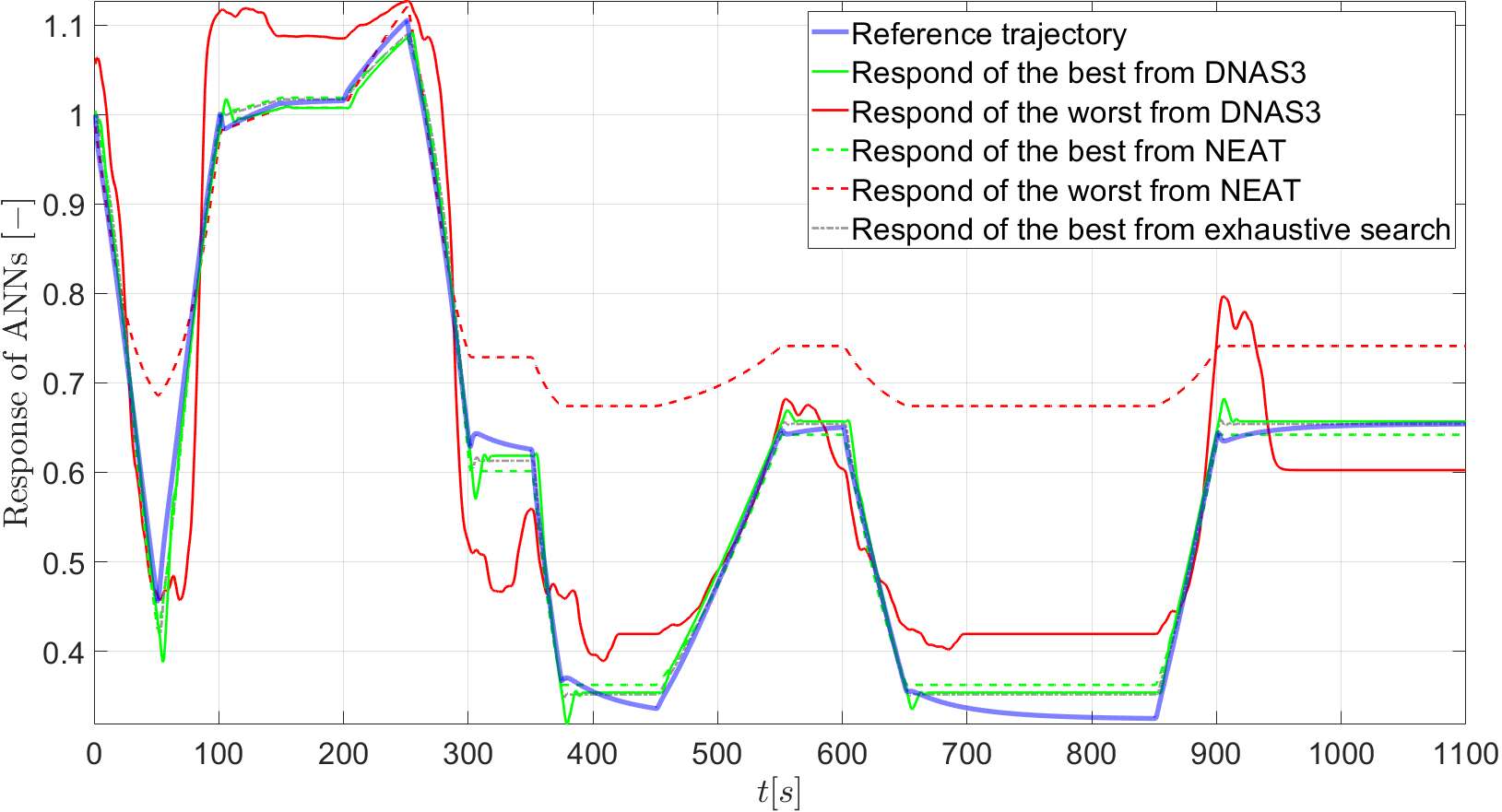} 
    \caption{The average thermal power from \ac{dnas3} algorithm -- data set 1} \label{fig:tp_dnas3_set_1}
  \end{subfigure}
  \hskip 2em
  \begin{subfigure}[t!]{0.45\textwidth}
    \includegraphics[width=0.99\textwidth]{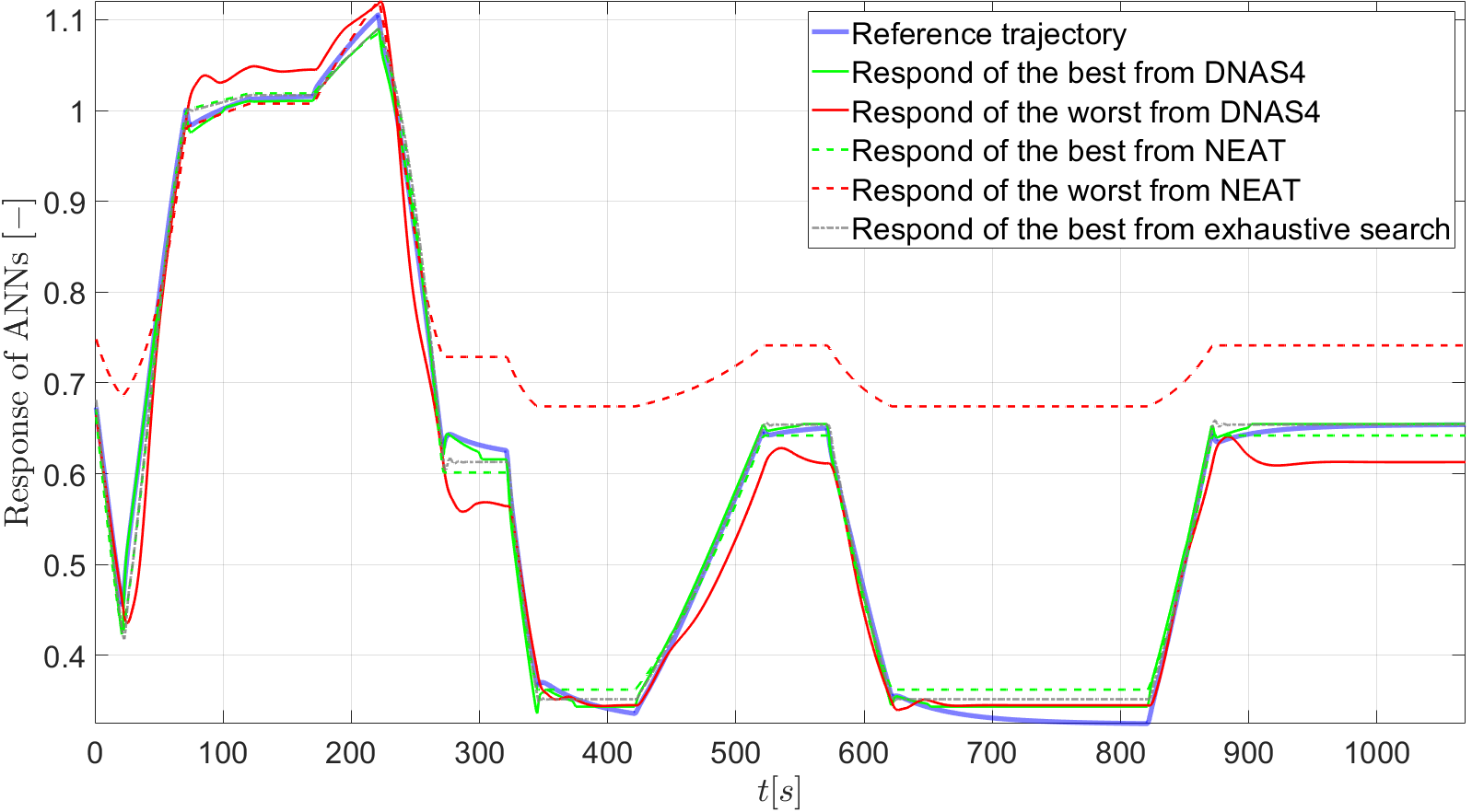} 
    \caption{The average thermal power from \ac{dnas4} algorithm -- data set 1} \label{fig:tp_dnas4_set_1}
  \end{subfigure}
  \vskip 1em
  \caption{The generated trajectories of the average thermal power of the reactor in the verification phase for the first data set.}
  \label{fig:thermal_power_verification_phase_1}
\end{figure}

\begin{figure}
\centering
   \begin{subfigure}[t!]{0.45\textwidth}
    \includegraphics[width=0.99\textwidth]{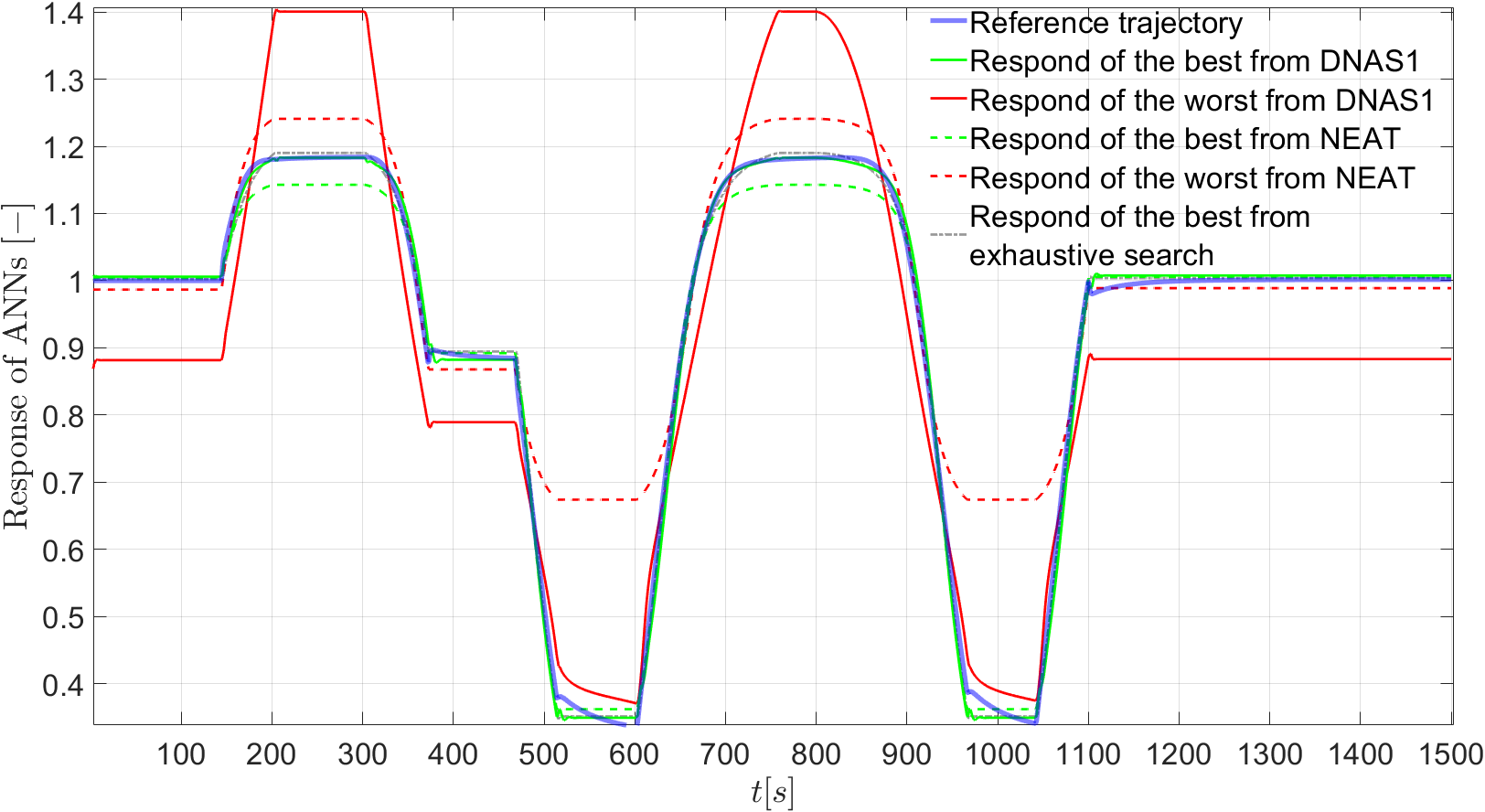} 
    \caption{The average thermal power from \ac{dnas1} algorithm -- data set 2} \label{fig:tp_dnas1_set_2}
  \end{subfigure}
  \hskip 2em
  \begin{subfigure}[t!]{0.45\textwidth}
    \includegraphics[width=0.99\textwidth]{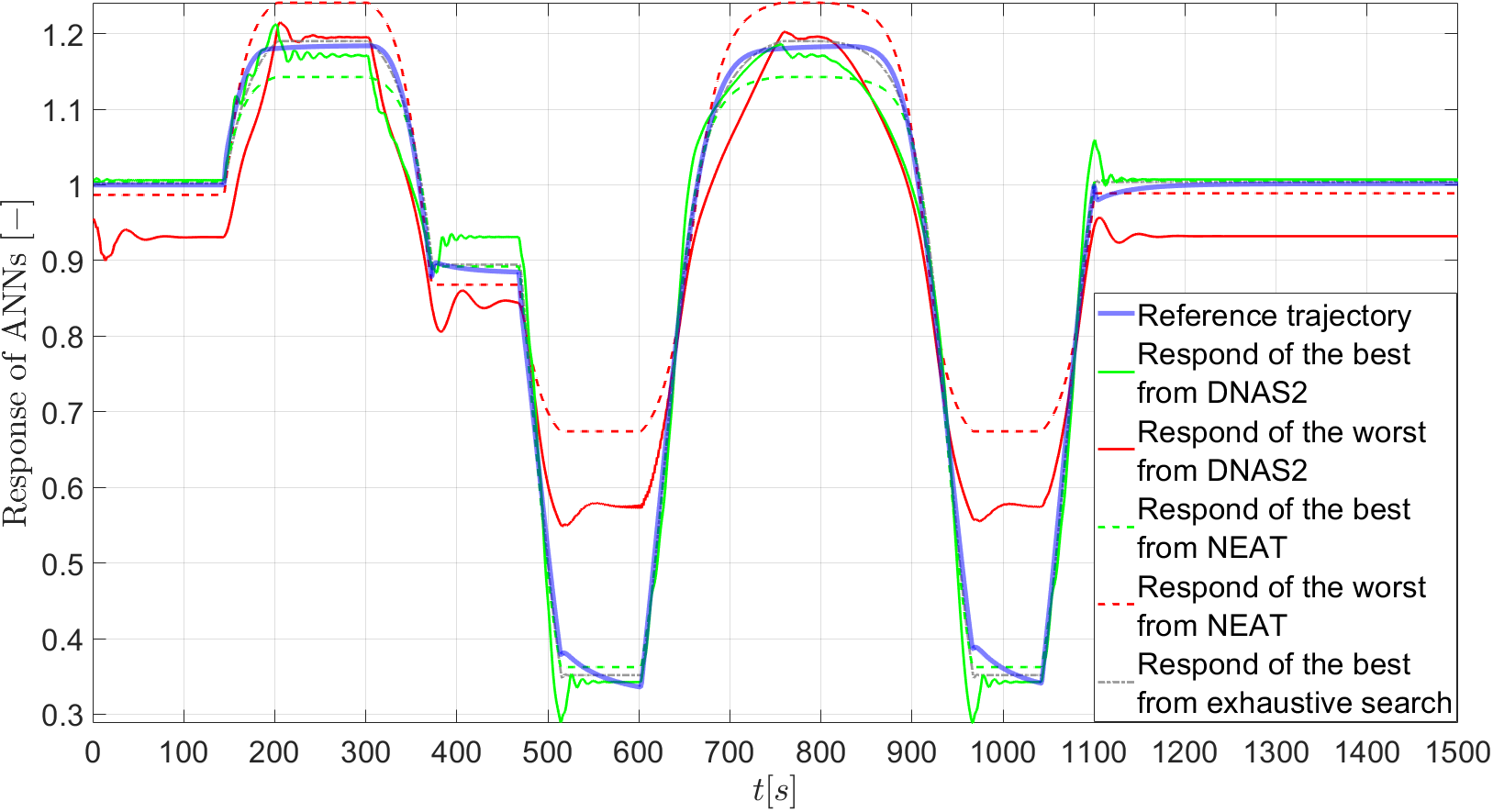} 
    \caption{The average thermal power from \ac{dnas2} algorithm -- data set 2} \label{fig:tp_dnas2_set_2}
  \end{subfigure}
  \vskip 2em
  \begin{subfigure}[t!]{0.45\textwidth}
    \includegraphics[width=0.99\textwidth]{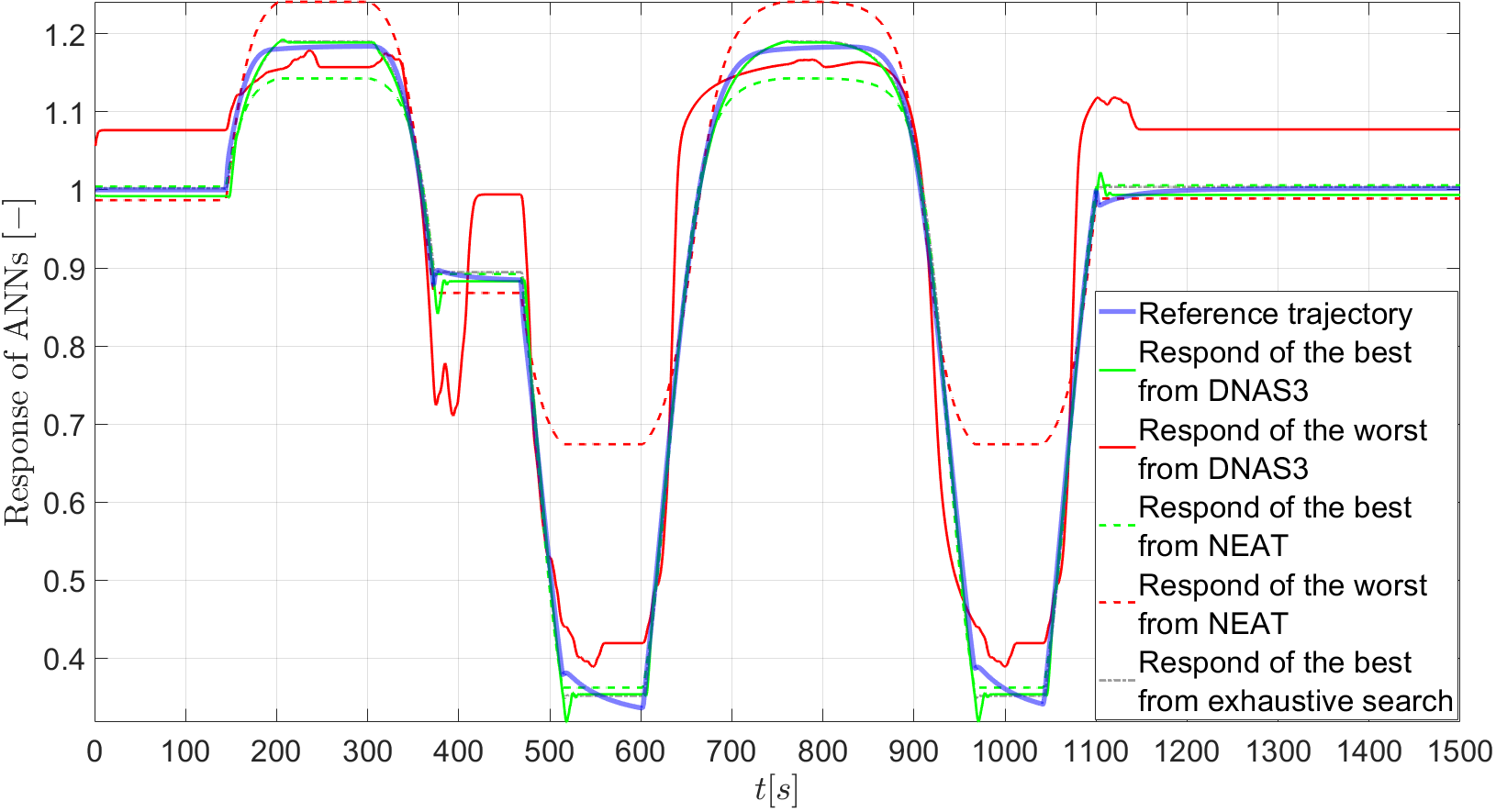} 
    \caption{The average thermal power from \ac{dnas3} algorithm -- data set 2} \label{fig:tp_dnas3_set_2}
  \end{subfigure}
  \hskip 2em
  \begin{subfigure}[t!]{0.45\textwidth}
    \includegraphics[width=0.99\textwidth]{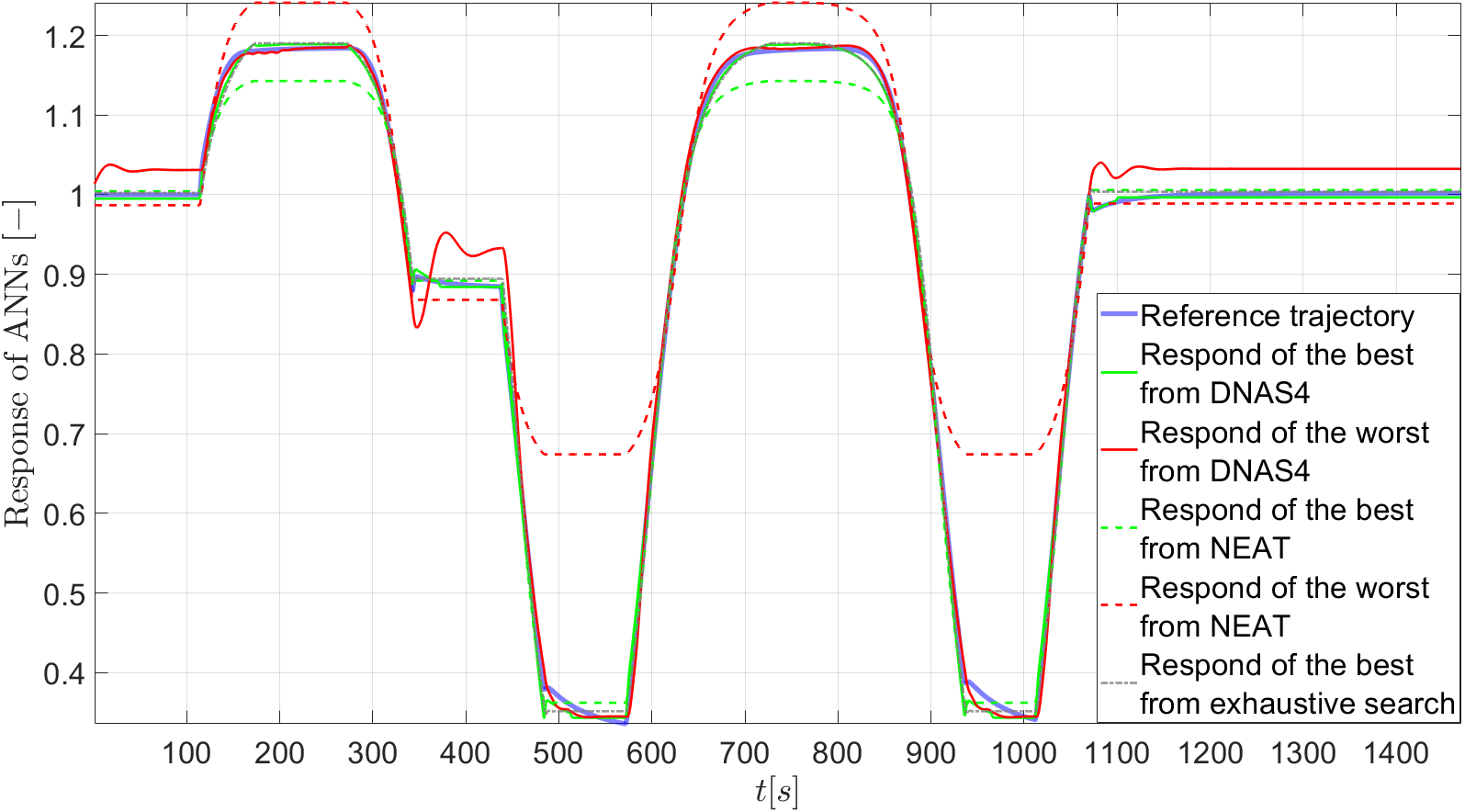} 
    \caption{The average thermal power from \ac{dnas4} algorithm -- data set 2} 
    \label{fig:tp_dnas4_set_2}
  \end{subfigure}
  \hskip 1em
  \caption{The generated trajectories of the average thermal power of the reactor in the verification phase for the second data set.}
  \label{fig:thermal_power_verification_phase_2}
\end{figure}

The reference trajectories using during the verification phase are shown in Fig.~\ref{fig:ref_verification_phase}. As it can be noticed the two data sets have been used during this phase. It is worth adding that those trajectories are prepared based on considerations contained in \cite{Puchalski2020}. The simulation results illustrating the generated trajectories of the average thermal power of the reactor by the best and worst individual from the set of best individuals (relative to the value of the fitness function) for the \ac{dnas1} -- \ac{dnas4} algorithms are given in Figs.~\ref{fig:thermal_power_verification_phase_1} and \ref{fig:thermal_power_verification_phase_2} for the first and second data set, respectively. In turn, the average value of mean errors of the response of the best individuals achieved in subsequent calls of particular algorithms for the data sets 1 and 2 are presented in table~\ref{tab:error_values}. 

\clearpage
\begin{table}
   \caption{The average value of mean errors for \ac{dnas1} -- \ac{dnas4} algorithms during the verification phase.}
   \label{tab:error_values}
   \begin{tabular}{|c|c|c|c|c|c|c|}
   \hline
    \multicolumn{7}{|c|}{\textbf{The average value of mean errors}} \\ 
    \hline \hline
    \textbf{Name} & \textbf{\ac{dnas1}} & \textbf{\ac{dnas2}} & \textbf{\ac{dnas3}} & \textbf{\ac{dnas4}} & \textbf{{NEAT}} & \textbf{Exhaustive search} \\ 
    \hline
    The average value of mean errors -- set 1 & 0.0379 & 0.0788 & 0.0308 & 0.0255 & 0.0410 & 0.0117* \\ 
    \hline
    The average value of mean errors -- set 2 & 0.0341 & 0.0487 & 0.0237 & 0.0117 & 0.0620 & 0.0083*\\ 
    \hline
    \multicolumn{7}{|c|}{* - values for the obtained best individual relative to the fitness function} \\
    \hline
\end{tabular}
\end{table}

Analysing the trajectories presented in Figs.~\ref{fig:thermal_power_verification_phase_1} and \ref{fig:thermal_power_verification_phase_2} and the average value of mean errors from table~\ref{tab:error_values} and taking into account conclusions of the learning phase (see section~\ref{subsubsec:learning_phase}),  it can be stated that the best developed \ac{nas} algorithm is \ac{dnas3} algorithm. Moreover, similarly good results are obtained with the \ac{dnas4} algorithm; however, their cost understood as the computing time is very high. Therefore, the following is stated:
\begin{itemize}
    \item the developed \ac{dnas3} and \ac{dnas4} algorithms may provide at least a good basis for \ac{nas} algorithms for the behavioural (black-box) modelling purposes,
    \item  the \acp{rnn} obtained by the \ac{dnas3} and \ac{dnas4} algorithms can be at least a prototype of the \ac{siso} black-box model of the fast processes in a \ac{pwr}.
\end{itemize}

Moreover, analysing the results obtained from exhaustive search algorithm it can be concluded that they confirm the results obtained with the \ac{dnas1} -- \ac{dnas4} algorithms. More specifically, in the framework of the exhaustive search algorithm performed, the best results are obtained by the network that has one neuron in the hidden layer, with negligible impact on the quality of its performance of the values $du$ and $dy$. Hence, the exhaustive search algorithm performed after the fact, using the knowledge gained from the research conducted by the authors, confirms the effectiveness of the developed \ac{dnas1} -- \ac{dnas4} algorithms to perform the role of a black-box model of the fast processes in a \ac{pwr}. Also, it can be concluded that the basic NEAT algorithm provides neural networks with response performance significantly worse than the developed algorithms, primarily \ac{dnas3} and \ac{dnas4} in the verification phase. Thus, the obtained black-box model of the considered processes in a \ac{pwr} using the basic NEAT algorithm would be worse than the one obtained using the developed \ac{dnas1} -- \ac{dnas4} algorithms.

\section{Conclusions}\label{sec:conclusions}
In this paper, the problem of developing algorithms of artificial neural network architecture search for black-box modelling purposes has been investigated. In particular, four algorithms of artificial neural network architecture search using an evolutionary algorithm and also gradient descent methods have been devised to create the desired models. The specialised evolutionary operators have been delivered to ensure the proper activity of particular algorithms. Finally, it enabled devising the single-input single-output model black-box model of the fast processes in a pressurized water reactor. This model provides the desired trade-off between the size of the network structure and the accuracy of black-box model responses. The proposed algorithms have been implemented in Matlab environment and obtained results yield satisfying performance of the generated output trajectories.

The fact that the created artificial neural networks are not large in size raises the question of whether manual methods using gradient learning would not be enough to search for their optimal structures. The authors claim that no because the number of parameters defining even a small network is too large. Moreover, the developed solutions are further developed both for other types of artificial neural networks, i.e. spiking neural networks and deep neural networks, and for building MIMO models of processes in a pressurized water reactor, e.g., changes in coolant and fuel temperature.

\bibliographystyle{cas-model2-names}

\bibliography{literature}

\begin{thebibliography}{66}
\expandafter\ifx\csname natexlab\endcsname\relax\def\natexlab#1{#1}\fi
\providecommand{\url}[1]{\texttt{#1}}
\providecommand{\href}[2]{#2}
\providecommand{\path}[1]{#1}
\providecommand{\DOIprefix}{doi:}
\providecommand{\ArXivprefix}{arXiv:}
\providecommand{\URLprefix}{URL: }
\providecommand{\Pubmedprefix}{pmid:}
\providecommand{\doi}[1]{\href{http://dx.doi.org/#1}{\path{#1}}}
\providecommand{\Pubmed}[1]{\href{pmid:#1}{\path{#1}}}
\providecommand{\bibinfo}[2]{#2}
\ifx\xfnm\relax \def\xfnm[#1]{\unskip,\space#1}\fi
\bibitem[{Agarwal et~al.(2010)Agarwal, Jain, Kumar and Agrawal}]{Agarwal2010}
\bibinfo{author}{Agarwal, M.}, \bibinfo{author}{Jain, N.},
  \bibinfo{author}{Kumar, M.}, \bibinfo{author}{Agrawal, H.},
  \bibinfo{year}{2010}.
\newblock \bibinfo{title}{Face recognition using eigen faces and neural
  networks}.
\newblock \bibinfo{journal}{International Journal of Computer Theory and
  Engineering} \bibinfo{volume}{2}, \bibinfo{pages}{2--7}.
\newblock \DOIprefix\doi{10.7763/IJCTE.2010.V2.213}.
\bibitem[{Ahmadizar et~al.(2015)Ahmadizar, Soltanian, Akhlaghiantab and
  Tsoulos}]{Ahmadizar2015}
\bibinfo{author}{Ahmadizar, F.}, \bibinfo{author}{Soltanian, K.},
  \bibinfo{author}{Akhlaghiantab, F.}, \bibinfo{author}{Tsoulos, I.},
  \bibinfo{year}{2015}.
\newblock \bibinfo{title}{Artificial neural network development by means of a
  novel combination of grammatical evolution and genetic algorithm}.
\newblock \bibinfo{journal}{Engineering Applications of Artificial
  Intelligence} \bibinfo{volume}{39}, \bibinfo{pages}{1--13}.
\newblock \DOIprefix\doi{10.1016/j.engappai.2014.11.003}.
\bibitem[{Ahn et~al.(2000)Ahn, Cho and Kim}]{Ahn2000}
\bibinfo{author}{Ahn, B.S.}, \bibinfo{author}{Cho, S.S.}, \bibinfo{author}{Kim,
  C.Y.}, \bibinfo{year}{2000}.
\newblock \bibinfo{title}{Integrated methodology of rough set theory and
  artificial neural network for business failure prediction}.
\newblock \bibinfo{journal}{Expert Systems with Applications}
  \bibinfo{volume}{18}, \bibinfo{pages}{65--74}.
\newblock \DOIprefix\doi{10.1016/S0957-4174(99)00053-6}.
\bibitem[{Alom et~al.(2019)Alom, Taha, Yakopcic, Westberg, Sidike, Nasrin,
  Hasan, Van~Essen, Awwal and Asari}]{Alom2019}
\bibinfo{author}{Alom, M.Z.}, \bibinfo{author}{Taha, T.M.},
  \bibinfo{author}{Yakopcic, C.}, \bibinfo{author}{Westberg, S.},
  \bibinfo{author}{Sidike, P.}, \bibinfo{author}{Nasrin, M.S.},
  \bibinfo{author}{Hasan, M.}, \bibinfo{author}{Van~Essen, B.C.},
  \bibinfo{author}{Awwal, A.A.S.}, \bibinfo{author}{Asari, V.K.},
  \bibinfo{year}{2019}.
\newblock \bibinfo{title}{A state-of-the-art survey on deep learning theory and
  architectures}.
\newblock \bibinfo{journal}{Electronics} \bibinfo{volume}{8},
  \bibinfo{pages}{292--358}.
\newblock \DOIprefix\doi{https://doi.org/10.3390/electronics8030292}.
\bibitem[{Azzini and Tettamanzi(2011)}]{Azzini:2011}
\bibinfo{author}{Azzini, A.}, \bibinfo{author}{Tettamanzi, A.G.B.},
  \bibinfo{year}{2011}.
\newblock \bibinfo{title}{Evolutionary {ANN}s: State of the art survey}.
\newblock \bibinfo{journal}{Intelligenza Artificiale} \bibinfo{volume}{5},
  \bibinfo{pages}{19--35}.
\newblock \DOIprefix\doi{10.3233/IA-2011-0002}.
\bibitem[{Bartlett and Basu(1991)}]{Bartlett1991}
\bibinfo{author}{Bartlett, E.}, \bibinfo{author}{Basu, A.},
  \bibinfo{year}{1991}.
\newblock \bibinfo{title}{A dynamic node architecture scheme for
  backpropagation neural networks}, in: \bibinfo{booktitle}{Proceedings of the
  Artificial Neural Networks in Engineering}, pp. \bibinfo{pages}{559--564}.
\bibitem[{Basu(1992)}]{Basu1992}
\bibinfo{author}{Basu, A.}, \bibinfo{year}{1992}.
\newblock \bibinfo{title}{Nuclear Power Plant Status Diagnostics Using an
  Artificial Neural Network with Dynamic Node Architecture}.
\newblock Ph.D. thesis. Iowa State University. \bibinfo{address}{Ames, Iowa,
  US}.
\bibitem[{Bhattacharyya et~al.(2011)Bhattacharyya, Kim and
  Lee}]{Bhattacharyya2011}
\bibinfo{author}{Bhattacharyya, D.}, \bibinfo{author}{Kim, T.H.},
  \bibinfo{author}{Lee, G.S.}, \bibinfo{year}{2011}.
\newblock \bibinfo{title}{Use of artificial neural network in bengali character
  recognition}.
\newblock \bibinfo{journal}{Communications in Computer and Information Science}
  \bibinfo{volume}{260}, \bibinfo{pages}{140--152}.
\newblock \DOIprefix\doi{10.1007/978-3-642-27183-0_15}.
\bibitem[{Bhushan(2009)}]{Bhushan2009}
\bibinfo{author}{Bhushan, B.}, \bibinfo{year}{2009}.
\newblock \bibinfo{title}{Biomimetics: Lessons from nature - an {O}verview}.
\newblock \bibinfo{journal}{Philosophical Transactions of the Royal Society A:
  Mathematical, Physical and Engineering Sciences} \bibinfo{volume}{367},
  \bibinfo{pages}{1445--1486}.
\newblock \DOIprefix\doi{10.1098/rsta.2009.0011}.
\bibitem[{Billings(2013)}]{Billings2013}
\bibinfo{author}{Billings, S.A.}, \bibinfo{year}{2013}.
\newblock \bibinfo{title}{Nonlinear System Identification: NARMAX Methods in
  the Time, Frequency, and Spatio-temporal Domains, 1st Edition}.
\newblock \bibinfo{publisher}{John Wiley {\&} Sons, Inc.},
  \bibinfo{address}{London, UK}.
\bibitem[{Bishop(2006)}]{Bishop2006}
\bibinfo{author}{Bishop, C.M.}, \bibinfo{year}{2006}.
\newblock \bibinfo{title}{Pattern Recognition and Machine Learning}.
\newblock \bibinfo{publisher}{Springer}.
\bibitem[{Brudzewski et~al.(2004)Brudzewski, Osowki and
  Markiewicz}]{Brudzewski2004}
\bibinfo{author}{Brudzewski, K.}, \bibinfo{author}{Osowki, S.},
  \bibinfo{author}{Markiewicz, T.}, \bibinfo{year}{2004}.
\newblock \bibinfo{title}{Classification of milk by means of an electronic nose
  and {SVM} neural network}.
\newblock \bibinfo{journal}{Sensors and Actuators, B: Chemical}
  \bibinfo{volume}{98}, \bibinfo{pages}{291--298}.
\newblock \DOIprefix\doi{10.1016/j.snb.2003.10.028}.
\bibitem[{Candy(2006)}]{Candy2006}
\bibinfo{author}{Candy, J.V.}, \bibinfo{year}{2006}.
\newblock \bibinfo{title}{Model-Based Signal Processing}.
\newblock \bibinfo{publisher}{John Wiley {\&} Sons, Inc.},
  \bibinfo{address}{Hoboken, New Jersey, US}.
\bibitem[{Chen(1996)}]{Chen1996}
\bibinfo{author}{Chen, C.H.}, \bibinfo{year}{1996}.
\newblock \bibinfo{title}{Fuzzy Logic and Neural Network Handbook}.
\newblock \bibinfo{publisher}{McGraw-Hill}, \bibinfo{address}{Michigan, US}.
\bibitem[{Chua and Yang(1988)}]{Chua1988}
\bibinfo{author}{Chua, L.O.}, \bibinfo{author}{Yang, L.}, \bibinfo{year}{1988}.
\newblock \bibinfo{title}{Cellular neural networks: {A}pplications}.
\newblock \bibinfo{journal}{IEEE Transactions on Circuits and Systems}
  \bibinfo{volume}{35}, \bibinfo{pages}{1273--1290}.
\newblock \DOIprefix\doi{10.1109/31.7601}.
\bibitem[{CodeReclaimers({A}ccessed on: Apr. 08, 2021)}]{Neat:2021}
\bibinfo{author}{CodeReclaimers}, \bibinfo{year}{{A}ccessed on: Apr. 08, 2021}.
\newblock \bibinfo{title}{Python implementation of {NEAT}}.
\newblock \URLprefix \url{https://neat-python.readthedocs.io/en/latest/}.
\bibitem[{Cybenko(1989)}]{Cybenko:1989}
\bibinfo{author}{Cybenko, G.}, \bibinfo{year}{1989}.
\newblock \bibinfo{title}{Approximation by superpositions of a sigmoidal
  function}.
\newblock \bibinfo{journal}{Mathematics of Control, Signals and Systems}
  \bibinfo{volume}{2}, \bibinfo{pages}{303--314}.
\newblock \DOIprefix\doi{https://doi.org/10.1007/BF02551274}.
\bibitem[{Dede and Sazli(2010)}]{Dede2010}
\bibinfo{author}{Dede, G.}, \bibinfo{author}{Sazli, M.H.},
  \bibinfo{year}{2010}.
\newblock \bibinfo{title}{Speech recognition with artificial neural networks}.
\newblock \bibinfo{journal}{Digital Signal Processing: A Review Journal}
  \bibinfo{volume}{20}, \bibinfo{pages}{763--768}.
\newblock \DOIprefix\doi{10.1016/j.dsp.2009.10.004}.
\bibitem[{Dreiseitl and Ohno-Machado(2002)}]{Dreiseitl2002}
\bibinfo{author}{Dreiseitl, S.}, \bibinfo{author}{Ohno-Machado, L.},
  \bibinfo{year}{2002}.
\newblock \bibinfo{title}{Logistic regression and artificial neural network
  classification models: a methodology review}.
\newblock \bibinfo{journal}{Journal of Biomedical Informatics}
  \bibinfo{volume}{35}, \bibinfo{pages}{352--359}.
\newblock \DOIprefix\doi{10.1016/S1532-0464(03)00034-0}.
\bibitem[{Ellefsen et~al.(2020)Ellefsen, Huizinga and Torresen}]{Ellefsen:2020}
\bibinfo{author}{Ellefsen, K.O.}, \bibinfo{author}{Huizinga, J.},
  \bibinfo{author}{Torresen, J.}, \bibinfo{year}{2020}.
\newblock \bibinfo{title}{Guiding neuroevolution with structural objectives}.
\newblock \bibinfo{journal}{Evolutionary Computation} \bibinfo{volume}{28},
  \bibinfo{pages}{115--140}.
\newblock \DOIprefix\doi{https://doi.org/10.1162/evco_a_00250}.
\bibitem[{Floreano et~al.(2008)Floreano, D{\"{u}}rr and
  Mattiussi}]{Floreano2008}
\bibinfo{author}{Floreano, D.}, \bibinfo{author}{D{\"{u}}rr, P.},
  \bibinfo{author}{Mattiussi, C.}, \bibinfo{year}{2008}.
\newblock \bibinfo{title}{Neuroevolution: From architectures to learning}.
\newblock \bibinfo{journal}{Evolutionary Intelligence} \bibinfo{volume}{1},
  \bibinfo{pages}{47--62}.
\newblock \DOIprefix\doi{10.1007/s12065-007-0002-4}.
\bibitem[{Fukuda and Shibata(1992)}]{Fukuda1992}
\bibinfo{author}{Fukuda, T.}, \bibinfo{author}{Shibata, T.},
  \bibinfo{year}{1992}.
\newblock \bibinfo{title}{Theory and applications of neural networks}.
\newblock \bibinfo{journal}{IEEE Transactions on Industrial Electrionics}
  \bibinfo{volume}{39}, \bibinfo{pages}{472--489}.
\newblock \DOIprefix\doi{10.1007/978-1-4471-1833-6}.
\bibitem[{Futuyma and Kirkpatrick(2018)}]{Futuyma2018}
\bibinfo{author}{Futuyma, D.J.}, \bibinfo{author}{Kirkpatrick, M.},
  \bibinfo{year}{2018}.
\newblock \bibinfo{title}{Evolution, 4th Edition}.
\newblock \bibinfo{publisher}{Oxford University Press},
  \bibinfo{address}{Oxford, UK}.
\bibitem[{Gadoue et~al.(2009)Gadoue, Giaouris and Finch}]{Gadoue2009}
\bibinfo{author}{Gadoue, S.M.}, \bibinfo{author}{Giaouris, D.},
  \bibinfo{author}{Finch, J.W.}, \bibinfo{year}{2009}.
\newblock \bibinfo{title}{Sensorless control of induction motor drives at very
  low and zero speeds using neural network flux observers}.
\newblock \bibinfo{journal}{IEEE Transactions on Industrial Electronics}
  \bibinfo{volume}{56}, \bibinfo{pages}{3029--3039}.
\newblock \DOIprefix\doi{10.1109/TIE.2009.2024665}.
\bibitem[{Gupta and Raza(2019)}]{Gupta:2019}
\bibinfo{author}{Gupta, T.K.}, \bibinfo{author}{Raza, R.},
  \bibinfo{year}{2019}.
\newblock \bibinfo{title}{Optimization of {ANN} architecture: {A} review on
  nature-inspired techniques}, in: \bibinfo{editor}{Dey, N.},
  \bibinfo{editor}{Borra, S.}, \bibinfo{editor}{Ashour, A.S.},
  \bibinfo{editor}{Shi, F.} (Eds.), \bibinfo{booktitle}{Machine Learning in
  Bio-Signal Analysis and Diagnostic Imaging}. \bibinfo{publisher}{Academic
  Press}, pp. \bibinfo{pages}{159--182}.
\bibitem[{Haykin(2011)}]{Haykin2011}
\bibinfo{author}{Haykin, S.O.}, \bibinfo{year}{2011}.
\newblock \bibinfo{title}{Neural Networks and Learning Machines, 3rd Edition}.
\newblock \bibinfo{publisher}{Pearson Education}, \bibinfo{address}{London,
  UK}.
\bibitem[{Hornik et~al.(1989)Hornik, Stinchcombe and White}]{Hornik:1989}
\bibinfo{author}{Hornik, K.}, \bibinfo{author}{Stinchcombe, M.},
  \bibinfo{author}{White, H.}, \bibinfo{year}{1989}.
\newblock \bibinfo{title}{Multilayer feedforward networks are universal
  approximators}.
\newblock \bibinfo{journal}{Neural Networks} \bibinfo{volume}{2},
  \bibinfo{pages}{359--366}.
\newblock \DOIprefix\doi{https://doi.org/10.1016/0893-6080(89)90020-8}.
\bibitem[{Kandel et~al.(2012)Kandel, Schwartz, Jessell, Siegelbaum and
  Hudspeth}]{Kandel2012}
\bibinfo{author}{Kandel, E.R.}, \bibinfo{author}{Schwartz, J.H.},
  \bibinfo{author}{Jessell, T.M.}, \bibinfo{author}{Siegelbaum, S.A.},
  \bibinfo{author}{Hudspeth, A.J.}, \bibinfo{year}{2012}.
\newblock \bibinfo{title}{Principles of Neural Science, 5th Edition}.
\newblock \bibinfo{publisher}{McGraw-Hill}.
\bibitem[{Kashiwao et~al.(2017)Kashiwao, Nakayama, Ando, Ikeda, Lee and
  Bahadori}]{Kashiwao:2017}
\bibinfo{author}{Kashiwao, T.}, \bibinfo{author}{Nakayama, K.},
  \bibinfo{author}{Ando, S.}, \bibinfo{author}{Ikeda, K.},
  \bibinfo{author}{Lee, M.}, \bibinfo{author}{Bahadori, A.},
  \bibinfo{year}{2017}.
\newblock \bibinfo{title}{A neural network-based local rainfall prediction
  system using meteorological data on the internet: {A} case study using data
  from the {J}apan {M}eteorological {A}gency}.
\newblock \bibinfo{journal}{Applied Soft Computing} \bibinfo{volume}{56},
  \bibinfo{pages}{317--330}.
\newblock \DOIprefix\doi{https://doi.org/10.1016/j.asoc.2017.03.015}.
\bibitem[{Kavzoglu(1999)}]{Kavzoglu1999}
\bibinfo{author}{Kavzoglu, T.}, \bibinfo{year}{1999}.
\newblock \bibinfo{title}{Determining optimum structure for artificial neural
  networks}, in: \bibinfo{booktitle}{Proceedings of the 25th Annual Technical
  Conference and Exhibition of the Remote Sensing Society}, pp.
  \bibinfo{pages}{675--682}.
\bibitem[{Kerlin(1978)}]{Kerlin1978}
\bibinfo{author}{Kerlin, T.W.}, \bibinfo{year}{1978}.
\newblock \bibinfo{title}{Dynamic analysis and control of pressurized water
  reactors}.
\newblock \bibinfo{journal}{Control and Dynamic Systems} \bibinfo{volume}{14},
  \bibinfo{pages}{103--212}.
\newblock \DOIprefix\doi{https://doi.org/10.1016/B978-0-12-012714-6.50008-8}.
\bibitem[{Kirkpatrick et~al.(1983)Kirkpatrick, Gelatt~Jr. and
  Vecchi}]{Kirkpatrick1983}
\bibinfo{author}{Kirkpatrick, S.}, \bibinfo{author}{Gelatt~Jr., C.D.},
  \bibinfo{author}{Vecchi, M.P.}, \bibinfo{year}{1983}.
\newblock \bibinfo{title}{Optimization by simulated annealing}.
\newblock \bibinfo{journal}{Science} \bibinfo{volume}{220},
  \bibinfo{pages}{671--680}.
\newblock \DOIprefix\doi{10.1126/science.220.4598.671}.
\bibitem[{Koza(1998)}]{Koza1998}
\bibinfo{author}{Koza, J.R.}, \bibinfo{year}{1998}.
\newblock \bibinfo{title}{Genetic Programming: On the Programming of Computers
  by Means of Natural Selection, 6th Edition}.
\newblock \bibinfo{publisher}{Massachusetts Institute of Technology},
  \bibinfo{address}{Cambridge, Massachusetts, US}.
\bibitem[{Krizhevsky et~al.(2017)Krizhevsky, Sutskever and
  Hinton}]{Krizhevsky2017}
\bibinfo{author}{Krizhevsky, A.}, \bibinfo{author}{Sutskever, I.},
  \bibinfo{author}{Hinton, G.E.}, \bibinfo{year}{2017}.
\newblock \bibinfo{title}{Imagenet classification with deep convolutional
  neural networks}.
\newblock \bibinfo{journal}{Communications of the ACM} \bibinfo{volume}{60},
  \bibinfo{pages}{84--90}.
\newblock \DOIprefix\doi{10.1145/3065386}.
\bibitem[{Kurkova(1992)}]{Kurkova1992}
\bibinfo{author}{Kurkova, V.}, \bibinfo{year}{1992}.
\newblock \bibinfo{title}{Kolmogorov's theorem and multilayer neural networks}.
\newblock \bibinfo{journal}{Neural Networks} \bibinfo{volume}{5},
  \bibinfo{pages}{501--506}.
\newblock \DOIprefix\doi{https://doi.org/10.1016/0893-6080(92)90012-8}.
\bibitem[{Llobet et~al.(1999)Llobet, Hines, Gardner and Franco}]{Llobet1999}
\bibinfo{author}{Llobet, E.}, \bibinfo{author}{Hines, E.L.},
  \bibinfo{author}{Gardner, J.W.}, \bibinfo{author}{Franco, S.},
  \bibinfo{year}{1999}.
\newblock \bibinfo{title}{Non-destructive banana ripeness determination using a
  neural network-based electronic nose}.
\newblock \bibinfo{journal}{Measurement Science and Technology}
  \bibinfo{volume}{10}, \bibinfo{pages}{538--548}.
\newblock \DOIprefix\doi{10.1088/0957-0233/10/6/320}.
\bibitem[{Michalewicz(1996)}]{Michalewicz1996}
\bibinfo{author}{Michalewicz, Z.}, \bibinfo{year}{1996}.
\newblock \bibinfo{title}{Genetic Algorithms + Data Structures = Evolution
  Programs}.
\newblock \bibinfo{publisher}{Springer}.
\bibitem[{Naghedolfeizi(1990)}]{Naghedolfeizi1990}
\bibinfo{author}{Naghedolfeizi, M.}, \bibinfo{year}{1990}.
\newblock \bibinfo{title}{Dynamic Modeling of a Pressurized Water Reactor Plant
  for Diagnostics and Control}.
\newblock Master's thesis. University of Tennessee.
  \bibinfo{address}{Knoxville, US}.
\bibitem[{Nolfi and Parisi(1997)}]{Nolfi1997}
\bibinfo{author}{Nolfi, S.}, \bibinfo{author}{Parisi, D.},
  \bibinfo{year}{1997}.
\newblock \bibinfo{title}{Evolution of artificial neural networks}.
\newblock \bibinfo{journal}{Neural Networks} \bibinfo{volume}{2},
  \bibinfo{pages}{1--8}.
\newblock \DOIprefix\doi{10.1016/j.microc.2008.10.006}.
\bibitem[{Norgaard et~al.(2000)Norgaard, Ravn, Poulsen and
  Hansen}]{Norgaard:2000}
\bibinfo{author}{Norgaard, M.}, \bibinfo{author}{Ravn, O.},
  \bibinfo{author}{Poulsen, N.K.}, \bibinfo{author}{Hansen, L.K.},
  \bibinfo{year}{2000}.
\newblock \bibinfo{title}{Neural Networks for Modelling and Control of Dynamic
  Systems: {A} Practitioner’s Handbook}.
\newblock \bibinfo{publisher}{Springer Verlag}, \bibinfo{address}{London, UK}.
\bibitem[{Papavasileiou et~al.(2021)Papavasileiou, Cornelis and
  Jansen}]{Papavasileiou:2021}
\bibinfo{author}{Papavasileiou, E.}, \bibinfo{author}{Cornelis, J.},
  \bibinfo{author}{Jansen, B.}, \bibinfo{year}{2021}.
\newblock \bibinfo{title}{A systematic literature review of the successors of
  'neuroevolution of augmenting topologies'}.
\newblock \bibinfo{journal}{Evolutionary Computation} \bibinfo{volume}{29},
  \bibinfo{pages}{1--73}.
\newblock \DOIprefix\doi{https://doi.org/10.1162/evco_a_00282}.
\bibitem[{Perrusquía and Yu(2021)}]{Perrusquia:2021}
\bibinfo{author}{Perrusquía, A.}, \bibinfo{author}{Yu, W.},
  \bibinfo{year}{2021}.
\newblock \bibinfo{title}{Identification and optimal control of nonlinear
  systems using recurrent neural networks and reinforcement learning: {A}n
  overview}.
\newblock \bibinfo{journal}{Neurocomputing} \bibinfo{volume}{438},
  \bibinfo{pages}{145--154}.
\newblock \DOIprefix\doi{https://doi.org/10.1016/j.neucom.2021.01.096}.
\bibitem[{Ponulak and Kasi{\'{n}}ski(2011)}]{Ponulak2011}
\bibinfo{author}{Ponulak, F.}, \bibinfo{author}{Kasi{\'{n}}ski, A.},
  \bibinfo{year}{2011}.
\newblock \bibinfo{title}{Introduction to spiking neural networks: Information
  processing, learning and applications}.
\newblock \bibinfo{journal}{Acta Neurobiologiae Experimentalis}
  \bibinfo{volume}{71}, \bibinfo{pages}{409--433}.
\bibitem[{Puchalski(2018)}]{Puchalski2018}
\bibinfo{author}{Puchalski, B.}, \bibinfo{year}{2018}.
\newblock \bibinfo{title}{Sterowanie z wykorzystaniem rachunku niecałkowitego
  rzędu reaktorem wodnym ciśnieniowym elektrowni jądrowej (in Polish)}.
\newblock Ph.D. thesis. Gda\'nsk University of Technology.
  \bibinfo{address}{Gda\'nsk, Poland}.
\bibitem[{Puchalski and Rutkowski(2020)}]{Puchalski2020a}
\bibinfo{author}{Puchalski, B.}, \bibinfo{author}{Rutkowski, T.A.},
  \bibinfo{year}{2020}.
\newblock \bibinfo{title}{Approximation of fractional order dynamic systems
  using elman, {GRU} and {LSTM} neural networks}, in:
  \bibinfo{editor}{Rutkowski, L.}, \bibinfo{editor}{Scherer, R.},
  \bibinfo{editor}{Korytkowski, M.}, \bibinfo{editor}{Pedrycz, W.},
  \bibinfo{editor}{Tadeusiewicz, R.}, \bibinfo{editor}{Zurada, J.M.} (Eds.),
  \bibinfo{booktitle}{Artificial Intelligence and Soft Computing. ICAISC 2020.
  Lecture Notes in Computer Science}, \bibinfo{publisher}{Springer, Cham}. pp.
  \bibinfo{pages}{215--230}.
\newblock \DOIprefix\doi{https://doi.org/10.1007/978-3-030-61401-0_21}.
\bibitem[{Puchalski et~al.(2017)Puchalski, Rutkowski and
  Duzinkiewicz}]{Puchalski2017}
\bibinfo{author}{Puchalski, B.}, \bibinfo{author}{Rutkowski, T.A.},
  \bibinfo{author}{Duzinkiewicz, K.}, \bibinfo{year}{2017}.
\newblock \bibinfo{title}{Nodal models of pressurized water reactor core for
  control purposes – a comparison study}.
\newblock \bibinfo{journal}{Nuclear Engineering and Design}
  \bibinfo{volume}{322}, \bibinfo{pages}{444--463}.
\newblock \DOIprefix\doi{https://doi.org/10.1016/j.nucengdes.2017.07.005}.
\bibitem[{Puchalski et~al.(2020)Puchalski, Rutkowski and
  Duzinkiewicz}]{Puchalski2020}
\bibinfo{author}{Puchalski, B.}, \bibinfo{author}{Rutkowski, T.A.},
  \bibinfo{author}{Duzinkiewicz, K.}, \bibinfo{year}{2020}.
\newblock \bibinfo{title}{Fuzzy multi-regional fractional {PID} controller for
  pressurized water nuclear reactor}.
\newblock \bibinfo{journal}{ISA Transactions} \bibinfo{volume}{103},
  \bibinfo{pages}{86--102}.
\newblock \DOIprefix\doi{https://doi.org/10.1016/j.isatra.2020.04.003}.
\bibitem[{Rios et~al.(2020)Rios, Alanis, Arana-Daniel and
  Lopez-Franco}]{Rios:2020}
\bibinfo{author}{Rios, J.D.}, \bibinfo{author}{Alanis, A.Y.},
  \bibinfo{author}{Arana-Daniel, N.}, \bibinfo{author}{Lopez-Franco, C.},
  \bibinfo{year}{2020}.
\newblock \bibinfo{title}{Neural networks modeling and control: Applications
  for unknown nonlinear delayed systems in discrete time}, in:
  \bibinfo{editor}{Sanchez, E.N.} (Ed.), \bibinfo{booktitle}{Neural Networks
  Modeling and Control: Applications for Unknown Nonlinear Delayed Systems in
  Discrete Time}. \bibinfo{publisher}{Academic Press}.
\bibitem[{Roffel and Betlem(2006)}]{Roffel2006}
\bibinfo{author}{Roffel, B.}, \bibinfo{author}{Betlem, B.},
  \bibinfo{year}{2006}.
\newblock \bibinfo{title}{Process Dynamic and Control. Modelling for Control
  and Prediction}.
\newblock \bibinfo{publisher}{John Wiley {\&} Sons, Inc.},
  \bibinfo{address}{Chichester, West Sussex, UK}.
\bibitem[{SaiSindhuTheja and Shyam(2021)}]{SaiSindhuTheja:2021}
\bibinfo{author}{SaiSindhuTheja, R.}, \bibinfo{author}{Shyam, G.K.},
  \bibinfo{year}{2021}.
\newblock \bibinfo{title}{An efficient metaheuristic algorithm based feature
  selection and recurrent neural network for {D}o{S} attack detection in cloud
  computing environment}.
\newblock \bibinfo{journal}{Applied Soft Computing} \bibinfo{volume}{100}.
\newblock \DOIprefix\doi{https://doi.org/10.1016/j.asoc.2020.106997}.
\bibitem[{Saxena and Saad(2007)}]{Saxena2007}
\bibinfo{author}{Saxena, A.}, \bibinfo{author}{Saad, A.}, \bibinfo{year}{2007}.
\newblock \bibinfo{title}{Evolving an artificial neural network classifier for
  condition monitoring of rotating mechanical systems}.
\newblock \bibinfo{journal}{Applied Soft Computing Journal}
  \bibinfo{volume}{7}, \bibinfo{pages}{441--454}.
\newblock \DOIprefix\doi{10.1016/j.asoc.2005.10.001}.
\bibitem[{Siebel and Sommer(2007)}]{Siebel2007}
\bibinfo{author}{Siebel, N.T.}, \bibinfo{author}{Sommer, G.},
  \bibinfo{year}{2007}.
\newblock \bibinfo{title}{Evolutionary reinforcement learning of artificial
  neural networks}.
\newblock \bibinfo{journal}{International Journal of Hybrid Intelligent
  Systems} \bibinfo{volume}{4}, \bibinfo{pages}{171--183}.
\newblock \DOIprefix\doi{10.3233/HIS-2007-4304}.
\bibitem[{Stanley et~al.(2019)Stanley, Clune, Lehman and
  Miikkulainen}]{Stanley2019}
\bibinfo{author}{Stanley, K.O.}, \bibinfo{author}{Clune, J.},
  \bibinfo{author}{Lehman, J.}, \bibinfo{author}{Miikkulainen, R.},
  \bibinfo{year}{2019}.
\newblock \bibinfo{title}{Designing neural networks through neuroevolution}.
\newblock \bibinfo{journal}{Nature Machine Intelligence} \bibinfo{volume}{1},
  \bibinfo{pages}{24--35}.
\newblock \DOIprefix\doi{10.1038/s42256-018-0006-z}.
\bibitem[{Stanley et~al.(2009)Stanley, D'Ambrosio and Gauci}]{Stanley2009}
\bibinfo{author}{Stanley, K.O.}, \bibinfo{author}{D'Ambrosio, D.B.},
  \bibinfo{author}{Gauci, J.}, \bibinfo{year}{2009}.
\newblock \bibinfo{title}{A hypercube-based encoding for evolving large-scale
  neural networks}.
\newblock \bibinfo{journal}{Artificial Life} \bibinfo{volume}{15},
  \bibinfo{pages}{185--212}.
\newblock \DOIprefix\doi{10.1162/artl.2009.15.2.15202}.
\bibitem[{Stanley and Miikkulainen(2002)}]{Stanley2002}
\bibinfo{author}{Stanley, K.O.}, \bibinfo{author}{Miikkulainen, R.},
  \bibinfo{year}{2002}.
\newblock \bibinfo{title}{Evolving neural networks through augmenting
  topologies}.
\newblock \bibinfo{journal}{Evolutionary Computation} \bibinfo{volume}{10},
  \bibinfo{pages}{99--127}.
\bibitem[{Stubberud et~al.(1995)Stubberud, Lobbia and Owen}]{Stubberud1995}
\bibinfo{author}{Stubberud, S.C.}, \bibinfo{author}{Lobbia, R.N.},
  \bibinfo{author}{Owen, M.}, \bibinfo{year}{1995}.
\newblock \bibinfo{title}{An adaptive extended kalman filter using artificial
  neural networks}, in: \bibinfo{booktitle}{Proceedings of the 34th IEEE
  Conference on Decision and Control}, pp. \bibinfo{pages}{1852--1856}.
\bibitem[{Sun et~al.(2013)Sun, Chen, Yang and Zhu}]{Sun2013}
\bibinfo{author}{Sun, X.}, \bibinfo{author}{Chen, L.}, \bibinfo{author}{Yang,
  Z.}, \bibinfo{author}{Zhu, H.}, \bibinfo{year}{2013}.
\newblock \bibinfo{title}{Speed-sensorless vector control of a bearingless
  induction motor with artificial neural network inverse speed observer}.
\newblock \bibinfo{journal}{IEEE/ASME Transactions on Mechatronics}
  \bibinfo{volume}{18}, \bibinfo{pages}{1357--1366}.
\newblock \DOIprefix\doi{10.1109/TMECH.2012.2202123}.
\bibitem[{Sutskever et~al.(2011)Sutskever, Martens and Hinton}]{Sutskever2011}
\bibinfo{author}{Sutskever, I.}, \bibinfo{author}{Martens, J.},
  \bibinfo{author}{Hinton, G.E.}, \bibinfo{year}{2011}.
\newblock \bibinfo{title}{Generating text with recurrent neural networks}, in:
  \bibinfo{booktitle}{Proceedings of the 28th International Conference on
  Machine Learning}, pp. \bibinfo{pages}{528--535}.
\bibitem[{Tam and Kiang(1992)}]{Tam1992}
\bibinfo{author}{Tam, K.Y.}, \bibinfo{author}{Kiang, M.Y.},
  \bibinfo{year}{1992}.
\newblock \bibinfo{title}{Managerial applications of neural networks: The case
  of bank failure predictions}.
\newblock \bibinfo{journal}{Management Science} \bibinfo{volume}{38},
  \bibinfo{pages}{926--947}.
\newblock \DOIprefix\doi{10.1287/mnsc.38.7.926}.
\bibitem[{Tatjewski(2007)}]{Tatjewski2007}
\bibinfo{author}{Tatjewski, P.}, \bibinfo{year}{2007}.
\newblock \bibinfo{title}{Advanced Control of Industrial Processes. Structures
  and Algorithms}.
\newblock \bibinfo{publisher}{Springer-Verlag}, \bibinfo{address}{London, UK}.
\bibitem[{Tetko et~al.(1995)Tetko, Livingstone and Luik}]{Tetko1995}
\bibinfo{author}{Tetko, I.V.}, \bibinfo{author}{Livingstone, D.J.},
  \bibinfo{author}{Luik, A.I.}, \bibinfo{year}{1995}.
\newblock \bibinfo{title}{Neural network studies. 1. comparison of overfitting
  and overtraining}.
\newblock \bibinfo{journal}{Journal of Chemical Information and Modeling}
  \bibinfo{volume}{35}, \bibinfo{pages}{826--833}.
\newblock \DOIprefix\doi{10.1021/ci00027a006}.
\bibitem[{The~MathWorks({A}ccessed on: Apr. 08, 2021)}]{Narx:2021}
\bibinfo{author}{The~MathWorks, I.}, \bibinfo{year}{{A}ccessed on: Apr. 08,
  2021}.
\newblock \bibinfo{title}{Design time series {NARX} feedback neural networks}.
\newblock \URLprefix
  \url{https://ch.mathworks.com/help/deeplearning/ug/design-time-series-narx-feedback-neural-networks.html}.
\bibitem[{Turner and Miller(2014)}]{Turner2014}
\bibinfo{author}{Turner, A.J.}, \bibinfo{author}{Miller, J.F.},
  \bibinfo{year}{2014}.
\newblock \bibinfo{title}{Neuroevolution: Evolving heterogeneous artificial
  neural networks}.
\newblock \bibinfo{journal}{Evolutionary Intelligence} \bibinfo{volume}{7},
  \bibinfo{pages}{135--154}.
\newblock \DOIprefix\doi{10.1007/s12065-014-0115-5}.
\bibitem[{Vincent et~al.(2006)Vincent, Bogatyreva, Bogatyrev, Bowyer and
  Pahl}]{Vincent2006}
\bibinfo{author}{Vincent, J.F.V.}, \bibinfo{author}{Bogatyreva, O.A.},
  \bibinfo{author}{Bogatyrev, N.R.}, \bibinfo{author}{Bowyer, A.},
  \bibinfo{author}{Pahl, A.K.}, \bibinfo{year}{2006}.
\newblock \bibinfo{title}{Biomimetics: Its practice and theory}.
\newblock \bibinfo{journal}{Journal of The Royal Society Interface}
  \bibinfo{volume}{3}, \bibinfo{pages}{471--482}.
\newblock \DOIprefix\doi{10.1098/rsif.2006.0127}.
\bibitem[{Yaot(1993)}]{Yaot1993}
\bibinfo{author}{Yaot, X.}, \bibinfo{year}{1993}.
\newblock \bibinfo{title}{A review of evolutionary artificial neural networks}.
\newblock \bibinfo{journal}{International Journal of Intelligent Systems}
  \bibinfo{volume}{8}, \bibinfo{pages}{539--567}.
\newblock \DOIprefix\doi{https://doi.org/10.1002/int.4550080406}.
\bibitem[{Yu et~al.(2021)Yu, Gonzalez and Li}]{Yu:2021}
\bibinfo{author}{Yu, W.}, \bibinfo{author}{Gonzalez, J.}, \bibinfo{author}{Li,
  X.}, \bibinfo{year}{2021}.
\newblock \bibinfo{title}{Fast training of deep {LSTM} networks with guaranteed
  stability for nonlinear system modeling}.
\newblock \bibinfo{journal}{Neurocomputing} \bibinfo{volume}{422},
  \bibinfo{pages}{85--94}.
\newblock \DOIprefix\doi{https://doi.org/10.1016/j.neucom.2020.09.030}.

\end{thebibliography}

\pagebreak

\bio[width=0.1\textwidth]{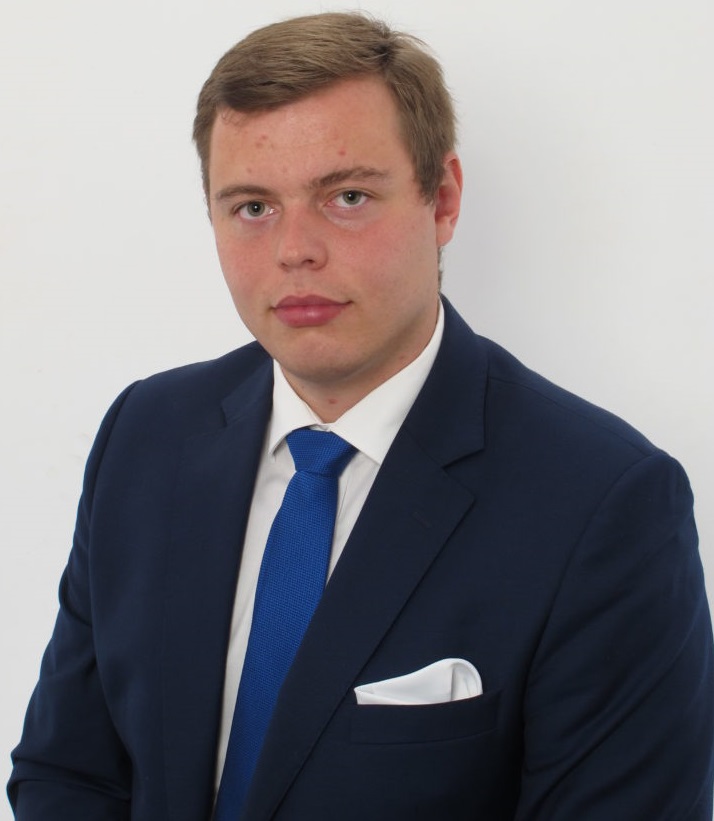}
Krzysztof Laddach received the M.Sc. degree (Hons.) in control engineering from the Faculty of Electrical and Control Engineering, the Gda\'nsk University of Technology in 2019. His master's dissertation was focused on algorithms of artificial neural network structure search for processes dynamics modelling purposes. Since October 2019 he has been a Ph.D. student in the Doctoral School at the Gda\'nsk University of Technology. His research interests involve mathematical modelling, especially in connection with the use of computational intelligence (artificial intelligence), optimal selection of the architecture of artificial neural networks, and the use of artificial neural networks in estimation.
\endbio

\bigskip
\bigskip
\bigskip

\bio[width=0.1\textwidth]{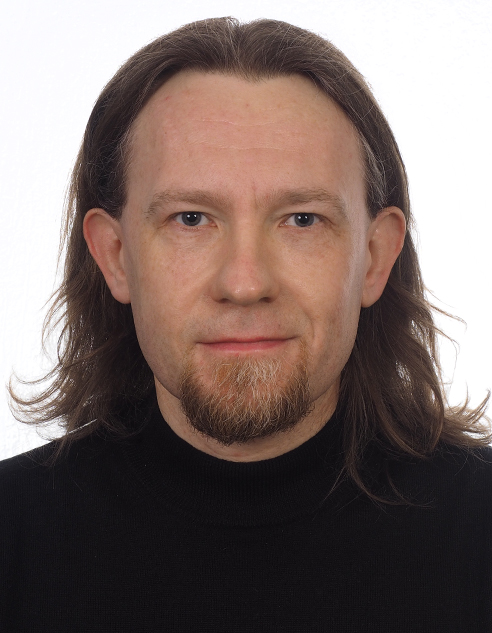}
Rafa\l{} \L{}angowski received the M.Sc. and the Ph.D. degrees (Hons.) in control engineering from the Faculty of Electrical and Control Engineering at the Gda\'nsk University of Technology in 2003 and 2015, respectively. From 2007 to 2014, he held the specialist as well as manager positions at ENERGA, one of the biggest energy enterprises in Poland. Since February 2014, he has been an owner of VIDEN a business in energy and control areas. He provides theoretical and practical experience, especially in front and back office at energy company and operation of the energy market in Poland. From 2016 to 2017, he was a Senior Lecturer with the Department of Control Systems Engineering at the Gda\'nsk University of Technology. He is currently an Assistant Professor with the Department of Electrical Engineering, Control Systems and Informatics. His research interests involve mathematical modelling and identification, estimation methods, especially state observers, and monitoring of large scale complex systems.
\endbio

\bigskip
\bigskip
\bigskip

\bio[width=0.1\textwidth]{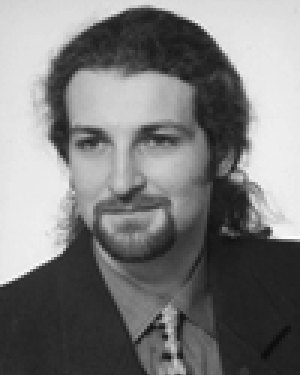}
Tomasz A. Rutkowski received the M.Sc. degree in automatics and robotics and the Ph.D. (Hons.) degree in automatic control from the Gda\'nsk University of Technology in 2000 and 2004, respectively. He has experience in control systems of critical infrastructure, such as environmental systems (drinking water distribution systems and wastewater treatment plant systems) and power systems, including nuclear power plants. He is currently an Assistant Professor with the Department of Electrical Engineering, Control Systems and Informatics, Gda\'nsk University of Technology. His current research interests include mathematical modelling, advanced control algorithms, estimation algorithms, computational intelligence techniques, and industrial control systems. 
\endbio

\bigskip
\bigskip
\bigskip

\bio[width=0.1\textwidth]{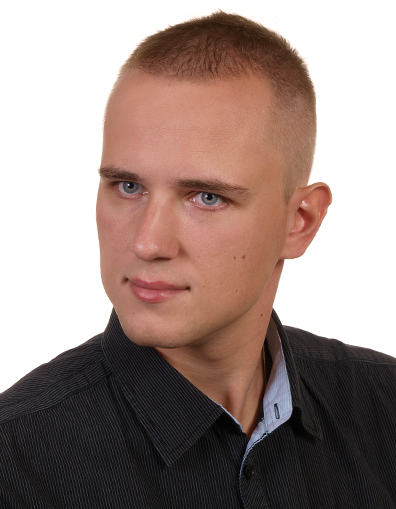}
Bartosz Puchalski received a MSc and PhD (Hons.) degree in control engineering from the Faculty of Electrical and Control Engineering at the Gda\'nsk University of Technology in 2011 and 2018 respectively. In 2012 he began to work at the Gda\'nsk University of Technology as a lecturer. In 2013, he received the engineer degree in the discipline of Electrical Engineering at the same Faculty. From 2018, he is employed at his alma mater as an assistant professor. Currently, his main scientific interests are focused on the control, identification and modelling of dynamic systems, fractional order calculus and recursive neural networks. 
\endbio

\end{document}